\documentclass[twoside,11pt]{article}

\usepackage{jmlr2e}

\usepackage{array}
\usepackage{booktabs}
\usepackage[dvipsnames]{xcolor}
\usepackage{tikz}
\newcolumntype{M}[1]{>{\centering\arraybackslash}m{#1}}

\usepackage[ruled,linesnumbered]{algorithm2e}

\usepackage{multirow}

\usepackage{longtable}

\usepackage{xcolor}
\usepackage{array}
\newcolumntype{L}[1]{>{\raggedright\let\newline\\\arraybackslash\hspace{0pt}}m{#1}}
\newcolumntype{C}[1]{>{\centering\let\newline\\\arraybackslash\hspace{0pt}}m{#1}}
\newcolumntype{R}[1]{>{\raggedleft\let\newline\\\arraybackslash\hspace{0pt}}m{#1}}
\newcommand{\bx}{\ensuremath{\mathbf{x}}}
\newcommand{\bs}{\ensuremath{\mathbf{s}}}

\newcommand{\bz}{\ensuremath{\mathbf{z}}}

\newcommand{\ba}{\ensuremath{\mathbf{a}}}

\usepackage{bbold}

\definecolor{dgreen}{RGB}{63, 175, 115}

\usepackage{breqn}

\DeclareMathAlphabet{\mathpzc}{OT1}{pzc}{m}{it}
\DeclareMathOperator*{\argmin}{arg\,min}

\usepackage{subcaption}

\newcolumntype{M}[1]{>{\centering\arraybackslash}m{#1}}

\jmlrheading{1}{2022}{1-36}{05/31}{N/A}{lash22a}{Michael T.~Lash}

\ShortHeadings{HEX via Deep RL}{Lash}
\firstpageno{1}

\begin{document}

\title{HEX: Human-in-the-loop Explainability via Deep Reinforcement Learning}

\author{\name Michael T. Lash 
        \email michael.lash@ku.edu\\
        \addr School of Business \\ University of Kansas \\ Lawrence, KS 66045}

\editor{None Yet Assigned}

\maketitle

\begin{abstract}
The use of machine learning (ML) models in decision-making contexts, particularly those used in high-stakes decision-making, are fraught with issue and peril since a person – not a machine – must ultimately be held accountable for the consequences of the decisions made using such systems. Machine learning explainability (MLX) promises to provide decision-makers with prediction-specific rationale, assuring them that the model-elicited predictions are made \textit{for the right reasons} and are thus reliable. Few works explicitly consider this key human-in-the-loop (HITL) component, however. In this work we propose HEX, a human-in-the-loop deep reinforcement learning approach to MLX. HEX incorporates 0-distrust projection to synthesize decider-specific explanation-providing policies from any arbitrary classification model. HEX is also constructed to operate in limited or reduced training data scenarios, such as those employing federated learning. Our formulation explicitly considers the decision boundary of the ML model in question, rather than the underlying training data, which is a shortcoming of many model-agnostic MLX methods. Our proposed methods thus synthesize HITL MLX policies that explicitly capture the decision boundary of the model in question for use in limited data scenarios.
\end{abstract}


\begin{keywords}
explainability, human-in-the-loop, deep reinforcement learning, federated learning, interpretability
\end{keywords}

\section{Introduction}
Over the years machine learning has become increasingly adopted and incorporated into decision-focused platforms. Such platforms provide predictions that entail some follow-up action depending on the outcome predicted. For instance, Infervision provides deep learning models to hospitals to detect the presence of cancer in CT image scans. These are clinically employed in China to help supplement the radiological workforce deficit \citep{marr2019artificial}. In this example, if the system detects that a patient has cancer, then a course of treatment is prescribed (follow-up action). Otherwise, no further action is taken. In either case, an incorrect prediction produces serious consequences.

Ultimately, however, machines cannot be held accountable for such high-stakes decisions \citep{neri2020artificial,doshi2017accountability}. When a decision based on a machine-elicited prediction goes awry, an individual (i.e., person) is necessarily held accountable for the consequences of that decision. Such human-machine accountability has consequently given rise to a variety of human-in-the-loop (HITL) machine learning paradigms. The idea of such paradigms is to incorporate, augment, or aid the human decision-maker by providing machine-elicited insights that are readily interpreted by the decision-maker.

One such HITL paradigm is machine learning explainability (MLX), otherwise referred to as artificial intelligence explainability (AIX). In this context decision platform-embedded machines not only provide predictions (classifications) as to the outcome of the decision task at hand, but a readily interpretable explanation as to why such a prediction was made as well. The explanation provided by the machine provides a contextual lens that allows the decision-maker to understand why the machine made the prediction that it did. If the rationale seems reasonable to the decision-maker they can choose to trust the prediction. If, on the other hand, the rationale does not make sense or seems counter-intuitive, the decision-maker can choose to disregard the prediction and seek solutions and insights elsewhere (using a different model, their own expertise, etc.).

Many MLX and AIX methods have been proposed over the years, including SHAP \citep{lundberg2017unified}, LIME \citep{ribeiro2016should}, TREPAN \citep{craven1995extracting}, MUSE \citep{lakkaraju2019faithful}, and MAPLE \citep{plumb2018model}, to name just a few. Relatively few works, however, explicitly incorporate the human element \citep{hitzler2022human, druce2021brittle, de2021classification}. Furthermore, many works rely on training a local surrogate model \citep{ribeiro2016should, lundberg2017unified,lakkaraju2019faithful,plumb2018model} or some other computationally complex procedure for each instance in need of explanation. These surrogate methods therefore model, and thus provide explanations for the training data, rather than ML model in question, a major short-coming. Finally, many sensitive decision scenarios may employ more recent privacy-preserving techniques, such as federated learning \citep{liu2021fate,salehkaleybar2021one}, which may greatly reduce the data points available to train local, instance-specific models.

In this work, we propose a novel model-agnostic human-in-the-loop classification explanation method, HEX, that addresses the above-mentioned challenges. We contextualize our method around the search for an instance-specific explanatory point -- the point nearest the instance in question that lies on the decision boundary, a new mode of explanation that is similar to, yet distinct from that of typical counterfactual explanation methods. The explanatory point promises to be superior since the decision boundary of the ML model in question is explicitly considered. We frame our search for such a point as a Markov Decision Process (MDP), thus allowing us to adopt a deep reinforcement learning (DRL) approach, from which we synthesize explanatory point-producing policies. We adopt two state-of-the-art actor-critic methods, DDPG \citep{lillicrap2015continuous} and TD3 \citep{fujimoto2018addressing}, subsequently augmenting these with several HITL and federated learning-motivated innovations and improvements. 

In particular, we propose and define two related phenomena, which we refer to as buffer degeneracy and policy degeneracy. Buffer degeneracy states that a distribution of sub-par actions, derived from a fixed policy, produces a distribution of sub-par rewards. Policy degeneracy, on the other hand, states that sub-par policies, synthesize from a fixed distribution of training instances, also produce sub-par rewards. We prove that buffer degeneracy implies policy degeneracy, using this proof as a basis for a greedy policy-learning strategy.

Furthermore, we incorporate a HITL element into our proposed method by considering the preferred explanatory features of a HITL decider and learning an MLX policy that is specific to these features (i.e., a decider-specific MLX policy). We do so by defining the domain of the decider's preferred features as the feasible action region, projecting the elicited actions onto this region during policy synthesis. We term this 0-distrust project and can thus guarantee that all policy-produced actions are intuitive to the decider in question.

Finally, we incorporate federated learning considerations into our method. First, our MDP conceptualization, coupled with the exploratory, off-policy behavior of our adopted DRL methods ensures that the probability space is explored beyond the training instances initially supplied to the learning process. Thus, the limited-data problem is naturally mitigated. Class imbalance may still permit the issue to exist, however. We therefore further augment our method with synthetic class balancing behavior to mitigate this additional concern.

We evaluate our proposed methods, HEX-DDPG and HEX-TD3, against their out-of-the-box DRL counterparts and two competing state-of-the-art explainability methods on two scenarios, one that is HITL decider-free and another that explicitly consider a HITL decider. Our investigation in the context of these scenarios spans five real-world decision-focused datasets and five machine learning (ML) models trained on each. Our experiments show that our methods perform well in both scenarios, particularly the HITL decider scenario.

\section{Our Problem Setting and Related Works}

To further clarify the disposition of this research we provide a brief discussion of related works, highlighting where our proposed method fits in to the broader MLX landscape as the discussion progresses. First, however, we provide some preliminary notation to clarify our problem setting and facilitate discussion of related work.

Our problem setting considers an arbitrary classification model $f$. We assume that $f$ takes an instance $\bx \in \mathcal{X}=[0,1]^p$ as input and produces a probability estimate or classification score $\hat{\mathbb{p}}$ as to the class $c \in \mathcal{C}$ the instance belongs to. To simplify our discussion we will assume w.l.o.g.~that $f$ outputs probabilities\footnote{If an adopted $f$ does not natively produce probabilities, Platt Scaling \citep{platt1999probabilistic} can be employed to obtain such outputs.}. This work specifically considers binary classification problems -- i.e., $\mathcal{C} = \{0,1\}$\footnote{Future work will consider multi-class formulations, which are discussed in the conclusions and future work section.}. The model-elicited probabilities are used with a decision rule $\omega$ and decision function $\Omega$ to produce classifications:
\begin{align}
\label{eq:decisionfunc}
    \hat{c} = \Omega\left(f\left(\bx \right),\omega \right) \triangleq
    \begin{cases}
        \text{if } f\left(\bx \right) < \omega: & 0 \\
        \text{else:} & 1
    \end{cases}.
\end{align}

This work proposes to learn an explanation-producing policy $\pi:\mathcal{X} \rightarrow \mathcal{Z}_x$, $\mathcal{Z}_x =\{\bz:\min\left(\mathcal{X}\right)-\bx \leq \bz \leq \max\left(\mathcal{X}\right)-\bx\}=\{\bz:\mathbf{0}-\bx \leq \bz \leq \mathbf{1}-\bx\}$, that considers $\bx$ and produces $\bz$ such that, when $\bz$ is additively applied to $\bx$, an explanatory point $\bx^{\prime}$ is produced -- i.e.,~$\bx^{\prime} = \bx + \bz$. Definition \ref{def:exp-point}, below, formalizes the notion of an explanatory point.
\begin{definition}
\label{def:exp-point}
 The \textbf{explanatory point} $\bx^{\prime}$ of $\bx$ is
\begin{align}
    \bx^{\prime} \triangleq \argmin_{\bx^{\prime} \in \mathcal{X}}\{\Vert \bx^{\prime} - \bx \Vert : f(\bx^{\prime}) = \omega\},
\end{align}
where $\Vert \cdot \Vert$ is  the Euclidean norm. In other words, $\bx^{\prime}$ is the point nearest $\bx$ that falls on decision boundary $\omega$.
\end{definition}
The explanatory point does not itself provide the explanation of instance $\bz$, but functions as a decision boundary reference. The explanation is provided by $\bz$. The policy $\pi$ thus directly produces explanations. To further clarify how $\bz$ provides an explanation of $\bx$ we provide two examples in Figure \ref{fig:exp-ex}, with Figure \ref{fig:exp-ex-pos} showing the explanation of a positive classification and Figure \ref{fig:exp-ex-neg} showing a negative classification.
\begin{figure}[!htp]
    \centering
    \begin{subfigure}[b]{0.49\textwidth}
        \centering
        \includegraphics[scale=3.0]{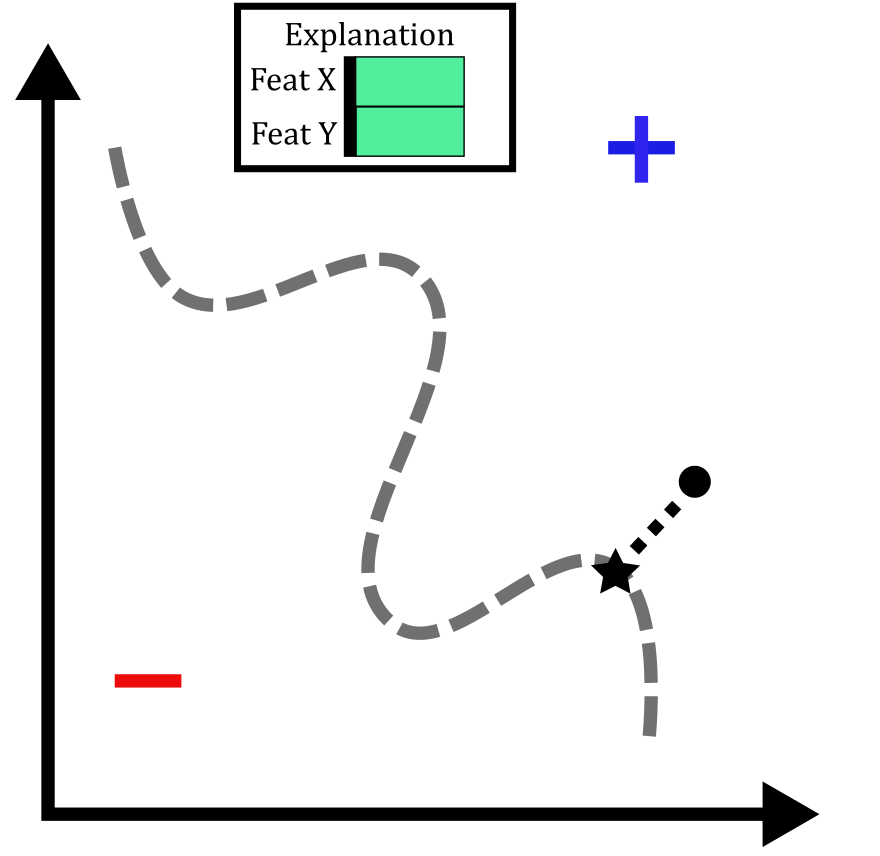}
        \caption{Positive example.}
        \label{fig:exp-ex-pos}
    \end{subfigure}
    \begin{subfigure}[b]{0.49\textwidth}
        \centering
        \includegraphics[scale=3.0]{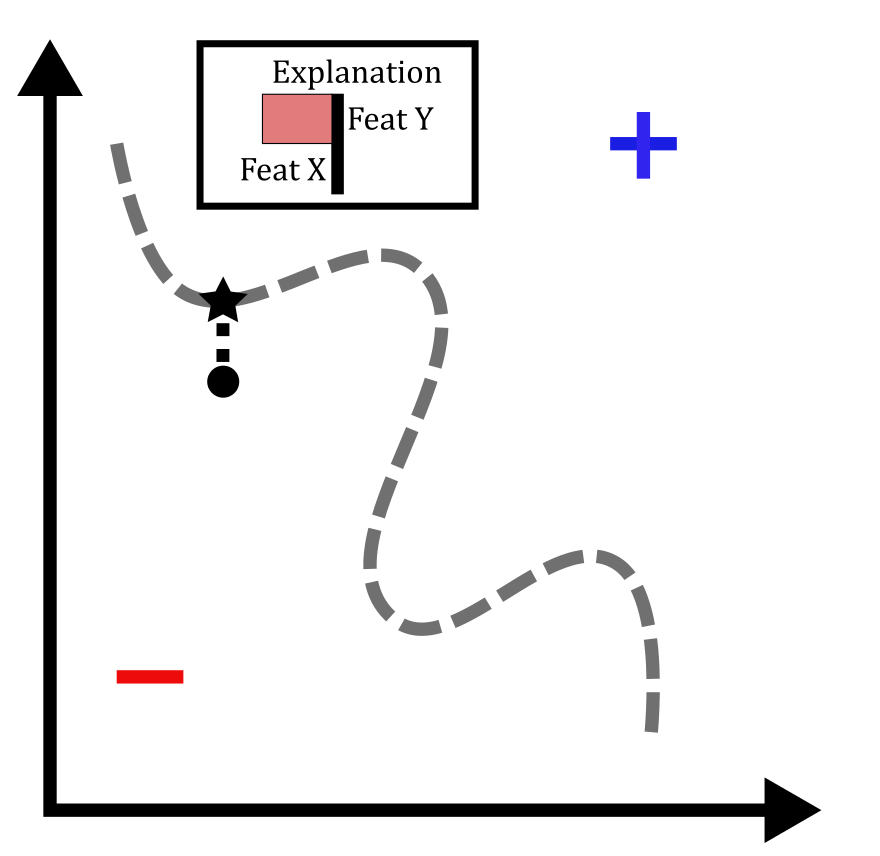}
        \caption{Negative example.}
        \label{fig:exp-ex-neg}
    \end{subfigure}
    \caption{Two toy explanatory point examples.}
    \label{fig:exp-ex}
\end{figure}
Explanations are provided by sorting the values $\vert \bz \vert$ from highest to lowest magnitude and then displaying these to a human-in-the-loop decider $k$. To ensure that $k$ is not overwhelmed and the explanation is simple, only the largest $q$ values need be actually displayed (in the form of a bar chart) \citep{ribeiro2016should}. Ideally these features will align with $k$'s intuition. We describe how our method provides decider-specific explanations, which guarantee this, in Section \ref{sec:hitl}. The benefits of using the proposed explanatory point over other modes of explanation are elucidated on at the end of the next subsection.

\subsection{Explainability Methods}

The general idea of MLX is to provide some type of \textit{explanation} as to why $f$ and $\omega$ produce the classification(s) that they do. Both the methods that provide such explanations and the manner in which the explanations are themselves provided are varied and can thus be decomposed and discussed along a variety of different dimensions.

One such dimension is local vs. global explainability. Methods that provide global explanations explain the model as a whole, often doing so through the lens of the entire (global) dataset. Such methods include iForest \citep{zhao2018iforest}, Partial Dependence Plots (PDP) \citep{goldstein2015peeking}, Gradient Feature Auditing (GFA) \citep{adler2018auditing}, and Golden Eye \citep{henelius2014peek}. On the other hand, local explainability methods provide explanations as to why individual instances are classified as they are. This paper focuses on local explanations since our problem setting specifically considers decisions made with respect to individual predictions. Furthermore, our method learns a so-called \textit{surrogate explainer model}, a separate model that provides explanations of $f$, $\Omega$, and $\omega$, which is a popular approach to the MLX problem. Local surrogate explanation methods include LIME \citep{ribeiro2016should}, SHAP \citep{lundberg2017unified} and ShapNet \citep{wang2021shapley}, PatternNet \citep{kindermans2017learning}, MUSE \citep{lakkaraju2019faithful}, MAPLE \citep{plumb2018model}, and Growing Spheres \citep{laugel2018comparison}, among many others.

Global surrogate methods are a special case of local explanation methods that attempt to approximate black box models through an interpretable surrogate \cite{hall2017machine,andrzejak2013interpretable,hinton2017distilling,hara2018making,ribeiro2016should,schetinin2007confident}. Our method is similar to these in that we attempt to globally approximate $f$ wrt. decision boundary $\omega$. However, rather than learn an interpretable approximation of $f$, our method learns a policy $\pi$ that directly produces explanations $\bz$. \cite{samoilescu2021model} also use RL to produce counterfactual explanations (discussed in the next paragraph), adopting an encoder-decoder approach to reduce the dimensionality of the feature space and focusing on the incorporation and modeling of various data modalities.

Local explanations are elicited in a variety of ways. Gradient-based methods explain differentiable models using extracted gradient information (or linear model coefficient) \citep{ribeiro2016should,adler2018auditing,lash2017budget,cortez2011opening,shrikumar2017learning}. Gradient-based methods are not totally model-agnostic, however, since the model must be differentiable, thus precluding decision trees, random forests, etc. Propagation- and sensitivity-based methods work by propagating the prediction backward through the model and examining the relevant inputs \citep{datta2016algorithmic,montavon2018methods,goldstein2015peeking,hall2017machine}. Perturbation-based methods manipulate the inputted feature values of an instance, examining the corresponding output as a means of explanation \citep{smilkov2017smoothgrad,lash2017generalized, samek2017explainable}. Similarly, counterfactual methods provide information about how the classification can be changed, thus providing an explanation through a counterexample \citep{laugel2018comparison,martens2014explaining,samoilescu2021model,tsirtsis2021counterfactual}. Prototype methods display (training) data points that are similar to the instance in need of explanation \citep{kim2014bayesian,kim2016examples}. This paper proposes the use of an explanatory point, which explicitly considers the decision boundary $\omega$. The explanatory point therefore explicitly considers classification model $f$, thus providing explanations of the ML model itself, rather than the training data, which is a common approach of other model-agnostic methods.

With the above in mind, our proposed method can simultaneously be considered a prototypical, counterfactual, and perturbation-based method. Readers interested in a more in-depth discussion of MLX and AIX methods and approaches may wish to consult \cite{burkart2021survey,linardatos2020explainable,beaudouin2020flexible,wang2022unified,biecek2021explanatory}, among other well-written pieces on the subject.

\subsection{Markov Decision Processes and Reinforcement Learning Methods}

This work proposes to learn a policy $\pi$ that produces an explanatory point $\bz$ provided some inputted instance $\bx$, arbitrary classification function $f$, decision function $\Omega$, and decision boundary $\omega$. To learn such a policy we take a Markov Decision Process (MDP) approach, ultimately adopting deep reinforcement learning methodologies to learn such a policy\footnote{Note that this is not a reinforcement learning explainability (XRL) method/work. XRL methods seek to explain reinforcement learning models themselves. In this work, we use RL as a means of providing classification explanations.}. We adopt this approach for five primary reasons:
\begin{enumerate}
    \item The explanatory point generating process is tantamount to approximating $f(\bx) = \omega: \bx \in \mathcal{X}$. Reinforcement learning is particularly suitable for function approximation \cite{sutton1999policy,melo2008analysis,xu2014reinforcement}.
    \item Deep reinforcement learning uses neural networks, which are universal function approximators \cite{hornik1989multilayer}, further strengthening the rationale of (1), above.
    \item The Markov Property is appropriate for the MLX problem setting since we are attempting to explain individual data points independent of one another (local explanations).
    \item Reinforcement learning methods belong to a broader class of approximate dynamic programming methods. Among this broader class of methods, reinforcement learning does not rely on limiting distributional assumptions, which is beneficial since we wish to make as few assumptions about $\mathcal{X}$ and $f$ as possible.
    \item By learning a policy that produces explanatory points, we avoid the computational overhead associated with many MLX methods that algorithmically construct an explanation for each explained instance \citep{ribeiro2016should,lundberg2017unified,lakkaraju2019faithful,samoilescu2021model}.
\end{enumerate}

\begin{figure}[!htp]
    \centering
    \includegraphics[scale=0.50]{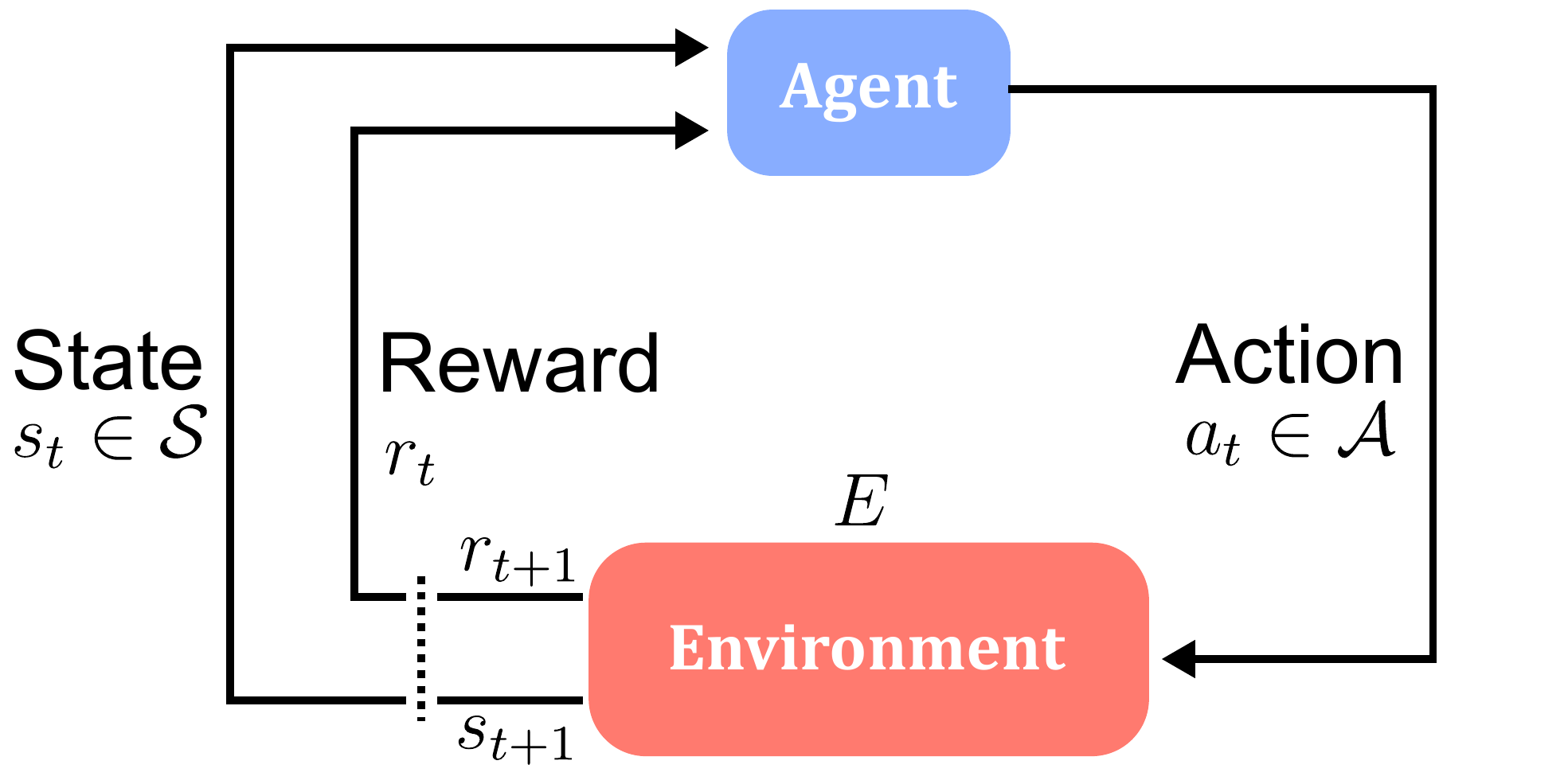}
    \caption{Standard MDP problem setting.}
    \label{fig:mdp-orig}
\end{figure}

MDPs are cast in terms depicted by Figure \ref{fig:mdp-orig} and include a state space $\mathcal{S}$, action space $\mathcal{A}$, environment $E$, and reward function $R$.
The current state $\bs_t \in \mathcal{S}$, observed at iteration $t$, is a numerical representation describing the ``current, observable state of affairs'' of the phenomena being modeled. Some form of so-called agent considers $\bs_t$ and produces actions $\ba_t$. These actions are then applied via a state transition function to produce $\bs_{t+1}$ and subsequently observed in environment $E$. The so-called \textit{goodness} of taking these actions is assessed via a defined reward function $R$, which produces reward $r_{t+1}$.

The agent is represented by some function or algorithmic process. Function approximation methods attempt to approximate either the policy, value, or both, which then dictate how the agent takes actions. Policy approximation, often referred to as actor-only methods, learn an action-producing policy directly \citep{schulman2015trust,sutton1998introduction,sutton1999policy}. Value approximation methods, on the other hand, attempt to approximate the so-called \textit{value} of being in a particular state from which a policy can be indirectly inferred \cite{watkins1992q,mnih2013playing,sutton1998introduction}. Some methods estimate both the policy (actor) and value (critic) and are thus referred to as actor-critic methods \citep{konda2000actor,lillicrap2015continuous,fujimoto2018addressing}. This work adopts an actor-critic approach for reasons that will be discussed shorty in the next section.

\section{The Proposed Method}

\begin{figure}[!htp]
    \centering
    \includegraphics[scale=0.50]{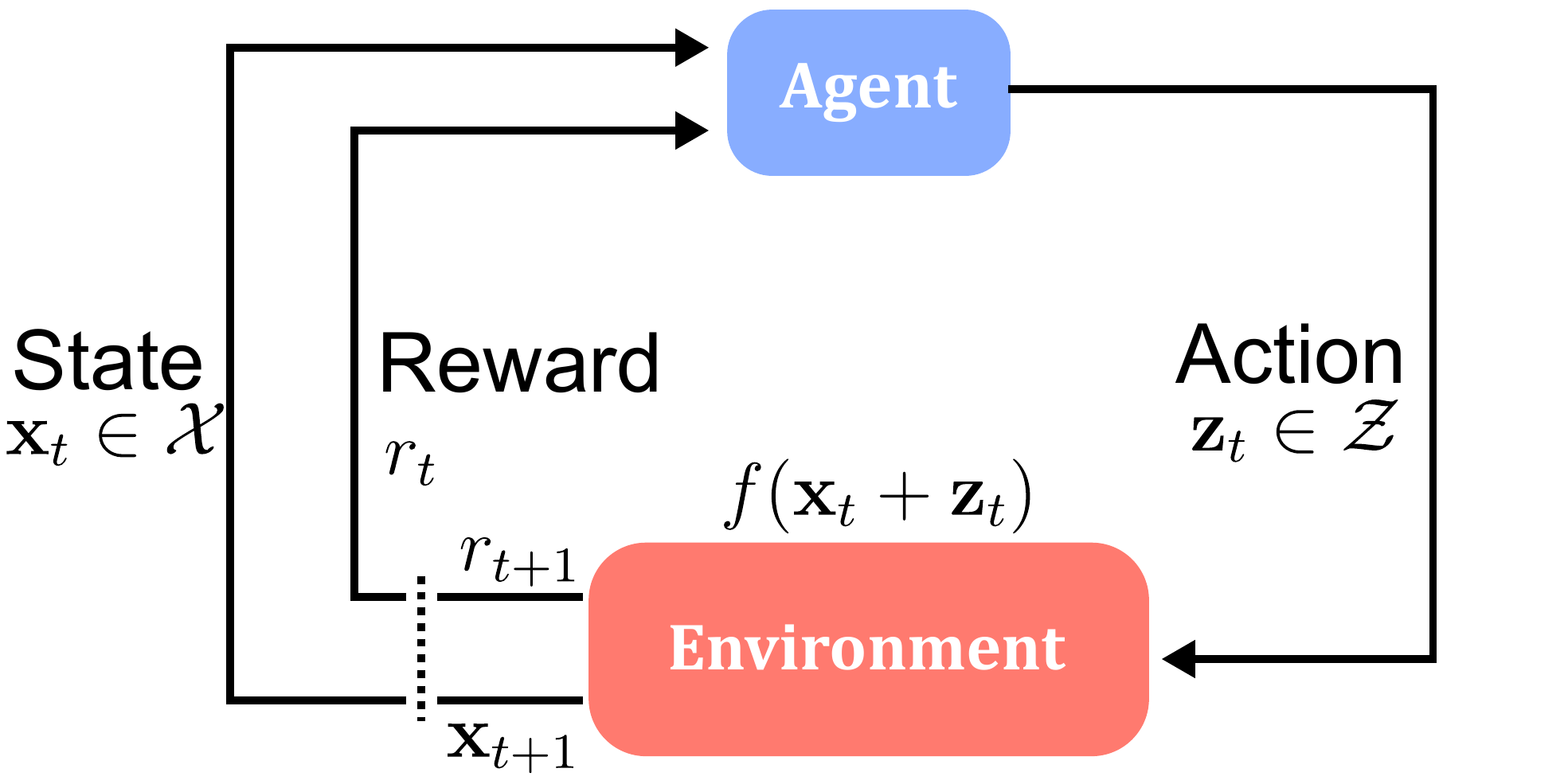}
    \caption{MLX conceptualized as an MDP.}
    \label{fig:mdp-alt}
\end{figure}

Our problem setting considers continuously-valued state and action spaces. We define $\mathcal{X}$ as our state space and $\mathcal{Z}_x$ as our action space\footnote{Future work may wish to consider discrete and ordinal explainability problems.}. Further, we define $f$ as our environment and element-wise vector addition as our transition function -- i.e.,~$\bx_{t+1} = \bx_{t} + \bz_{t}$. Figure \ref{fig:mdp-alt} clarifies the conceptualization of our problem setting as an MDP.

\subsection{An MLX Reward Function}

The reward $r_{t+1}$ of Figure \ref{fig:mdp-alt} is assessed using a defined explanatory point-based MLX reward function $R(\cdot)$, based on Definition \ref{def:exp-point}. Our reward function is disclosed in \eqref{eq:exp-reward}, below:
\begin{align}
    \label{eq:exp-reward}
    R\left(\mathbf{x}_t,\mathbf{z}_t \right) \triangleq & -\alpha \left[f\left(\mathbf{x}_{t+1}\right)- \left(\omega+ \epsilon_{x} \right)\right]^2 + \beta \left[ \Omega\left(f\left(\mathbf{x}_{t+1}\right),\omega \right) -\Omega\left(f\left(\mathbf{x}_{t}\right), \omega \right)\right]^2 \\ \nonumber
    & - \Vert \mathbf{z}_t \Vert_2 - \frac{1}{p}\vert \mathbf{z}_t \vert_0,
\end{align}
where $\alpha,\beta,\epsilon_x > 0$. The first term of \eqref{eq:exp-reward} encourages solutions that are as close to $\omega$ as possible. Solutions that are further from $\omega$ incur larger penalties. The additive term $\epsilon_x$ is included to facilitate discovery of the decision boundary $\omega$ by encouraging solutions that slightly crest this threshold. Experimentally, we found that the quality of our solutions improved once $\epsilon_x$ was included. While a variety of strategies and values were explored, we set $\epsilon_x$ according to:
\begin{align}
    \label{eq:eps-add}
    \epsilon_x = \left \{
    \begin{array}{cc}
         .01 & \text{if }\Omega(f(\mathbf{x}_t), \omega) = 0  \\
         -.01 &  \text{otherwise}
    \end{array} \right.
\end{align}
While the first term, which maximizes $R$ at $f(\bx_{t}+\bz_{t}) = \omega+\epsilon_x$, penalizes solutions that are further from the decision boundary, the second term of \eqref{eq:exp-reward} rewards solutions that flip the classification of the instance $\bx_{t}$. This term works in conjunction with the $\epsilon_x$ value of the first term and is motivated by the objective function of \citep{laugel2018comparison}. The inclusion of this term vastly improved the policy-learning process since it rewards, rather than penalizes discovery of the decision boundary\footnote{This was observed anecdotally as the reward function and methods were developed.}. $R$ is maximized through this term at $\Omega\left(f\left(\mathbf{x}_{t+1}\right),\omega \right) -\Omega\left(f\left(\mathbf{x}_{t}\right), \omega \right)=1$. The third term penalizes solutions that are further from $\bx_{t}$ and has an immediate correspondence to Definition \ref{def:exp-point}, since we want the point nearest $\bx_t$ that falls on the decision boundary. This term is also included in \cite{laugel2018comparison,samoilescu2021model} and maximizes $R$ at $\Vert \bz_{t} \Vert_2 \rightarrow 0$. The fourth (last) term $\frac{1}{p}\vert \bz_{t} \vert_0$ penalizes solutions that are less sparse, and therefore less interpretable, through the $0$-norm \citep{laugel2018comparison,samoilescu2021model}. Like the preceding term, $R$ is maximized as $\vert \bz_{t} \vert_0 \rightarrow 0$.

In our experiments, we set $\alpha=4, \beta=10$, and $\epsilon_x$ according to \eqref{eq:eps-add} after exploring a variety of values and observing that these worked well and produced stable results across several datasets.

\subsection{Learning an Explanatory Point-producing Policy}

To learn an explanatory point-producing policy $\pi$, we embed our reward function $R(\cdot)$ in an actor-critic reinforcement learning framework, adopting two well-known deep reinforcement learning methods, DDPG \citep{lillicrap2015continuous} and TD3 \citep{fujimoto2018addressing}. We tailor these to the MLX problem setting and then further propose several additional MLX-specific innovations which account for HITL deciders and federated learning scenarios.

\begin{figure}[!htp]
    \centering
    \includegraphics[scale=0.50]{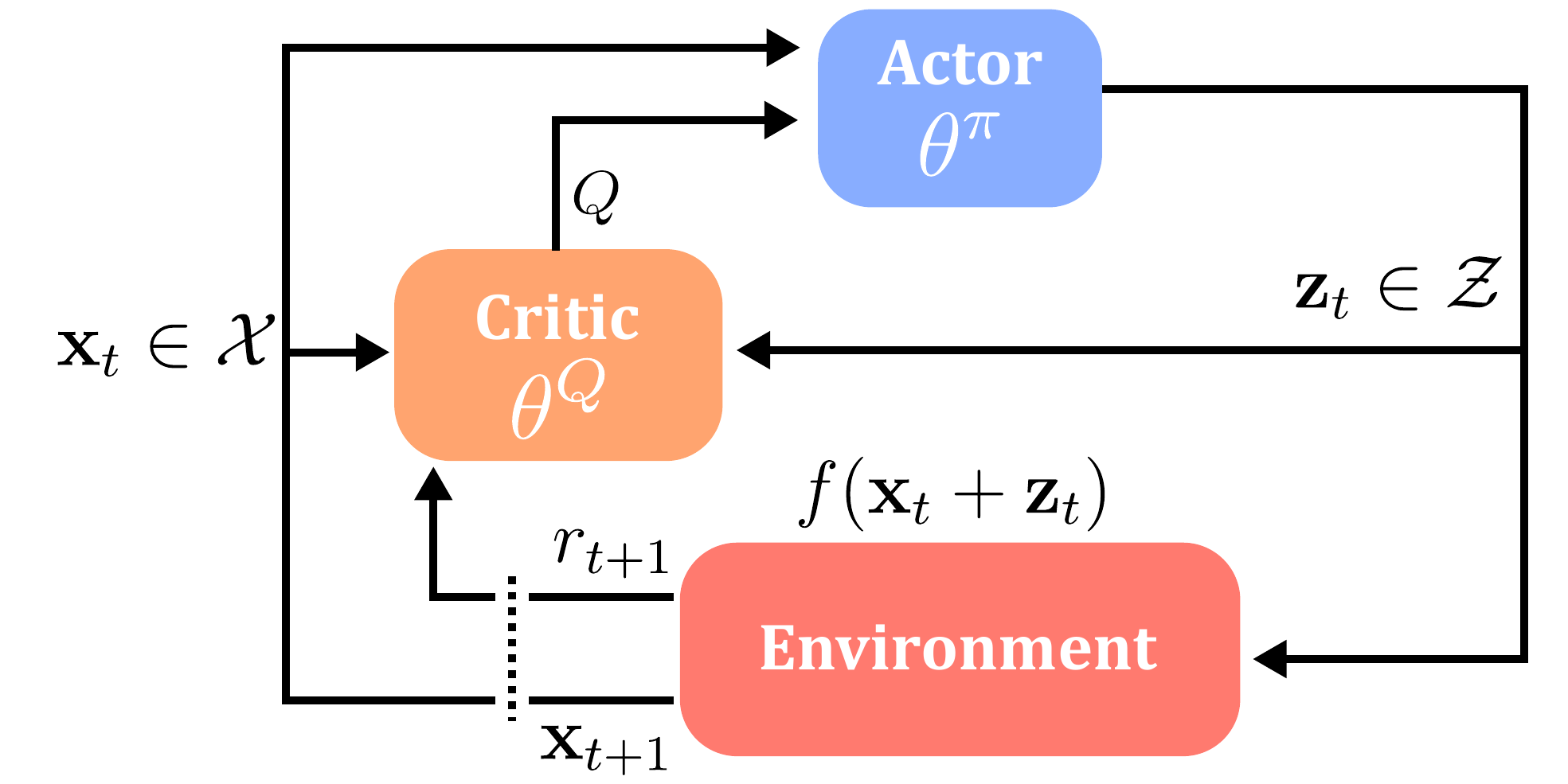}
    \caption{Caption}
    \label{fig:act-crit1}
\end{figure}

Figure \ref{fig:act-crit1} illustrates the adopted actor-critic framework in terms of our MLX MDP. Actor-critic methods work by approximating both the policy (actor) and the value (critic). The actor is responsible for estimating the policy and thus produces actions, while the critic is responsible for estimating the \textit{goodness} of the actions produced by the policy. Since the critic and actor are connected in this manner, if we can accurately assess/estimate action quality, we can learn an accurate policy.
More concretely, the actor and critic each have their own parameter sets -- i.e., they are two separate deep learning models -- but have a joint learning objective that begins with the critic. To elicit value estimates, the critic uses the well-known Q function, which is based on the Bellman Equation, in conjunction with our specific MLX reward function $R(\cdot)$. The learning objective of the critic is to therefore minimize the following loss function:
\begin{align}
    \label{eq:exp-loss}
    L\left(Q^{\theta}\right) \triangleq & \hspace{.2cm} \mathbb{E}_{\mathbf{x}_t \sim \rho^{\pi}, \mathbf{z}_t \sim \pi^{\theta}, r_t \sim R} \left[\left(Q \left(\mathbf{x}_t,\mathbf{z}_t \vert \theta^{Q} \right) - y_t^m \right)^2 \right],
\end{align}
where $\rho^{\pi}$ denotes a policy-dependent data distribution, $m\in\{DDPG,TD3\}$ such that
\begin{align}
    \label{eq:disc-fut-rewards-ddpg}
    y_t^{DDPG} = & \hspace{.2cm} R \left(\mathbf{x}_t,\mathbf{z}_t \right) + \gamma Q\left(\mathbf{x}_{t+1}, \check{\pi}\left( \mathbf{x}_{t+1} \vert \theta^{\pi}\right)\vert \theta^{Q} \right),
\end{align}
when the DDPG method is used and
\begin{align}
    \label{eq:disc-fut-rewards-td3}
    y_t^{TD3} = & \hspace{.2cm} R \left(\mathbf{x}_t,\mathbf{z}_t \right) + \gamma  \min_{i=1,2} Q\left(\mathbf{x}_{t+1}, \check{\pi}\left( \mathbf{x}_{t+1} \vert \theta^{\pi}\right)\vert \theta^{Q}_i \right),
\end{align}
when TD3 is used, and where
\begin{align}
    \label{eq:off-policy}
        \check{\pi}\left(\mathbf{x}_t \vert \theta \right) = & \pi\left(\mathbf{x}_t \vert \theta \right) + \mathcal{N}\left(\mathbf{0},\pmb{\epsilon} \right).
\end{align}
Put briefly, the loss is defined as the estimate elicited from the critic, parameterized by $\theta^Q$, minus the discounted, future rewards expressed by either \eqref{eq:disc-fut-rewards-ddpg} or \eqref{eq:disc-fut-rewards-td3}, where $0 < \gamma < 1$ is a discount parameter imposed on the estimate of future rewards. Note that the primary distinction between DDPG and TD3 is the use of two critics by TD3, over the single critic of DDPG. TD3 selects the minimum of the two estimates to help regularize the value estimation. Additionally, TD3 applies soft parameter updates periodically, rather than at every iteration -- we discuss this shortly.

To optimize the parameters $\theta^{\pi}$ of the actor the chain rule is applied to \eqref{eq:exp-loss}:
\begin{align}
    \label{eq:grad-pi}
    \nabla_{\theta^{\pi}}J \approx & \hspace{.2cm} \nabla_{\theta^{\pi}}L\left(Q^{\theta}\right)\\\nonumber
    = & \hspace{.2cm} \mathbb{E}\left[\nabla_{\theta^{\pi}}Q \left(\tilde{\bx},\tilde{\bz} \vert \theta^{Q} \right)\vert_{\tilde{\bx}=\bx_t,\tilde{\bz}=\pi\left(\bx_t \vert \theta^{\pi} \right)} \right]\\\nonumber
    = & \hspace{.2cm}\mathbb{E}\left[\nabla_{\tilde{\bz}}Q \left(\tilde{\bx},\tilde{\bz} \vert \theta^{Q} \right)\vert_{\tilde{\bx}=\bx_t,\tilde{\bz}=\pi\left(\bx_t\right)} \nabla_{\theta^{\pi}} \pi\left(\tilde{\bx} \vert \theta^{\pi} \right)\vert_{\tilde{\bx}=\bx_t} \right]\nonumber
\end{align}
This result follows that of \cite{lillicrap2015continuous}, who also point out that \cite{silver2014deterministic} proves this is the gradient that optimizes the policy (i.e.,~policy gradient). Note that the deterministic policy $\pi$ is used rather than the exploratory policy $\check{\pi}$.

To avoid the pitfalls associated with the use of Q Learning with deep neural networks, so-called \textit{target} Q and policy networks, $\theta^{Q^{\prime}}$ and $\theta^{\pi^{\prime}}$, are used. These networks are updated gradually using the parameters $\theta^{Q}$ and $\theta^{\pi}$ along with a scaling factor $\tau$. These are called \textit{soft} updates and can be expressed as:
\begin{align}
    \label{eq:soft-update}
    \theta^{Q^{\prime}} \leftarrow & \tau \theta^{Q} + (1-\tau) \theta^{Q^{\prime}} \\\nonumber
    \theta^{\pi^{\prime}} \leftarrow & \tau \theta^{\pi} + (1-\tau) \theta^{\pi^{\prime}}.
\end{align}

To compute the loss and optimize the parameters of both networks a sampling procedure is used in conjunction with a so-called \textit{experience replay buffer} \citep{mnih2013playing}, which we denote using $\pmb{\rho}$. The replay buffer stores tuples of the form $\left(\bx_t,\bz_t,r_t,\bx_{t+1}\right)$ across training episodes ($e=1,\dots,E$) and inner training iterations ($t=1,\dots,T$) according to some specified buffer size $\xi$:
\begin{align}
    \label{eq:buffer}
    \pmb{\rho}_{l \text{ mod } \xi} = \left(\bx_t,\bz_t,r_t,\bx_{t+1}\right)_l,
\end{align}
where $l = e \cdot t$. When $l > \xi$ previous buffer entries are overwritten. Therefore, if all observations are to be stored, a buffer of size $E \cdot T$ is needed. Note that we refer to distributions of specific sub-elements of $\pmb{\rho}$ using an index that refers to the element in question-- e.g.,~$\rho^{\pi}$ is the policy-dependent distribution.

At each update iteration the buffer is sampled uniformly $N$ times:
\begin{align}
    \left(\bx_t,\bz_t,r_t,\bx_{t+1}\right)_i \sim \mathcal{U}\left(\pmb{\rho}\right):i=1,\dots,N,
\end{align}
where the $N$ samples are used to compute the loss expressed in \eqref{eq:exp-loss}. To illustrate the training procedure we provide Algorithm \ref{algo:init-pol-learn}, highlighting the buffering of experience tuples in \textcolor{red}{red}. Projection onto the feasible action region (Line 6 of Algorithm \ref{algo:init-pol-learn}) is achieved by
\begin{align}
    \label{eq:proj}
    Proj_{\mathcal{Z}_x}\left(\bz\right) \triangleq & \min\left(\max\left(\mathbf{l}_z,\bz \right),\mathbf{u}_z\right)
\end{align}
where $\mathbf{l}_z$ and $\mathbf{u}_z$ are the lower and upper bounds of $\mathcal{Z}_x$
\begin{algorithm}[!htp]
	\caption{Initial MLX policy learning procedure}
	\label{algo:init-pol-learn}
    Initialize buffer $\mathcal{R} \gets \{\}$\\
    Execute a cold start procedure to provide initial samples to $\pmb{\rho}$.\\
	\For{$e=1,\dots,E$}{
        $\bx \sim \mathcal{D}$ (Sample an instance from policy-learning dataset $\mathcal{D}$)\\
		\For{$t=1,\dots,T$}{
		    $\bz_{t} \gets Proj_{\mathcal{Z}_x}\left(\check{\pi}\left(\mathbf{x}_t \vert \theta \right)\right)$ (Using \eqref{eq:off-policy} and \eqref{eq:proj}).\\
		    Estimate $r_t$ according to \eqref{eq:exp-reward}.\\
		    $\bx_{t+1} \gets \bx_{t} + \bz_{t}$ (Transition to the next state.)\\
		    \textcolor{red}{$\pmb{\rho}_{l \mod \xi} \gets (\bx_{t},\bz_{t},r_t,\bx_{t+1})_l$ (Buffer the sample.)}\\
		    \uIf{$m = DDPG$}{
                Sample $\pmb{\rho}$ and compute the loss according to \eqref{eq:exp-loss}.\\
                Backpropagate the loss to update $\theta^Q$ and use \eqref{eq:grad-pi} to update $\theta^\pi$.\\
                Apply soft updates according to \eqref{eq:soft-update}.
            }
            \uElseIf{$m = TD3$ and $t \mod 2 = 0$}{
                Sample $\pmb{\rho}$ and compute the loss according to \eqref{eq:exp-loss}.\\
                Backpropagate the loss to update $\theta^Q$ and use \eqref{eq:grad-pi} to update $\theta^\pi$.\\
                Apply soft updates according to \eqref{eq:soft-update}.
            }
		}
    }
	\KwResult{$\pi^{\theta^*}$}
\end{algorithm}

\subsection{Buffer and Policy Degeneracy}
The buffer is critically important to the successful learning of a policy that produces explanatory points. If the buffer fills up with ``bad'' examples, then the quality of the policy will be low: learning what is bad does not guarantee that the actions obtained from such a policy will be good. Furthermore, since the buffer depends in part on the policy itself, a negative feedback loop may be created such that the learning process becomes \textit{degenerate}. To better characterize this phenomenon we first define instance degeneracy, expanding this definition to characterize buffer degeneracy and policy degeneracy. We then provide a theorem and proof to show the linkage between the two.

Instance degeneracy is provided in Definition \ref{def:inst-degen}.
\begin{definition}
\label{def:inst-degen}
\textbf{Instance degeneracy}. A sampled instance $\bx \sim \rho^{\pi}_l$ from the policy-dependent data distribution at $l$ is said to experience degeneracy if
\begin{align}
    y_l > & y_{l^{\prime}}: l^{\prime} > l,
\end{align}
where $y_l$ is computed on \eqref{eq:disc-fut-rewards-ddpg} or \eqref{eq:disc-fut-rewards-td3} using $\pi_l^{\theta}$ and $\pi_{l^{\prime}}^{\theta}$ to produce actions and $Q_l^{\theta}$ and $Q_{l^{\prime}}^{\theta}$ to assess future discounted rewards.
\end{definition}
Definition \ref{def:inst-degen} state that an instance received a more favorable action at an earlier training iteration $l \leq l^{\prime}$ and forms the basis and motivation for our definition of both buffer degeneracy and policy degeneracy. Buffer degeneracy is characterized by Definition \ref{def:buff-degen}.
\begin{definition}
\label{def:buff-degen}
\textbf{Buffer degeneracy}. The buffer $\pmb{\rho}_{l^{\prime}}$ at iteration $l^{\prime}$ is to be degenerate if, under a fixed policy $\pi_{l}^{\theta}$ and fixed Q function $Q^{\theta}_l$,
\begin{align}
    \mathbb{E}_{y_l \sim \rho^{y}_l}\left[y_l\right] > & \mathbb{E}_{y_{l^{\prime}} \sim \rho^{y}_{l^{\prime}}}\left[y_{l^{\prime}}\right]: l^{\prime} > l,
\end{align}
where $\rho^{y}_l = \{y:\mathbf{x} \sim \rho^{\pi}_l, \mathbf{z} \sim \pi^{\theta}_l, r \sim R\}$ and  $\rho^{y}_{l^{\prime}} = \{y: \mathbf{x} \sim \rho^{\pi}_{l^{\prime}}, \mathbf{z} \sim \pi^{\theta}_{l}, r \sim R\}$.
\label{def:buff-degen}
\end{definition}
The definition of buffer degeneracy states that the expectation over the future discounted reward distribution, formed by assessing the buffered samples at $l$ using the policy and value (Q) function at $l$, is larger than using the same policy and value function on the buffered samples at $l^{\prime}$. This immediately gives rise to our characterization of policy degeneracy, provided in Definition \ref{def:pol-degen}, below.
\begin{definition}
\label{def:pol-degen}
\textbf{Policy degeneracy}. A policy $\pi_{l^{\prime}}$ at iteration $l^{\prime}$ is to be degenerate if
\begin{align}
    \mathbb{E}_{y_l \sim \rho^{y}_l}\left[y_l\right] > & \mathbb{E}_{\tilde{y}_{l^{\prime}} \sim \tilde{\rho}^{y}_{l^{\prime}}}\left[\tilde{y}_{l^{\prime}}\right]: l^{\prime} > l,
\end{align}
where $\tilde{\rho}^{y}_{l^{\prime}} = \{y: \mathbf{x} \sim \rho^{\pi}_{l}, \mathbf{z} \sim \pi^{\theta}_{l^{\prime}}, r \sim R\}$.
\end{definition}
Policy degeneracy occurs when the expectation taken over the future discounted reward distribution, formed using actions elicited from a policy at $l$ is better than the expectation over the distribution from from actions elicited from the policy at $l^{\prime}$. With the definitions of buffer degeneracy and policy degeneracy in mind, we propose Theorem \ref{thm:bd-pd}, below.
\begin{theorem}
\label{thm:bd-pd}
\textbf{(Buffer degeneracy $\implies$ policy degeneracy)}
\begin{align}
    \mathbb{E}_{y_l \sim \rho^{y}_l}\left[y_l\right] > \mathbb{E}_{y_{l^{\prime}} \sim \rho^{y}_{l^{\prime}}}\left[y_{l^{\prime}}\right] \implies \mathbb{E}_{y_l \sim \rho^{y}_l}\left[y_l\right] > \mathbb{E}_{\tilde{y}_{l^{\prime}} \sim \tilde{\rho}^{y}_{l^{\prime}}}\left[\tilde{y}_{l^{\prime}}\right]: l^{\prime} > l
\end{align}
\end{theorem}
The proof of Theorem \ref{thm:bd-pd} is in Appendix A. The theorem and proof show the relationship between buffer degeneracy and policy degeneracy and map intuitively onto the actor-critic policy-learning ecosystem. When the buffer fills up with low-reward examples, the policy synthesized from these samples consequently produces actions that have lower rewards. This behavior is attributable to the loss of \eqref{eq:exp-loss}, and consequent gradient updates of \eqref{eq:grad-pi}, which is minimized by learning such a policy. Note that the adopted off-policy approach may either help prevent or help encourage degenerate learning. On the one hand, off-policy learning can help prevent degeneracy during early learning phases. On the other hand, a policy that has learned to produce good rewards may become degenerate if the off-policy component produces enough degenerate action-having experience replay tuples. This issue, however, is mitigated by having a sufficiently large buffer.

\subsubsection{Selective Buffering}

To inhibit buffer degeneracy and thus prevent policy degeneracy, we propose a simple greedy procedure that selectively adds samples to the buffer. In essence, we add the experience tuple observed to have the highest reward over a prescribed number of iterations. This procedure is described by Algorithm \ref{algo:select-buffer}, below.
\begin{algorithm}[!htp]
	\caption{Selective Buffering}
	\label{algo:select-buffer}
	Initialize $r_{best} \gets -\infty$\\
	\uIf{$r_t > r_{best}$}{
	    $best \gets (\bx_{t},\bz_{t},r_t,\bx_{t+1})$\\
	}
	\uIf{$t \mod w = 0$}{
	    Add $best$ to $\pmb{\rho}$\\
	    $r_{best} \gets -\infty$\\
	}
\end{algorithm}
Since Algorithm \ref{algo:select-buffer} screens samples according to their reward, the buffer is prevented from filling up with increasingly worse samples, thereby mitigating the buffer degeneracy issue. Note that, $w$ can be set lower or higher to tailor the procedure to be more or less greedy. We show how Algorithm \ref{algo:select-buffer} fits into an updated MLX policy-learning procedure at the end of this section with all ensuing considerations added in.

\subsection{Data Corruption and Federate Learning Considerations}

The optimized policy $\pi^{\theta^*}$ is synthesized by sampling instances from a dataset. This dataset may not be the same as that used to induce $f$ for a variety of reasons. First, instances may be lost or corrupted over time. Second, some data may not be accessible due to federated learning or other privacy-preserving considerations. This may give rise to two related issues, illustrated by Figure \ref{fig:exp-ex}:
\begin{enumerate}
\item The first, illustrated by Figure \ref{fig:exp-ex-pos} is that there may be unsupported, and therefore unrepresented areas of the probability space, leading to an unreliable approximation of $f$.
\item A second, related artifact, illustrated by Figure \ref{fig:exp-ex-neg}, also leading to an unreliable approximation of $f$, is class imbalance, where substantially fewer samples of one class are available for learning $\pi^{\theta^*}$. The latter may also be attributable to class imbalance within the training data itself. In this case, a reliable policy may be learned for either $c=0$ or $c=1$, but not the other.
\end{enumerate}
\begin{figure}[!htp]
    \centering
    \begin{subfigure}[b]{0.49\textwidth}
        \centering
        \includegraphics[scale=3.0]{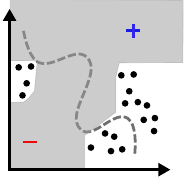}
        \caption{Unsupported areas of the probability space.}
        \label{fig:exp-ex-pos}
    \end{subfigure}
    \begin{subfigure}[b]{0.49\textwidth}
        \centering
        \includegraphics[scale=3.0]{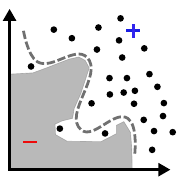}
        \caption{Class imbalance.}
        \label{fig:exp-ex-neg}
    \end{subfigure}
    \caption{Two examples of unsupported probability space regions, shaded in \textcolor{gray}{gray}. The decision boundary, which we are trying to approximate, is represented by the dark gray dotted line and the available policy-learning training samples are represented by the black dots.}
    \label{fig:exp-ex}
\end{figure}

The ``unsupported probability space'' issue is naturally mitigated as a consequence of our conceptualization of the problem and adopted off-policy RL methods. Off-policy RL explores the environment in question (see \eqref{eq:off-policy}) -- i.e.,~the probability space -- when searching for the explanatory point of each instance during policy synthesis. Thus, if a sufficiently large number of policy training iterations are executed, the probability space is naturally explored, mitigating this concern.

To alleviate the class imbalance issue, which amounts to under-exploration of one side of the probability space wrt.~the decision boundary in question, we can apply any well-known class-balancing technique to the policy-learning dataset. Such methods include the various majority class under-sampling \citep{liu2008exploratory,tsai2019under,yen2009cluster} and minority class oversampling \citep{chawla2002smote,barua2012mwmote} techniques. Since we wish to keep as many organic instances as possible, we focus on oversampling the minority class, adopting the well-known synthetic minority oversampling technique (SMOTE) \cite{chawla2002smote}.

\subsection{Human-in-the-loop Feature Trustworthiness \label{sec:hitl}}

An intrinsic component of MLX is the HITL element \citep{lipton2018mythos}. It is this element that necessitates elicitation of an explanation in the first place. In this paper we consider a HITL decision-making scenario, where an individual $k$ interacts with $f(\bx)$ and $\Omega(\bx,\omega)$ by inputting $\bx$ and obtaining a probability estimate $\hat{\mathbb{p}}$ and a discrete prediction $\hat{c}$ of the outcome. The decider $k$ then also obtains an explanation $\bz$ of this prediction from $\pi^{\theta^*}$. Subsequently, $k$ considers $\bz$ in order to decide whether the prediction was made \textit{for the right reasons} -- i.e.,~the prediction is trustworthy/reliable  \citep{ribeiro2016should}. Recall that $\bz = \bx - \bx^{\prime}$, where $\bx^\prime$ is the explanatory point nearest $\bx$ on the class-separating decision boundary. Therefore, $\bz$ tells $k$ which features (and by how much) ``pushed'' the instance towards the classification that was produced and, if those features seem reasonable, $k$ can choose to trust and subsequently act on the prediction (purchase the stock, treat the patient, etc.). To better formalize this notion, let $\bz_k \in \{0,1\}^{p}$ denote an acceptability vector for decider $k$, where
\begin{align}
    z_k^{(j)} =
    \begin{cases}
        1 & \text{if }k \text{ trusts feature }j \\
        0 & \text{Otherwise}
    \end{cases}.
\end{align}
Further, denote $\tilde{\bz}$ to be an indicator vector of $\bz$ such that
\begin{align}
    \tilde{z}^{(j)} =
    \begin{cases}
        1 & \text{if }z^{(j)} \neq 0\\
        0 & \text{Otherwise}
    \end{cases}.
\end{align}
In other words, $\tilde{\bz}$ indicates the non-zero entries in $\bz$ that were used to explain $\bx$. Using $\tilde{\bz}$ and $\bz_k$ we can compute an explanation-decider disagreement score as
\begin{align}
    \label{eq:agree-score}
    \zeta_k^{z}= \sum_{j=1}^{p}\mathbb{1}_{ (\tilde{z}^{(j)} - z_k^{(j)}) > 0}.
\end{align}
Note that $(\tilde{z}^{(j)} - z_k^{(j)}) > 0$ increases the score when a non-trusted feature is used to make an explanation, but not when a trusted feature is \textit{not} used as part of the explanation. This is intuitive since there may be times when only a subset of trustworthy features are used to provide an explanation. The explanation-decider disagreement score is used to motivate the UEP assessment measure, defined in the Experiments section. Finally, note that this measure can be considered a measure of explanation fidelity, similar to those proposed in \cite{yeh2021objective}.

\subsubsection{Individual-specific Explanatory Policies via 0-Distrust Projection}

Ideally, the learned explanatory policy $\pi^{\theta^*}$ will be sensitive to an individual $k$'s preferred explanatory features, i.e.,~ the non-zero entries of $\bz_k$. To learn a policy that is sensitive and specific to $k$ we propose the use of what we term \textit{0-Distrust Projection}. To present this idea, we define an instance-decider action feasibility domain $\mathcal{Z}^{(k)}_x$ as
\begin{align}
    \label{eq:0-distrust}
    \mathcal{Z}_x^{(k,j)} = 
        \begin{cases}
        \mathcal{Z}_x^{(j)} & \text{if }z_k^{(j)} = 1\\
        \{0\} & \text{Otherwise}
    \end{cases} j=1,\dots,p.
\end{align}
We subsequently update \eqref{eq:proj} to project onto instance-decider domain $\mathcal{Z}^{(k)}_x$ if we wish to synthesize a policy that is sensitive to $k$'s preferences, thus ensuring $\zeta_k^{z}=0$.

With selective buffering, federated learning/data corruption, and 0-distrust (i.e.,~HITL) considerations in mind, we provide an updated MLX policy-learning procedure in Algorithm \ref{algo:hitl-pol-learn}, below.
\begin{algorithm}[!htp]
	\caption{HEX MLX policy learning procedure}
	\label{algo:hitl-pol-learn}
    Initialize buffer $\pmb{\rho} \gets \{\}$ and $r_{best} \gets -\infty$\\
    Execute a cold start procedure to provide initial samples $\pmb{\rho}$.\\
	\For{$e=1,\dots,E$}{
        $\bx \sim \mathcal{D}$ (Sample an instance from policy-learning dataset $\mathcal{D}$)\\
		\For{$t=1,\dots,T$}{
		    \uIf{$0$-distrust}{
		        $\bz_{t} \gets Proj_{\mathcal{Z}_x^{(k)}}\left(\check{\pi}\left(\mathbf{x}_t \vert \theta \right)\right)$ (Using \eqref{eq:off-policy} and \eqref{eq:proj} with \eqref{eq:0-distrust}).\\
		    }
		    \uElse{
		         $\bz_{t} \gets Proj_{\mathcal{Z}_x}\left(\check{\pi}\left(\mathbf{x}_t \vert \theta \right)\right)$ (Using \eqref{eq:off-policy} and \eqref{eq:proj}).\\
		    }
		    Estimate $r_t$ according to \eqref{eq:exp-reward}.\\
		    $\bx_{t+1} \gets \bx_{t} + \bz_{t}$ (Transition to the next state.)\\
		    \uIf{$r_t > r_{best}$}{
	            $best \gets (\bx_{t},\bz_{t},r_t,\bx_{t+1})$\\
	        }
	        \uIf{$t \mod w = 0$}{
	            $\pmb{\rho}_{l/w \mod \xi} \gets best$\\
	            $r_{best} \gets -\infty$\\
	        }
		    \uIf{$m = DDPG$}{
                Sample $\pmb{\rho}$ and compute the loss according to \eqref{eq:exp-loss}.\\
                Backpropagate the loss to update $\theta^Q$ and use \eqref{eq:grad-pi} to update $\theta^\pi$.\\
                Apply soft updates according to \eqref{eq:soft-update}.
            }
            \uElseIf{$m = TD3$ and $t \mod 2 = 0$}{
                Sample $\pmb{\rho}$ and compute the loss according to \eqref{eq:exp-loss}.\\
                Backpropagate the loss to update $\theta^Q$ and use \eqref{eq:grad-pi} to update $\theta^\pi$.\\
                Apply soft updates according to \eqref{eq:soft-update}.
            }
		}
    }
	\KwResult{$\pi^{\theta^*}$}
\end{algorithm}

\section{Experiments}
To assess our proposed method we adopt five real-world datasets -- Bank, MIMIC, Movie, News, and Student -- spanning a variety of decision-focused domains. Descriptions of these datasets and summary statistics are found in Appendix B. We learn five well-known machine learning algorithms, logistic regression, SVM, neural networks, decision trees, and random forest on each dataset. The procedure for training each of these models on each of our datasets is found in Appendix C, along with predictive performance assessments of each model on each dataset.

Our first experiments do not consider a HITL decider (i.e.,~are decider-free). To begin assessing explainability method performance, we first examine the policy learning curves of each of our proposed HEX methods on each dataset and model against out-of-the-box DDPG and TD3 methods, which we henceforth refer to only as DDPG and TD3. Policies are learned by setting $E=1000$, corresponding to 1000 sampled training instances, and $T=300$. Inner training iterations terminate early when an explanatory point is discovered. Each method learns 10 policies (10 trials) for each dataset and method to ensure the results obtained are robust. Episode-wise averages are taken over the 10 trials, with a rolling window average subsequently taken over these averages (window size $= 40$) and plotted to produce learning curves. Subsequently, we examine the performance of these methods at producing explanations on held-out sets of data, each consisting of $100$ instances, comparing to two well-established model-agnostic explanation methods, LIME \citep{ribeiro2016should} and Growing Spheres \citep{laugel2018comparison}, which we display as ``Grow'' when reporting results. The learning curves are shown in terms of average explanatory reward, while the test set evaluations are shown in terms of average explanatory point decision boundary deviance (DBD), defined as follows:
\begin{align}
DBD =& \hspace{.1cm} \vert  f(\bx^{\prime}) - \omega \vert.
\end{align}
$\omega=0.50$ for all ML models for all datasets.

In our second set of experiments, we repeat the above for a scenario that explicitly considers a human-in-the-loop decider who is sensitive to some subset of features. To simulate this scenario we randomly select 10, 50, and 90\% of features and deem those selected as being explanation-unacceptable to the simulated decider (i.e.,~$90,50,10$\% are acceptable). We refer to these percentages as the unacceptable percentage, or UAP for short. We vary this random selection across each of the 10 trials, holding each selected feature set consistent for each MLX method being evaluated. Subsequently, we examine the average DBD on a held-out set of test data, comparing our HEX methods against the non-HEX DDPG, TD3, LIME, and Grow benchmark methods. Furthermore, we show the percentage of undesirable features used in explaining predictions, averaging across the 10 trials. We refer to this percentage as the unacceptable explanatory percentage, or UEP for short. This measure is motivated by our defined explanation-decider disagreement score disclosed in \eqref{eq:agree-score}. The UEP of an explanatory point is calculated as:
\begin{align}
    UEP = & \hspace{.1cm} \frac{\sum_{j=1}^{p} \mathbb{1}_{x^{\prime}_j > 0 \hspace{0.1cm} \cap \hspace{0.1cm} \mathcal{Z}_x^{(j)} = \{0\}}}{p \times UAP}
\end{align}

\section{Results}
We present the results of our experiments below, stratifying their presentation into ``decider-free' and ``HITL decider'' scenarios.

\subsection{Decider-free Scenario}
This subsection presents the results of our decider-free evaluations, where we do not consider or simulate a HITL decider. This scenario is intended to establish our proposed methods as being comparable or superior to the state-of-the-art, agnostic of the specific HITL problem setting considered by this work. We begin by examining the learning curves of the four DRL methods: DDPG, TD3, HEX-DDPG, and HEX-TD3. Subsequenntly, we examine the performance of all methods on held-out test data.

\subsubsection{Learning Curves}

The learning curves of the four DRL methods are depicted in the table of Figures, below (Table \ref{tab:lc-free}). Generally speaking, all DRL methods were able to learn explanatory point-producing policies on all datasets for all ML methods considered. That said, several trends are apparent.
\begin{table}[!htp]
        \centering
        \begin{tabular}{cM{24mm}M{24mm}M{24mm}M{24mm}M{24mm}}
           \toprule
             & Bank & MIMIC & Movie & News & Student\\
            \midrule
            DT & \includegraphics[scale=0.12]{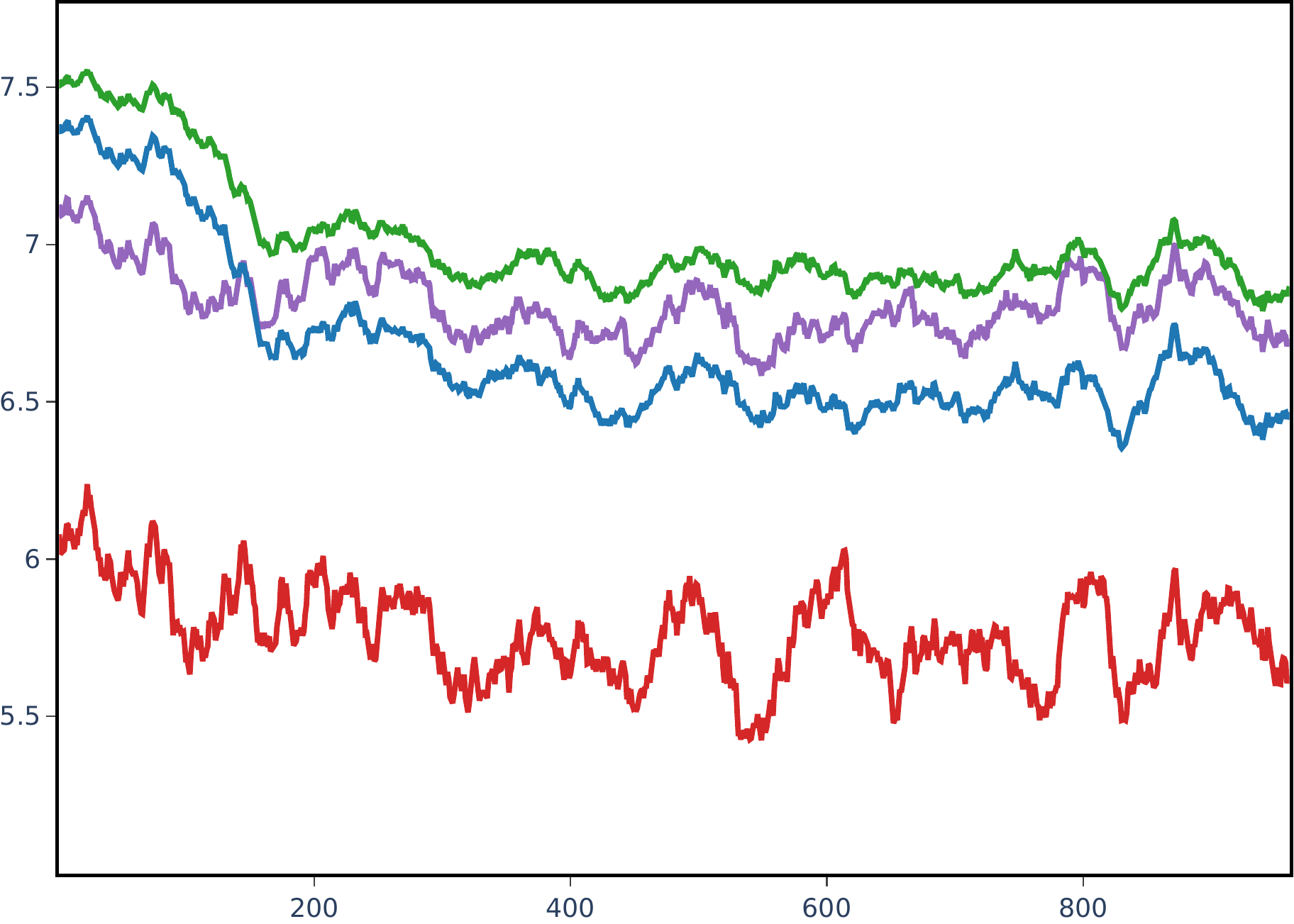} &  \includegraphics[scale=0.12]{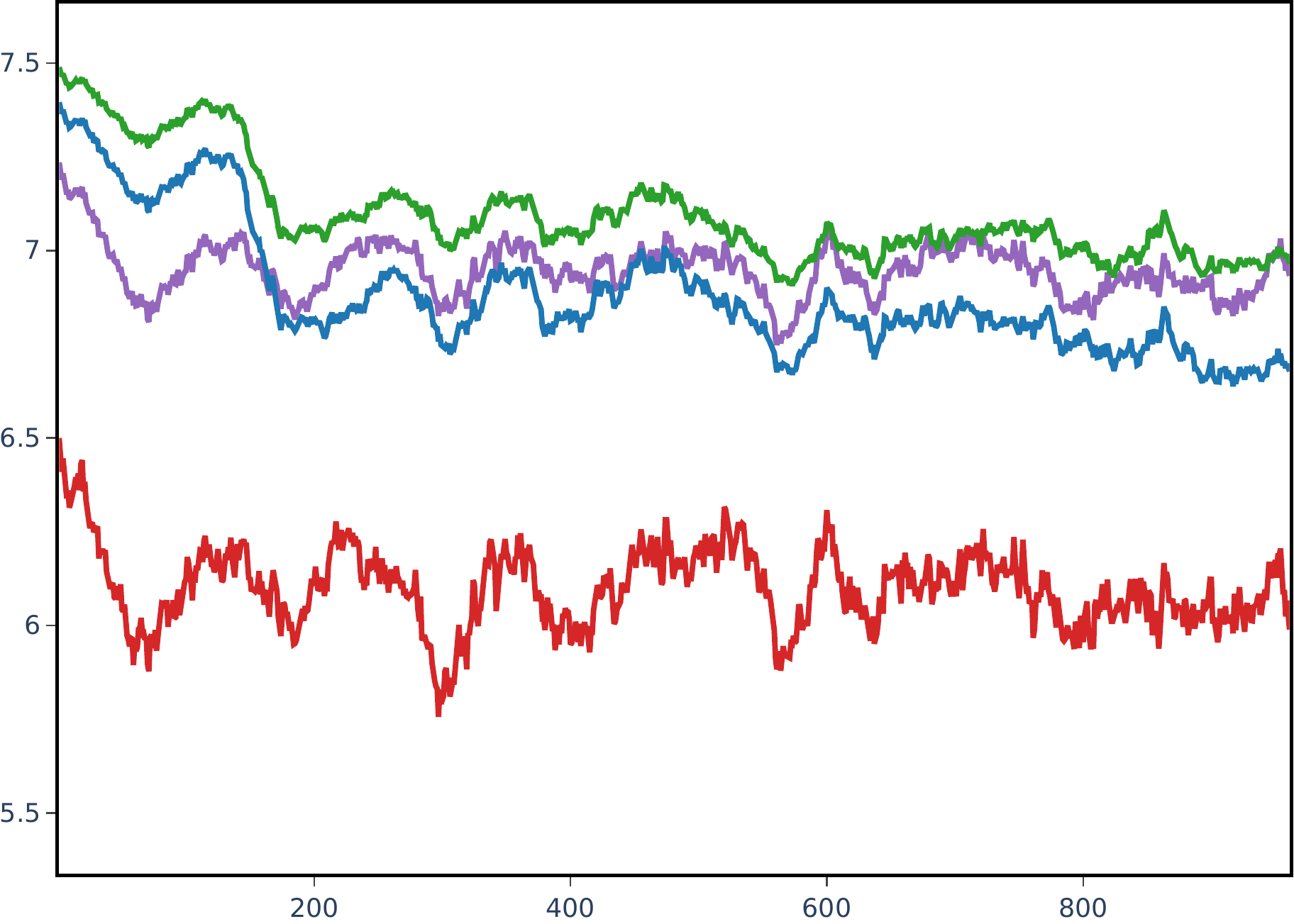} & \includegraphics[scale=0.12]{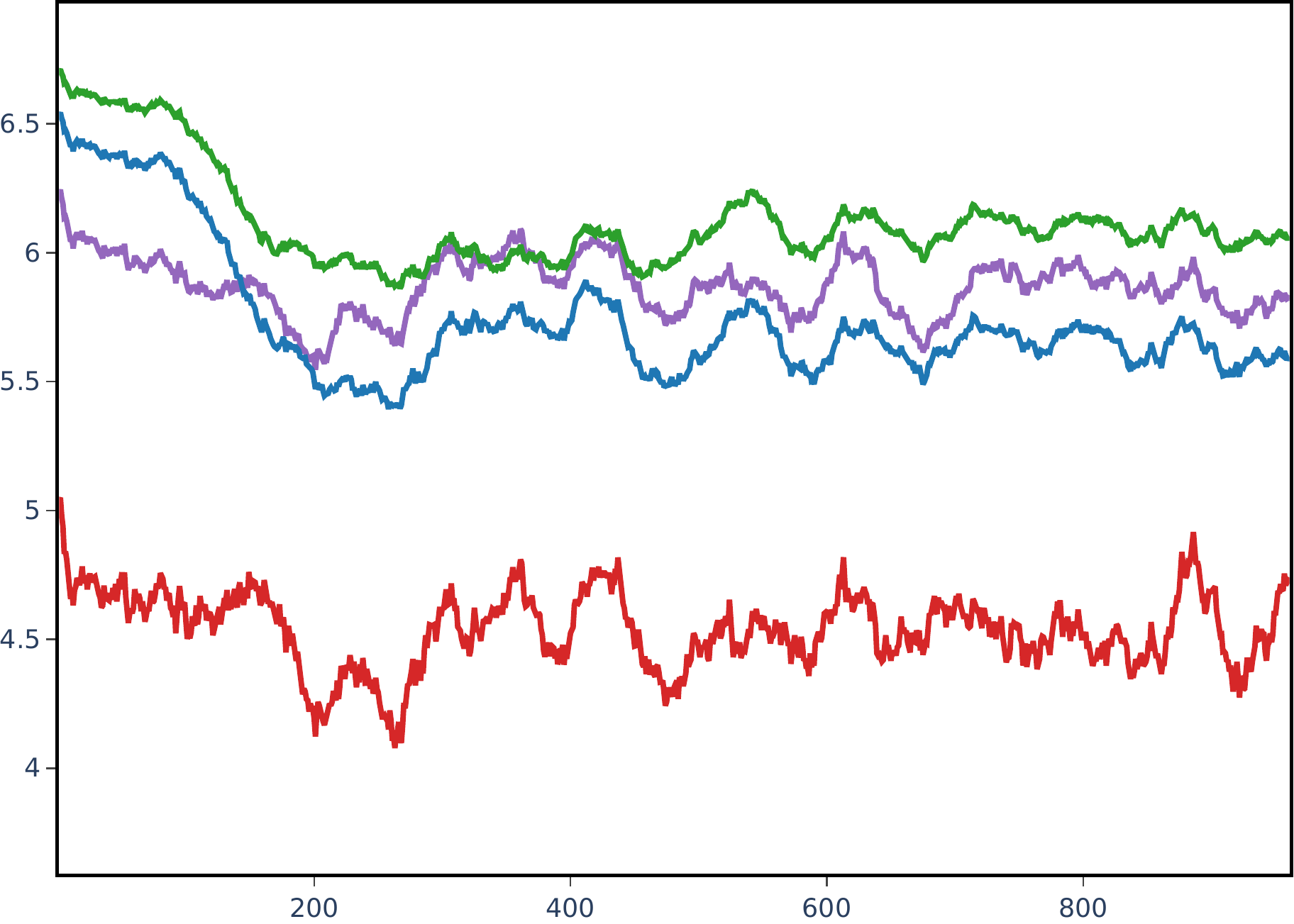} & \includegraphics[scale=0.12]{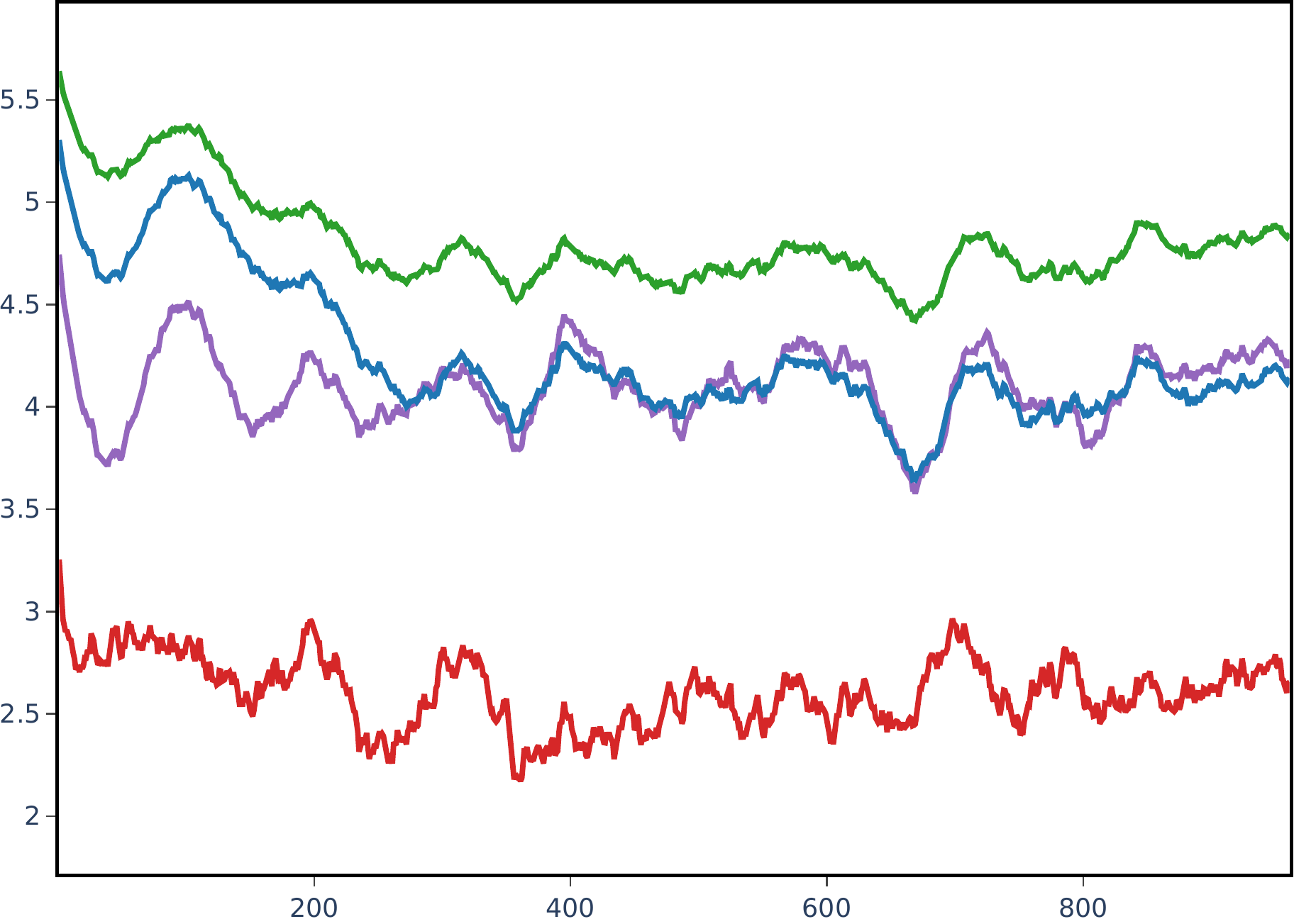} & \includegraphics[scale=0.12]{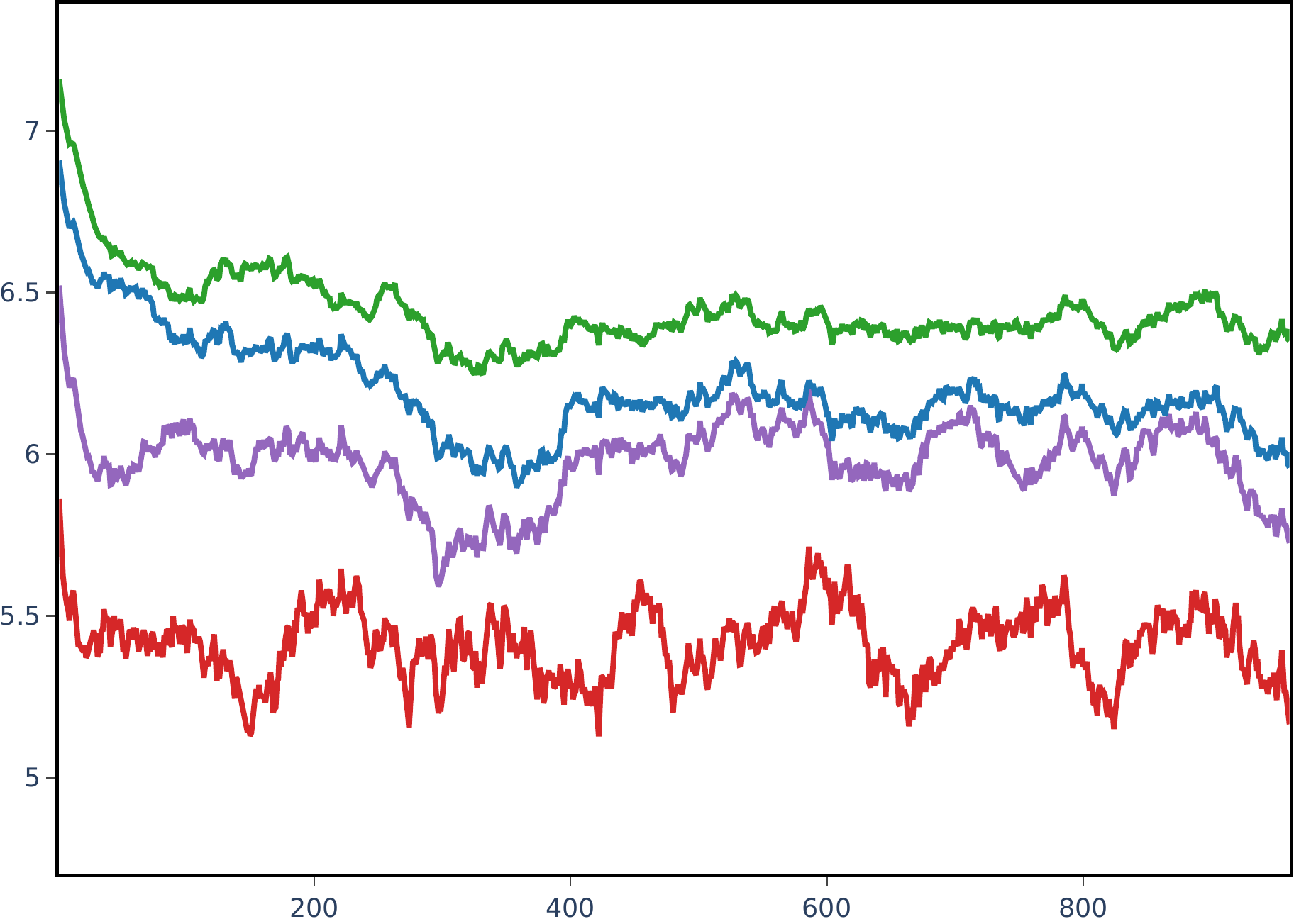}\\
            LR & \includegraphics[scale=0.12]{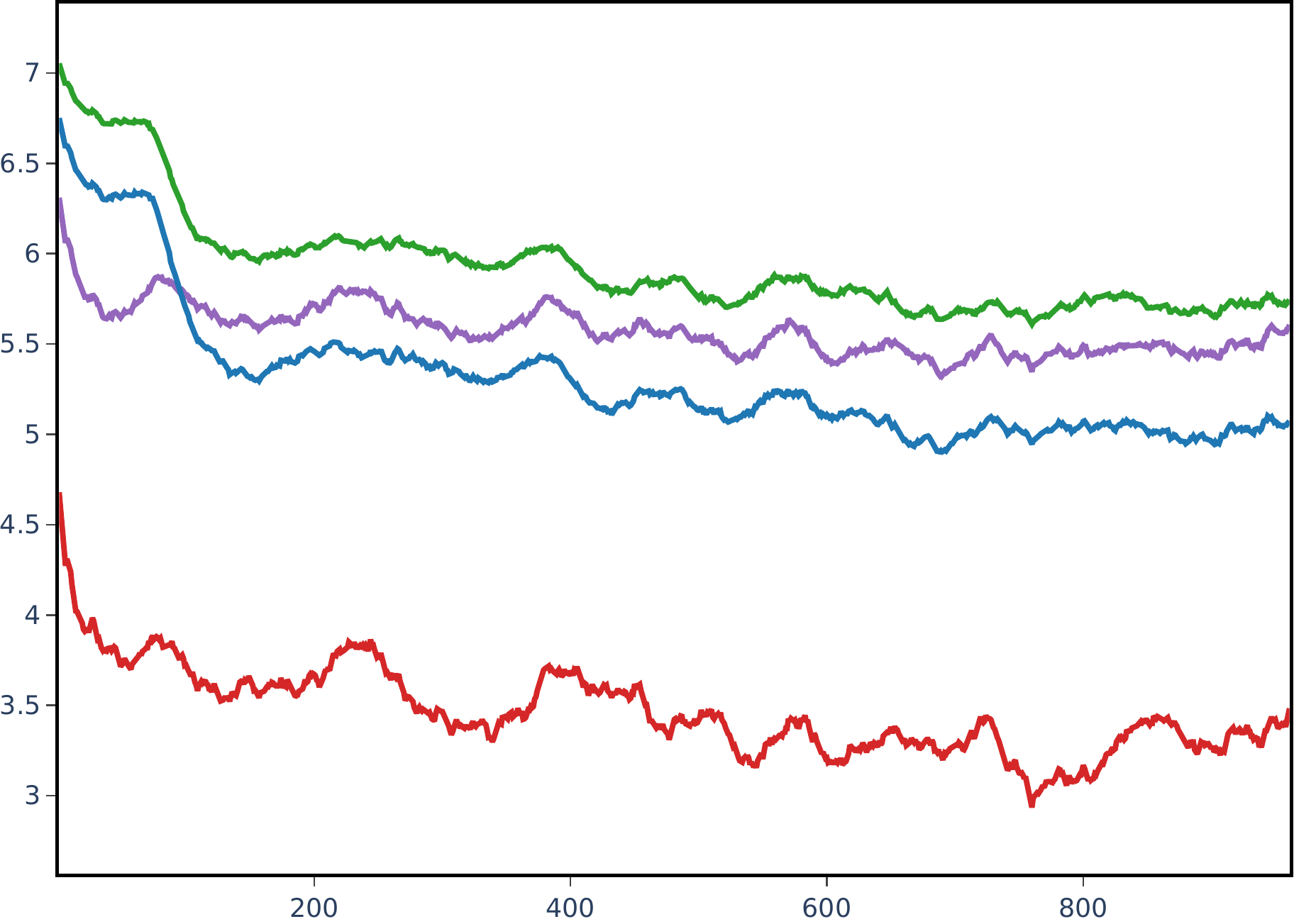}  &  \includegraphics[scale=0.12]{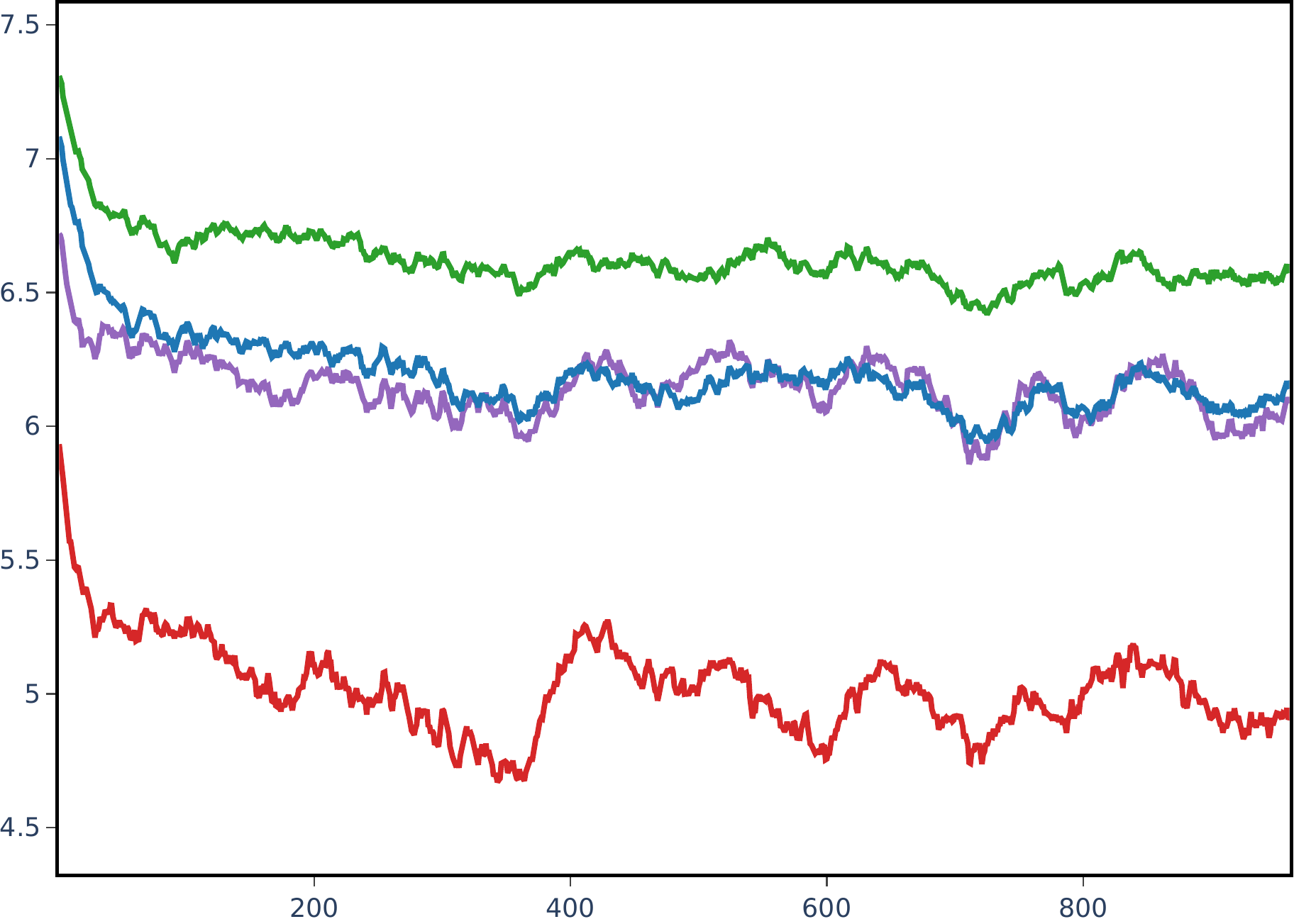} & \includegraphics[scale=0.12]{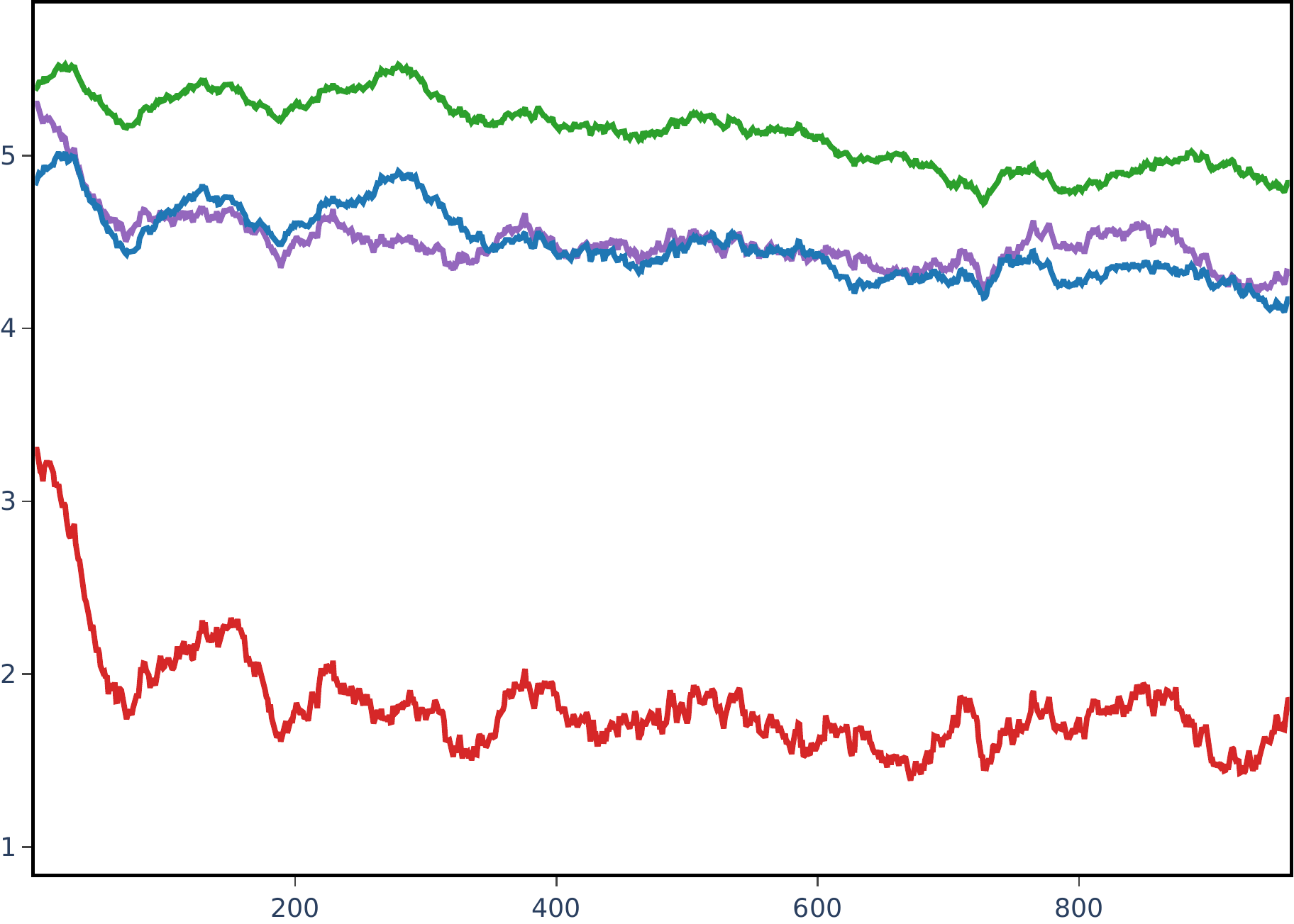} & \includegraphics[scale=0.12]{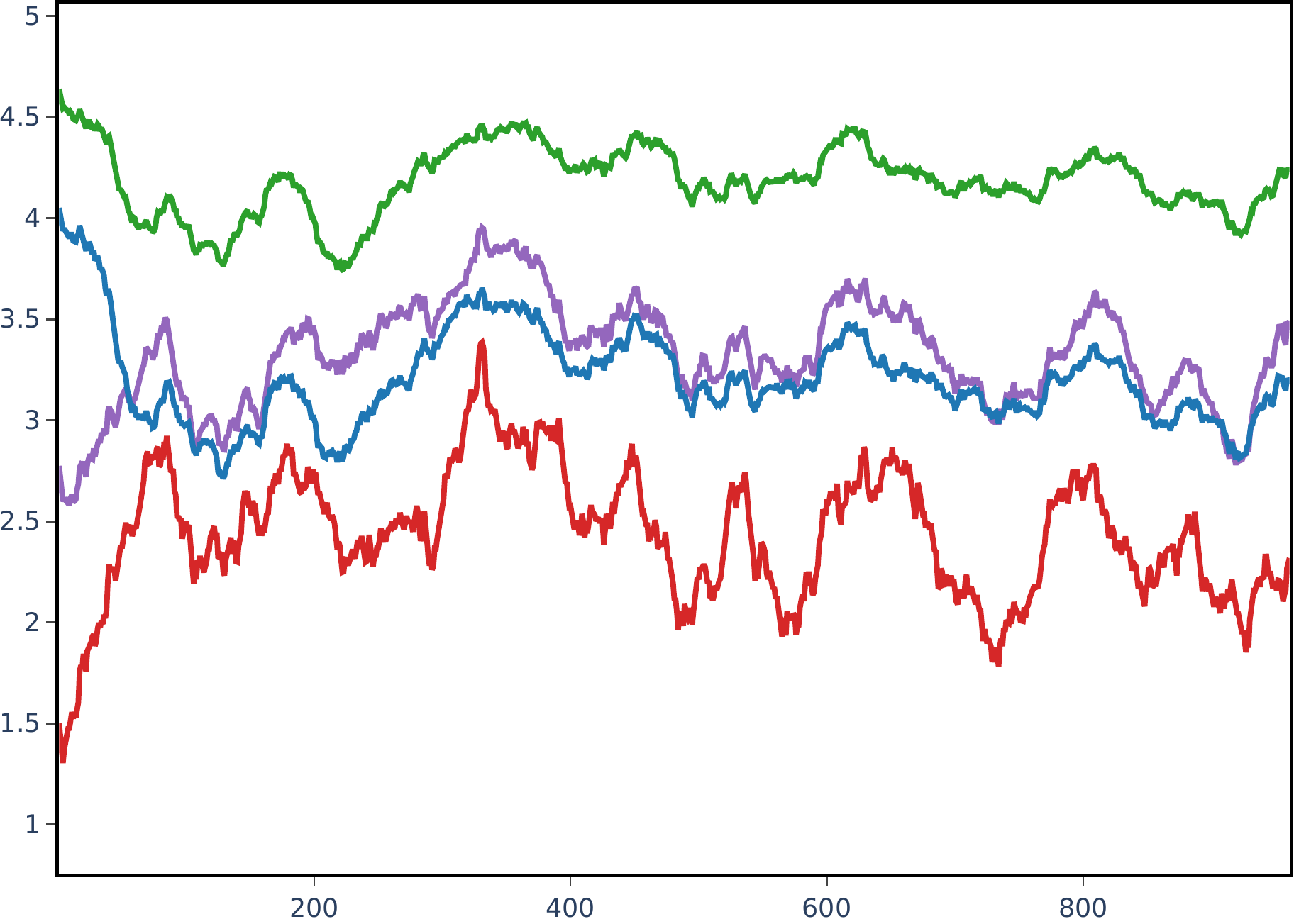} & \includegraphics[scale=0.12]{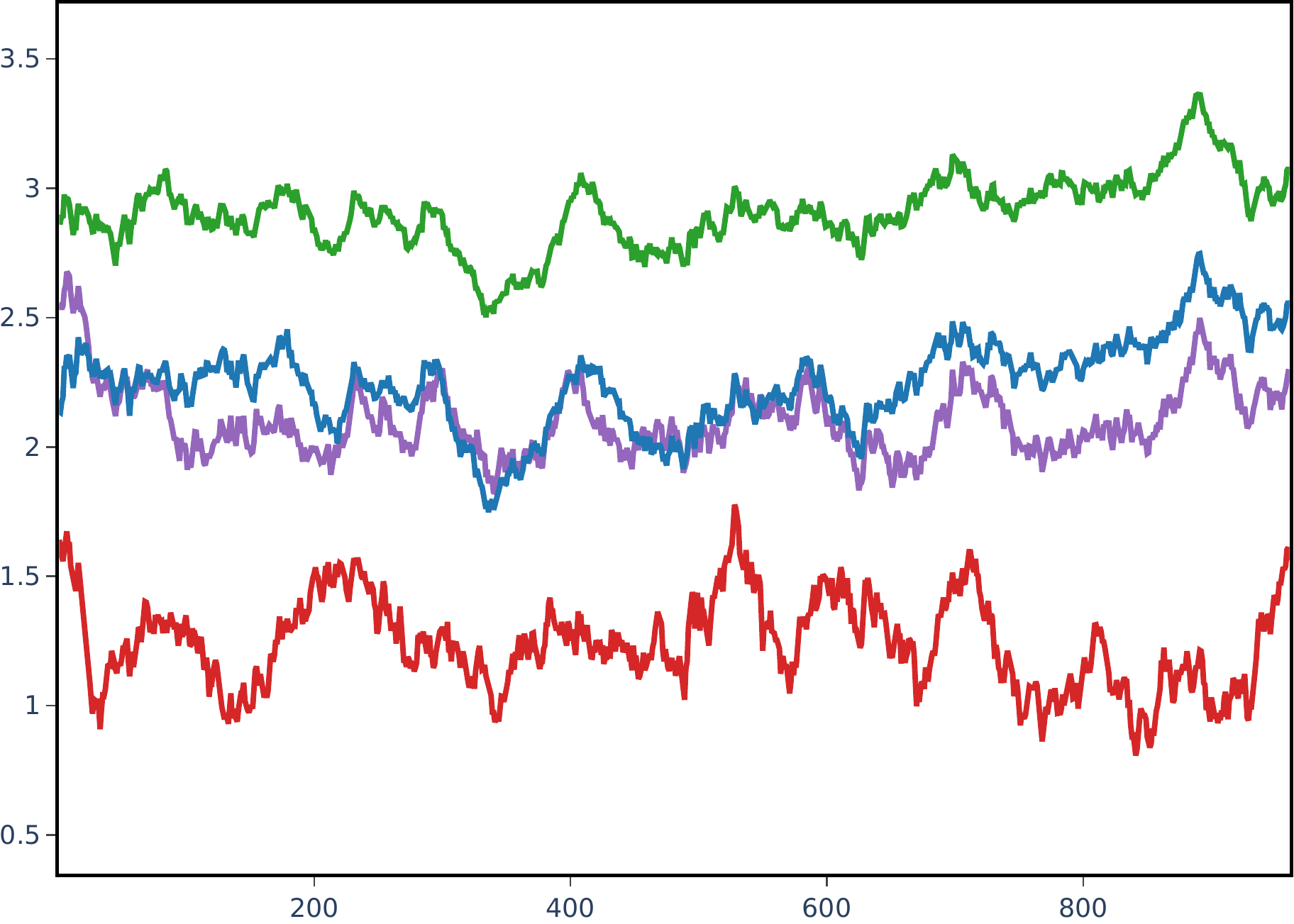} \\
            NN & \includegraphics[scale=0.12]{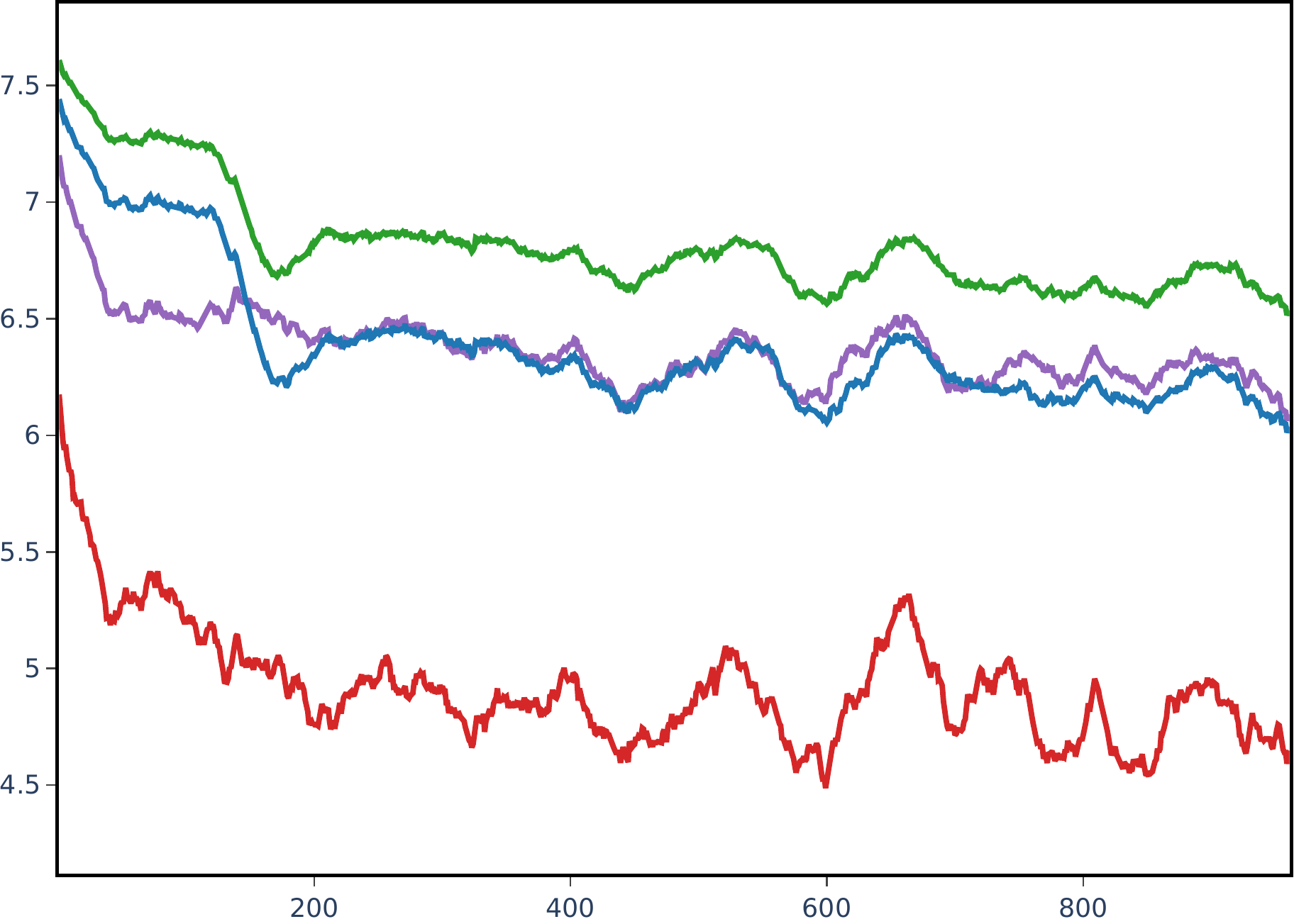}  &  \includegraphics[scale=0.12]{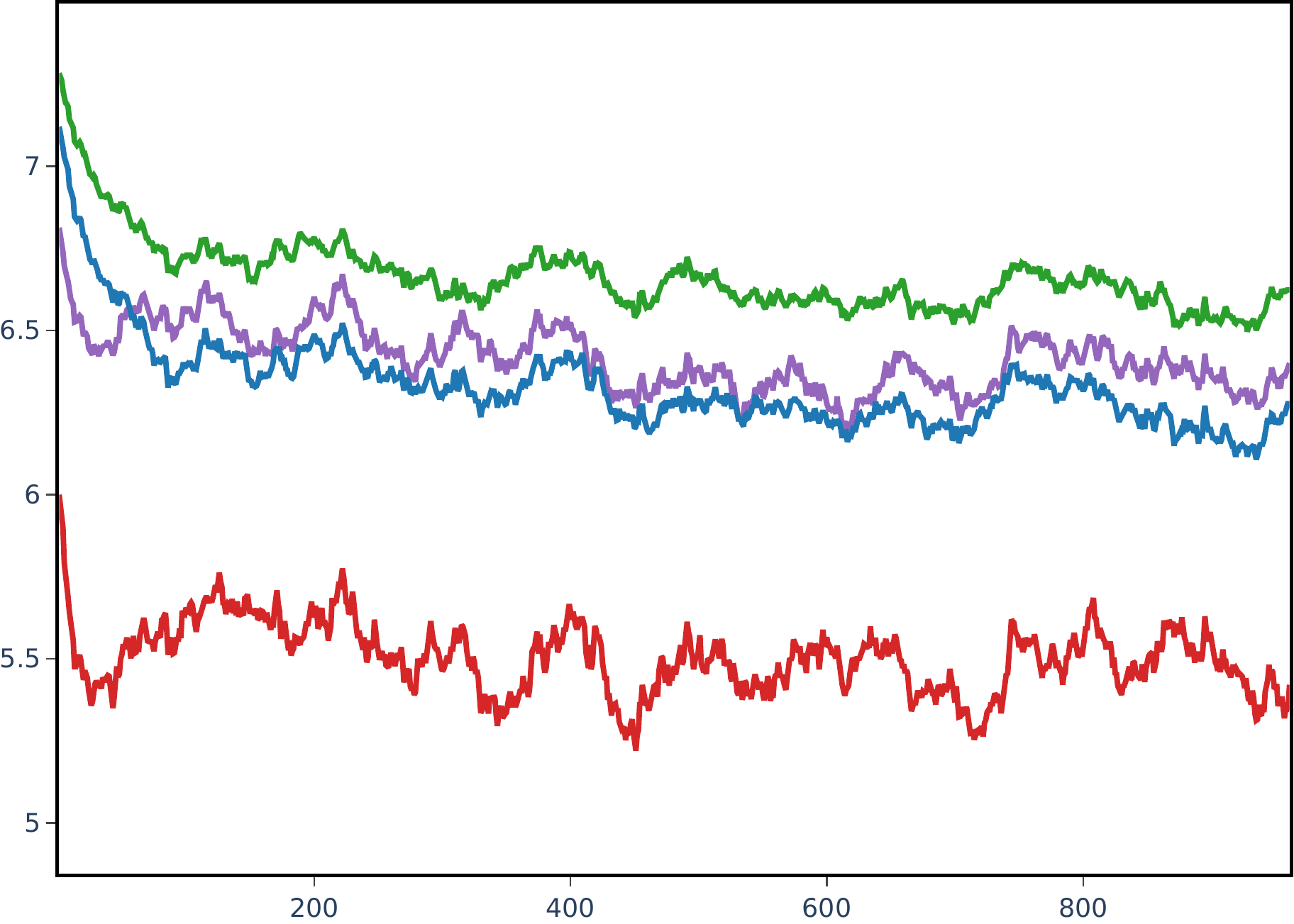} & \includegraphics[scale=0.12]{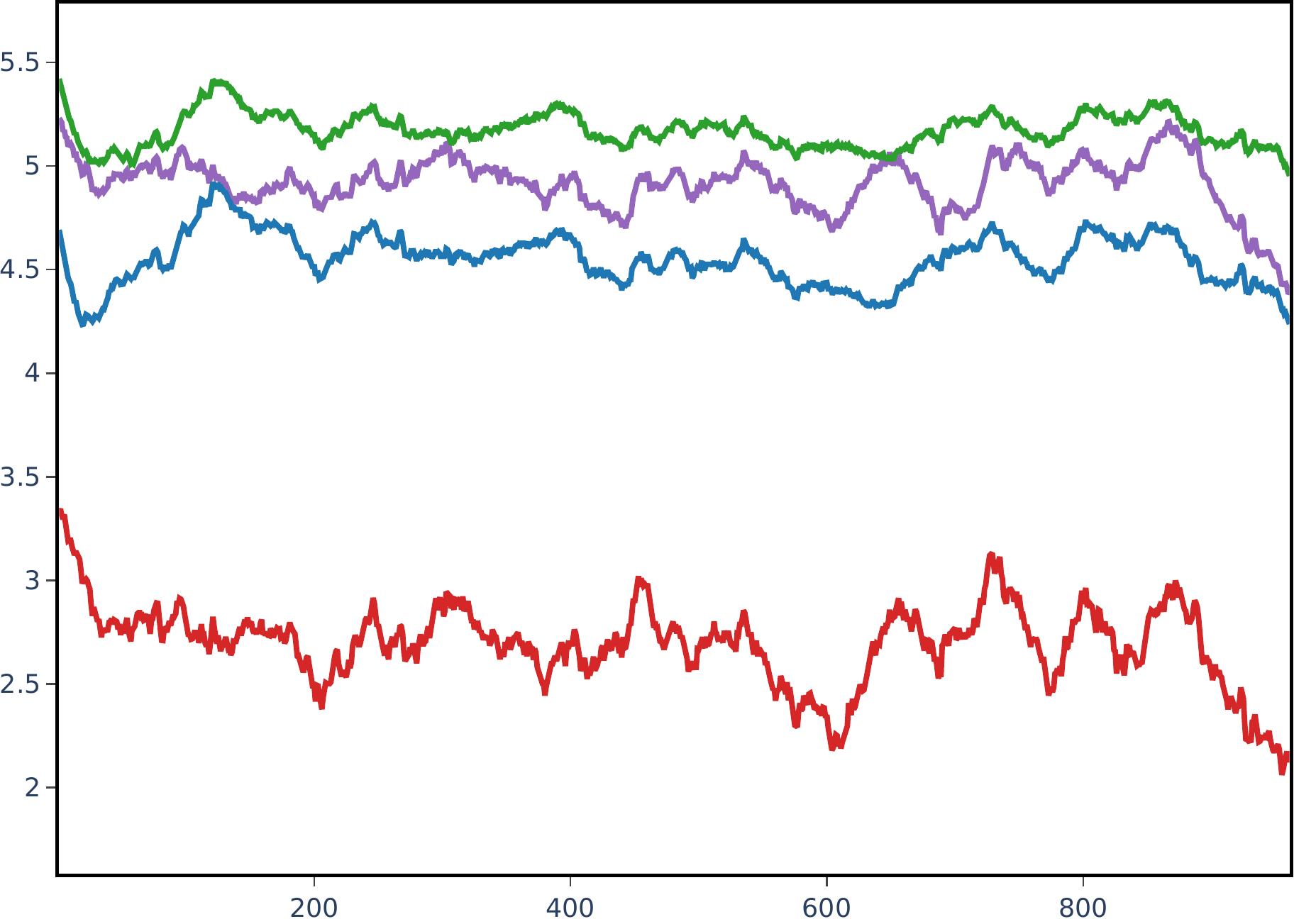} & \includegraphics[scale=0.12]{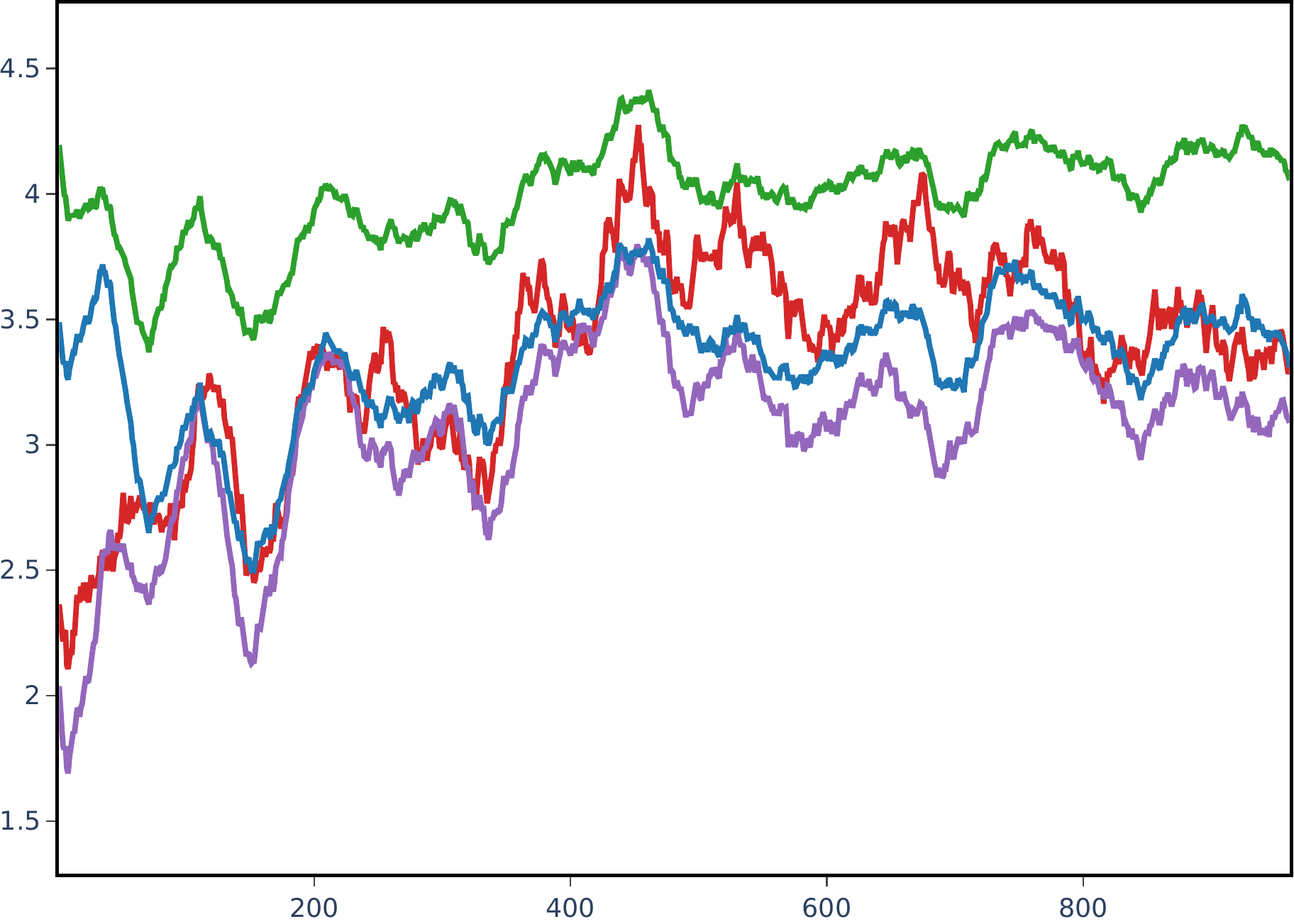} & \includegraphics[scale=0.12]{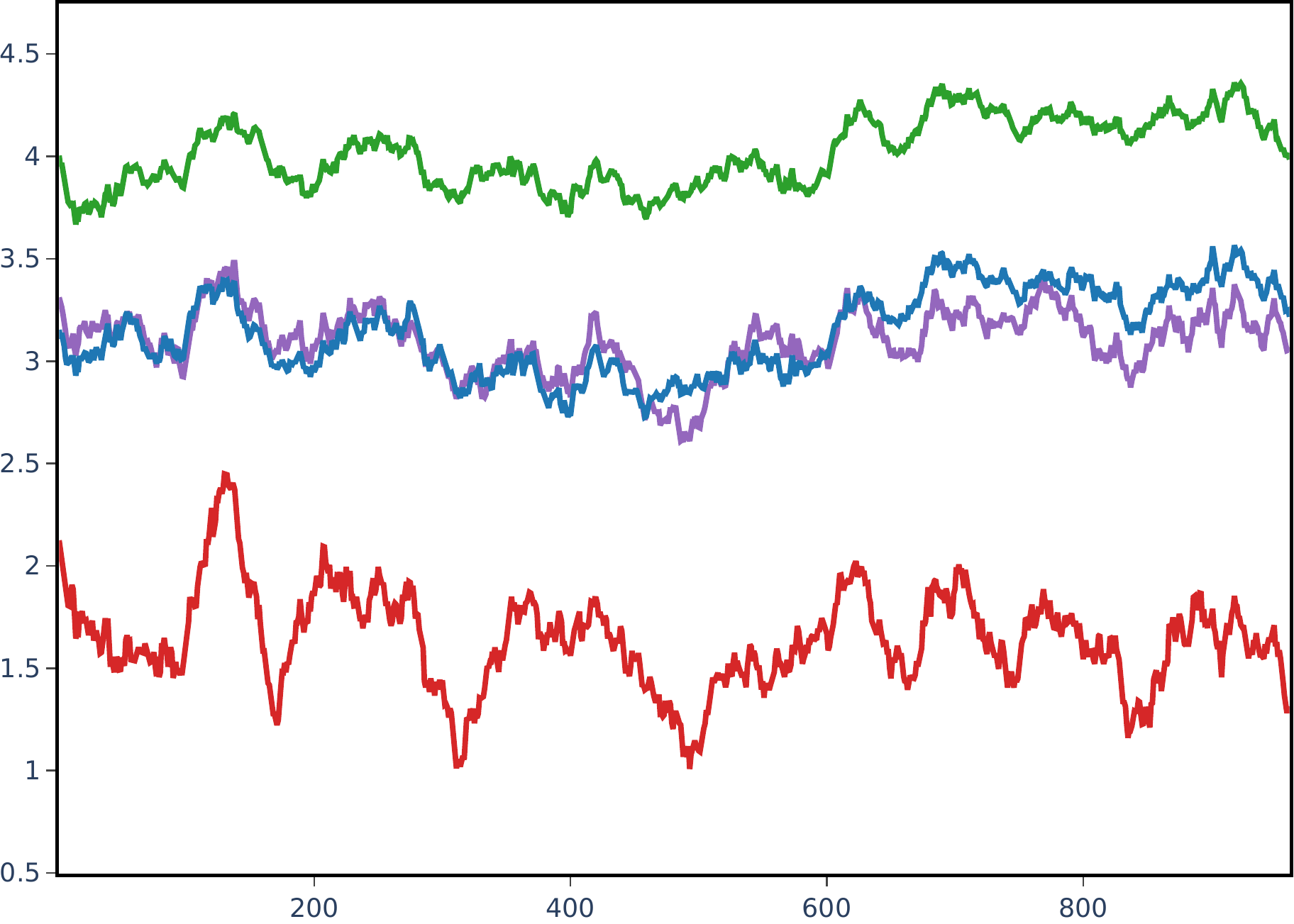} \\
            RF & \includegraphics[scale=0.12]{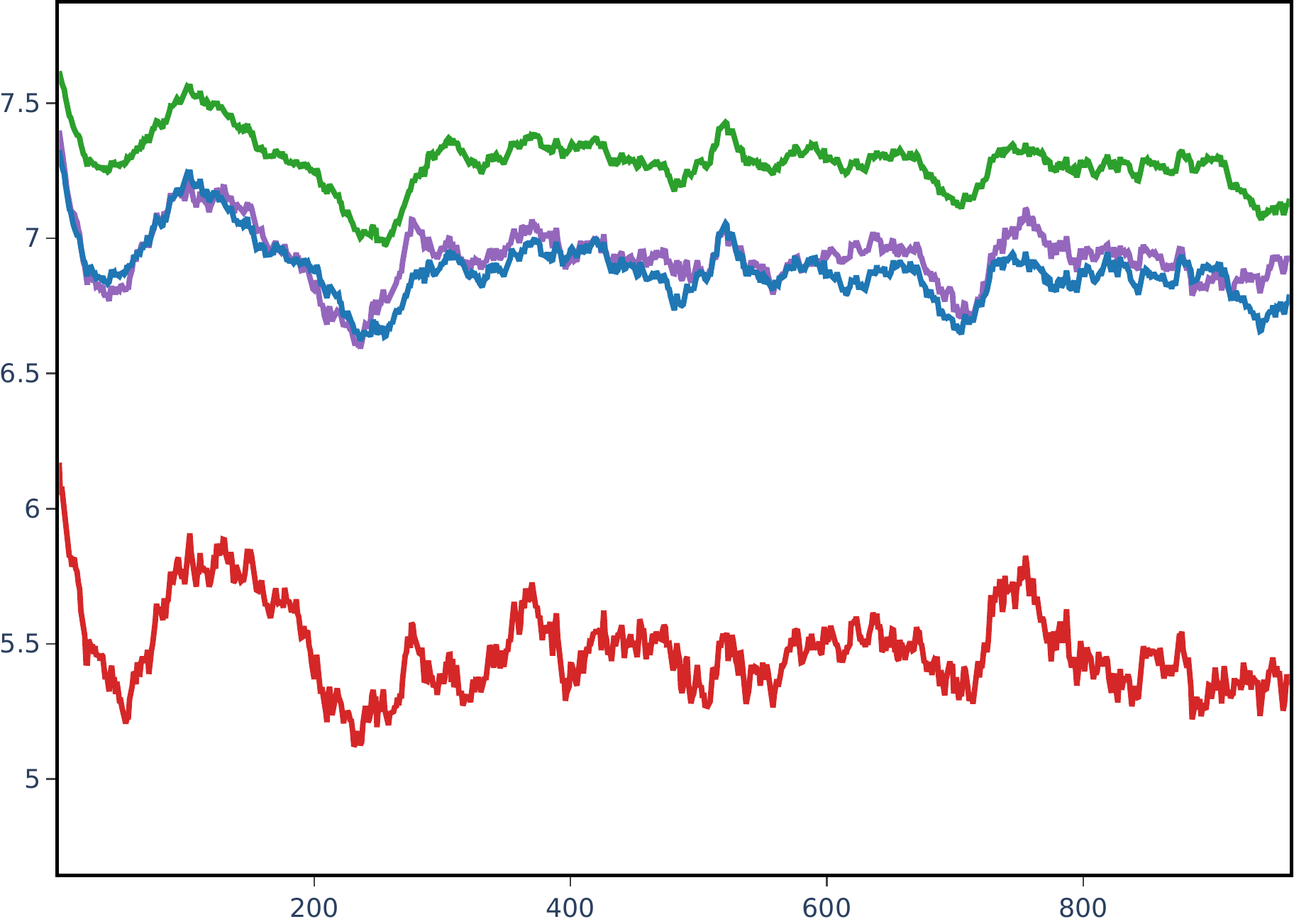}  &  \includegraphics[scale=0.12]{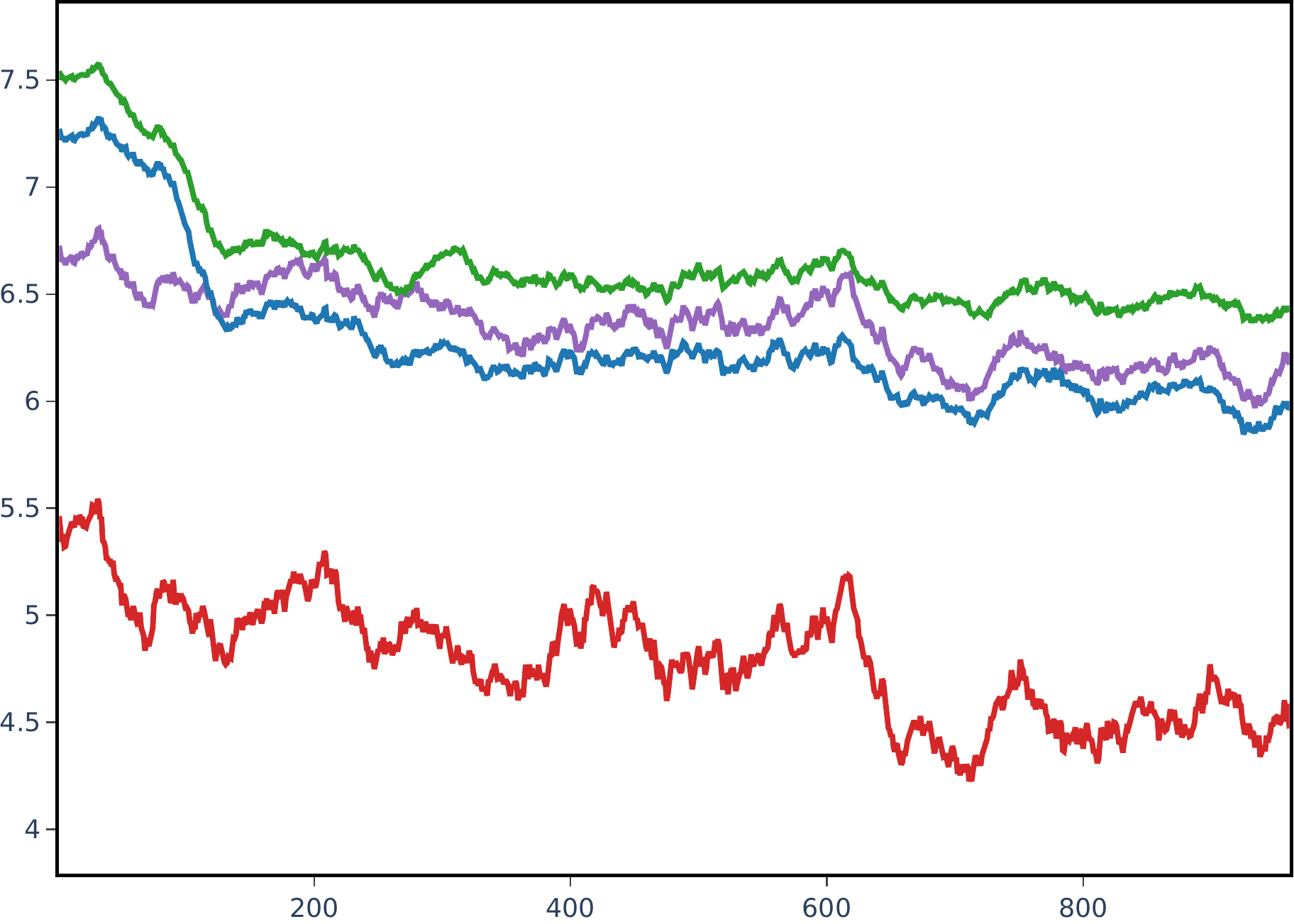} & \includegraphics[scale=0.12]{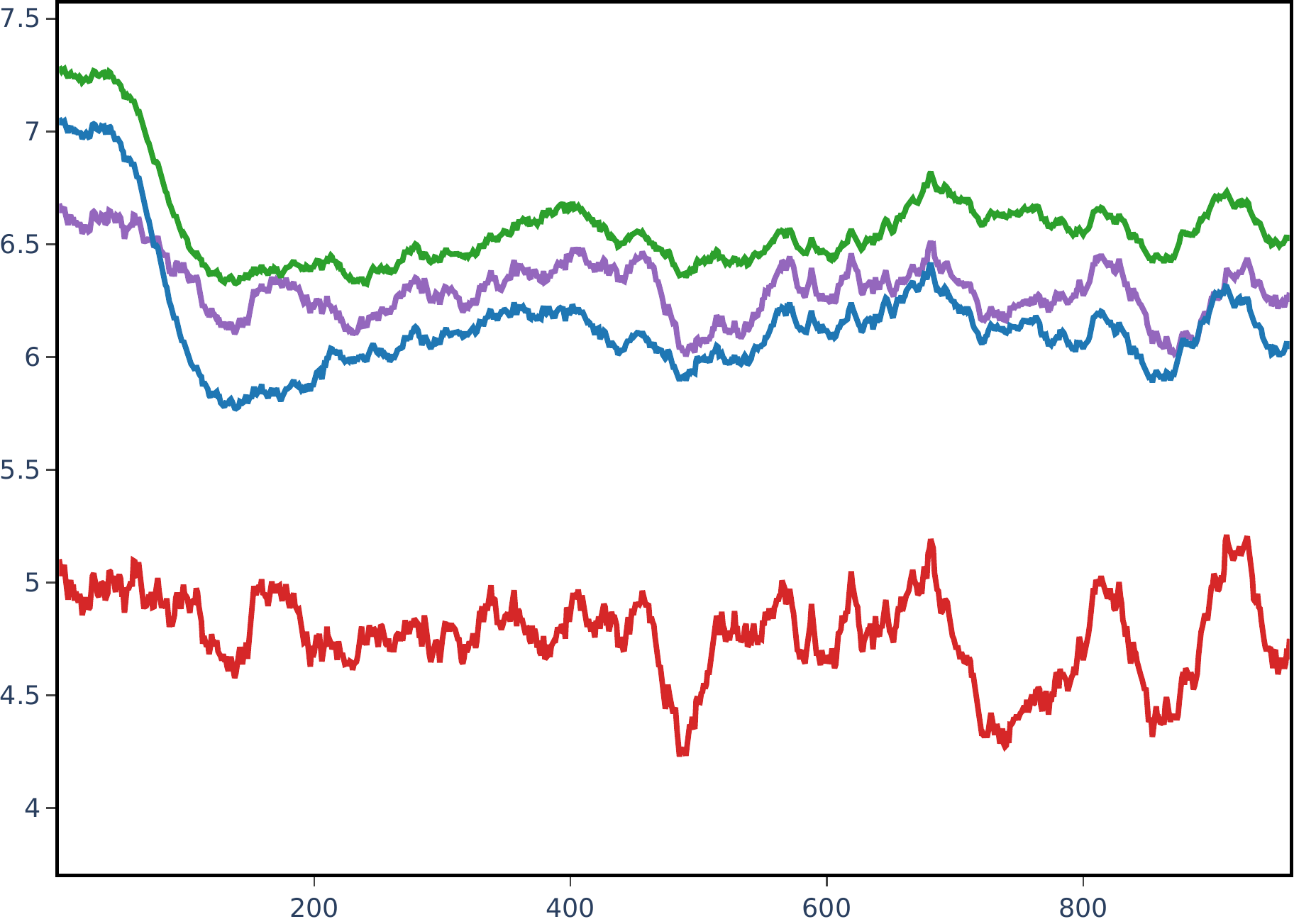} & \includegraphics[scale=0.12]{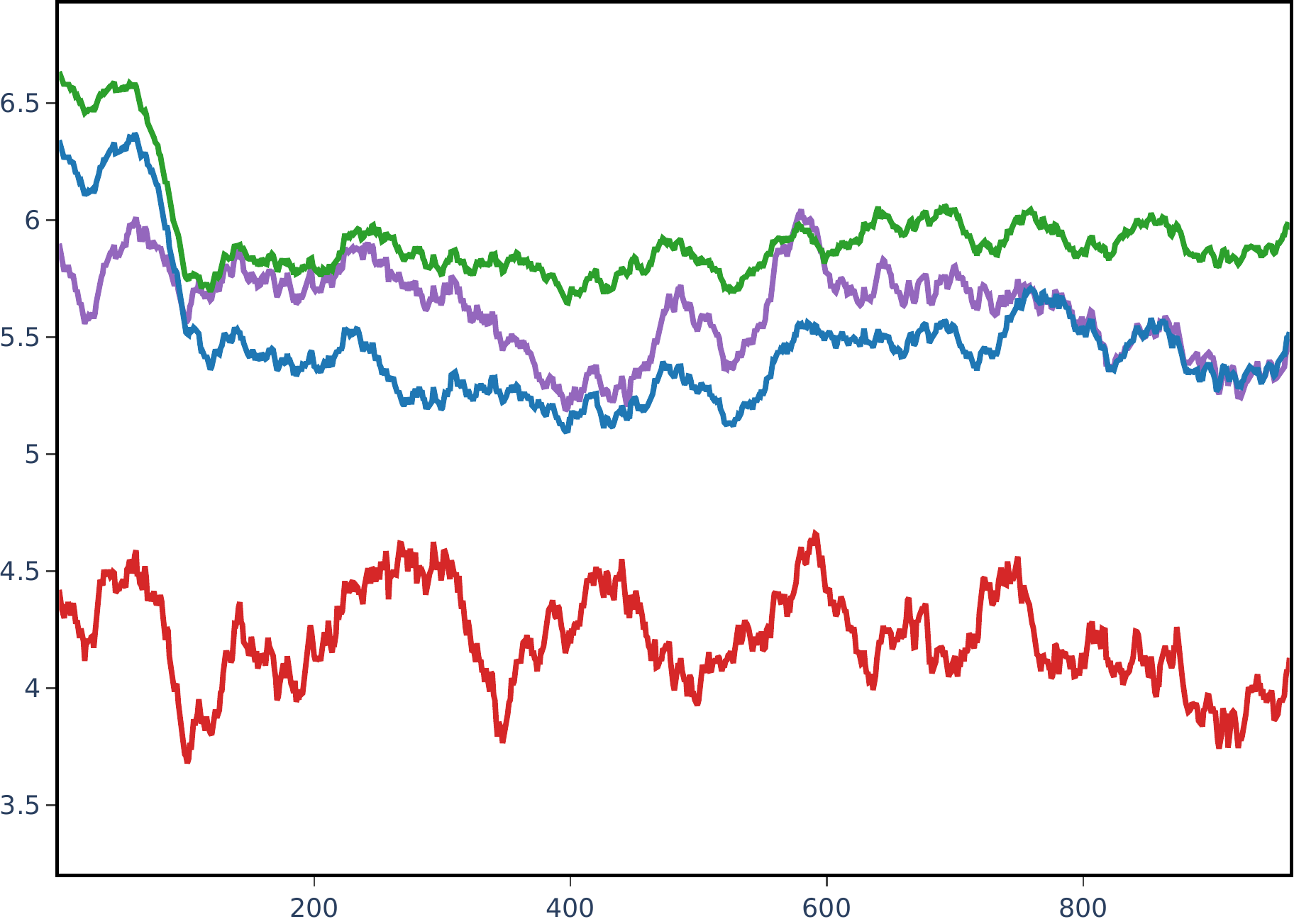} & \includegraphics[scale=0.12]{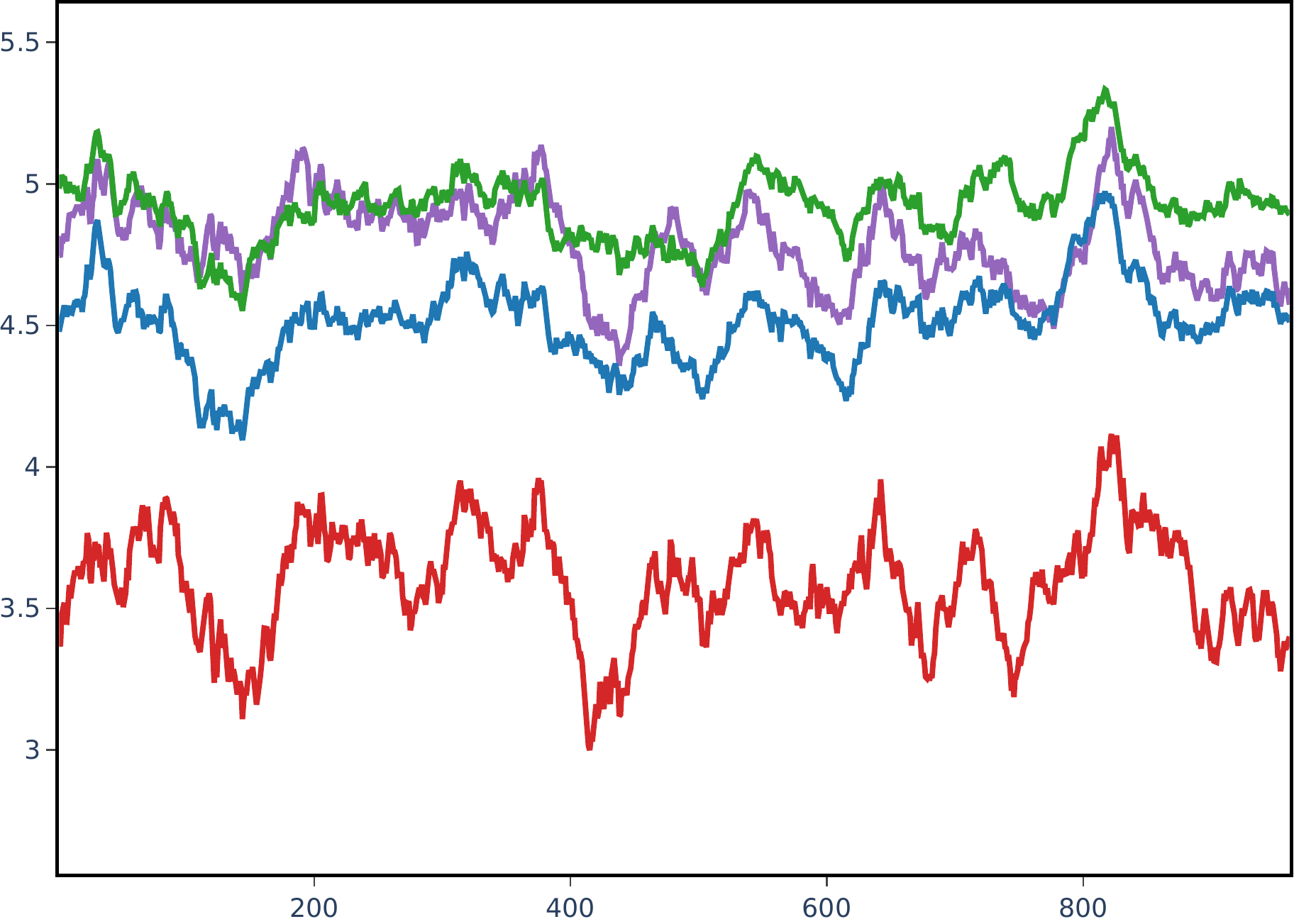} \\
            SVM & \includegraphics[scale=0.12]{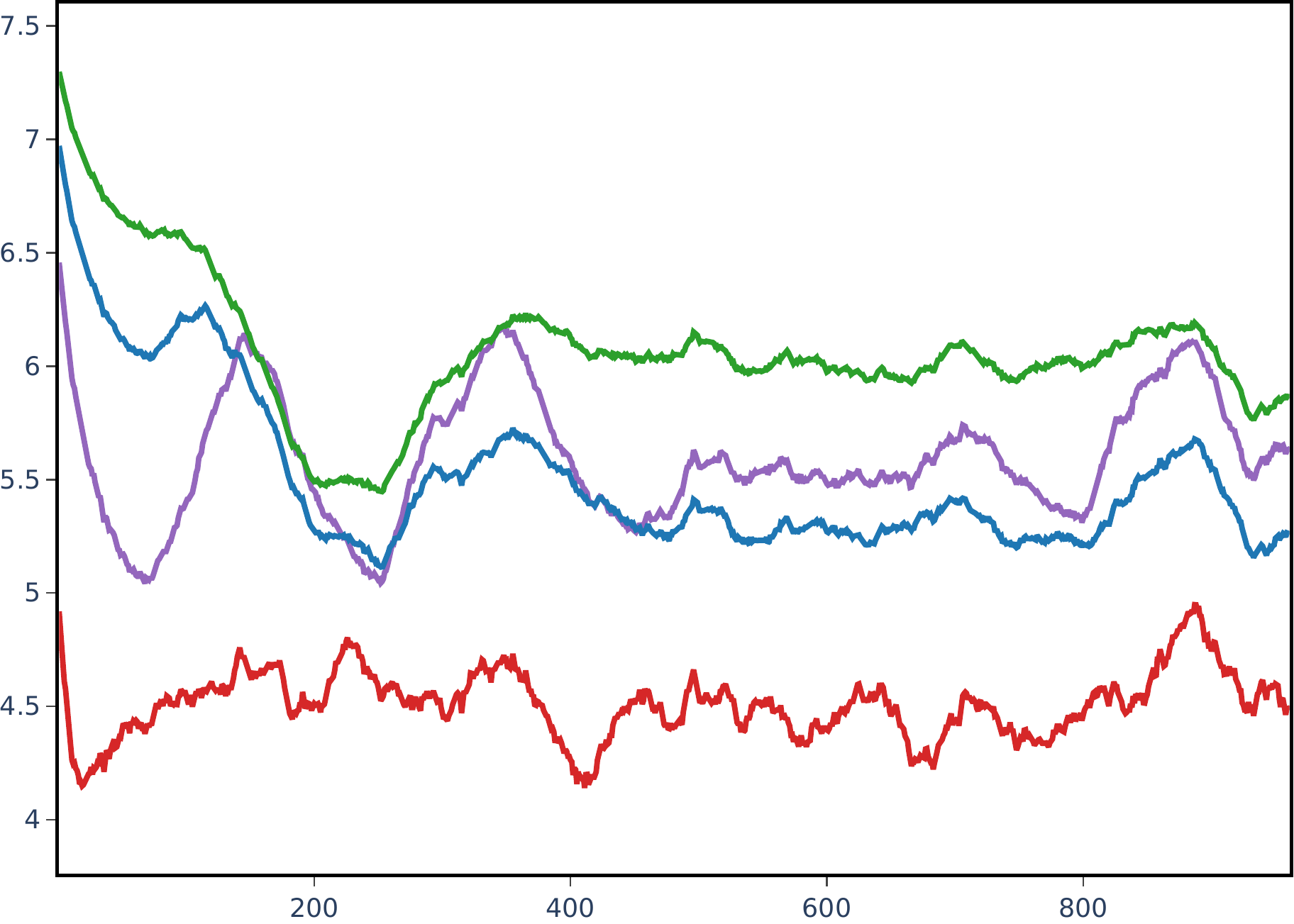}  &  \includegraphics[scale=0.12]{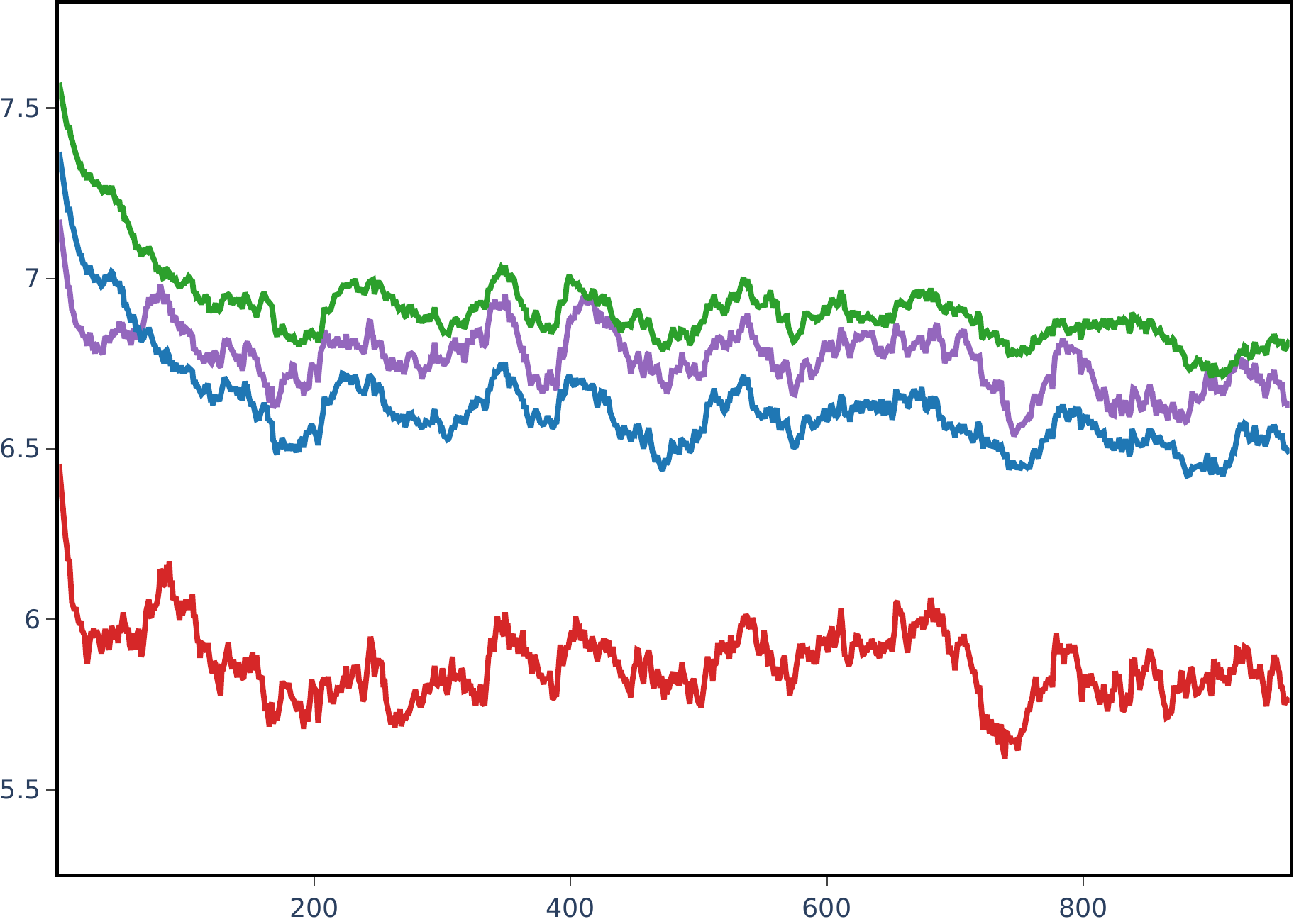} & \includegraphics[scale=0.12]{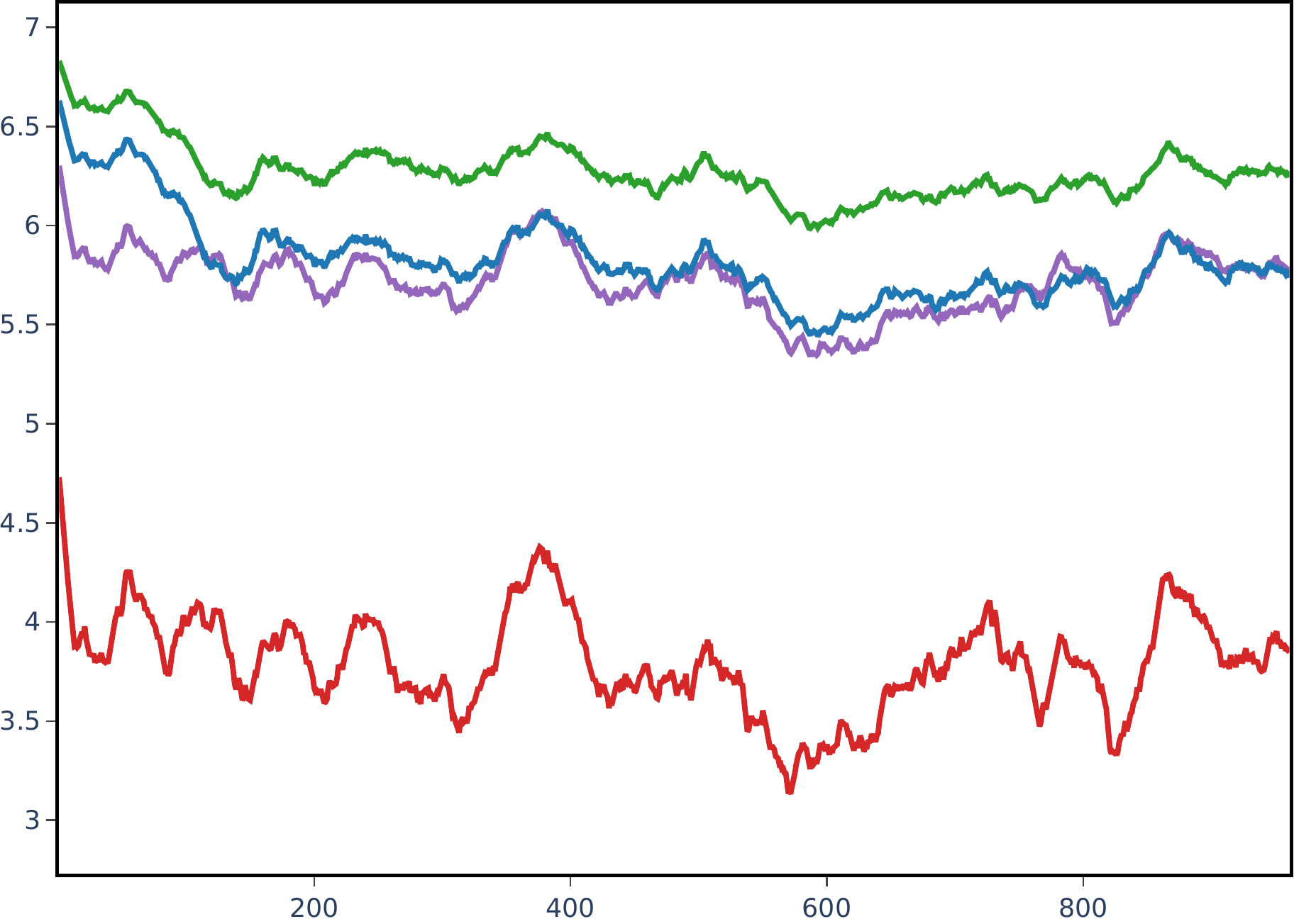} & \includegraphics[scale=0.12]{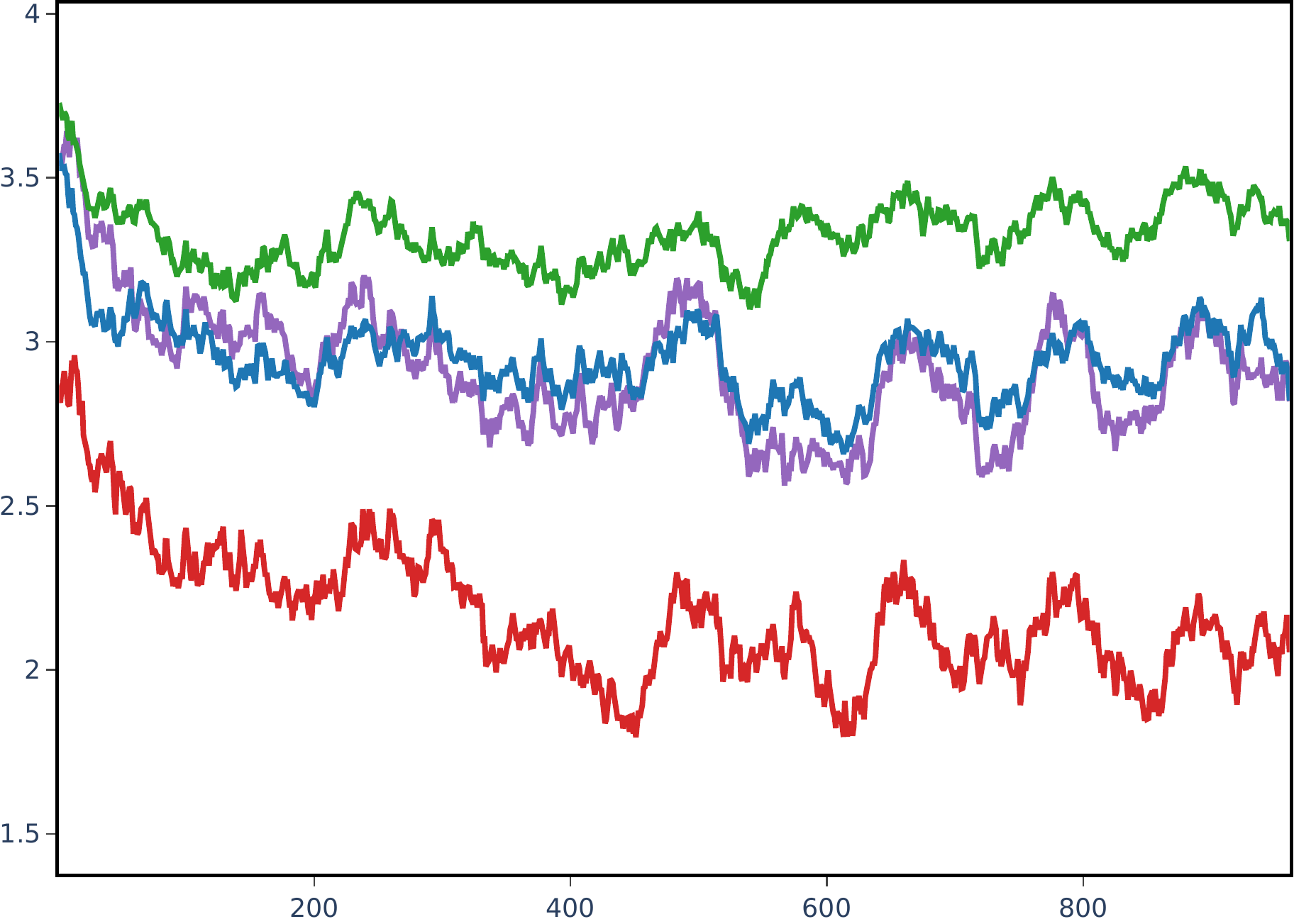} & \includegraphics[scale=0.12]{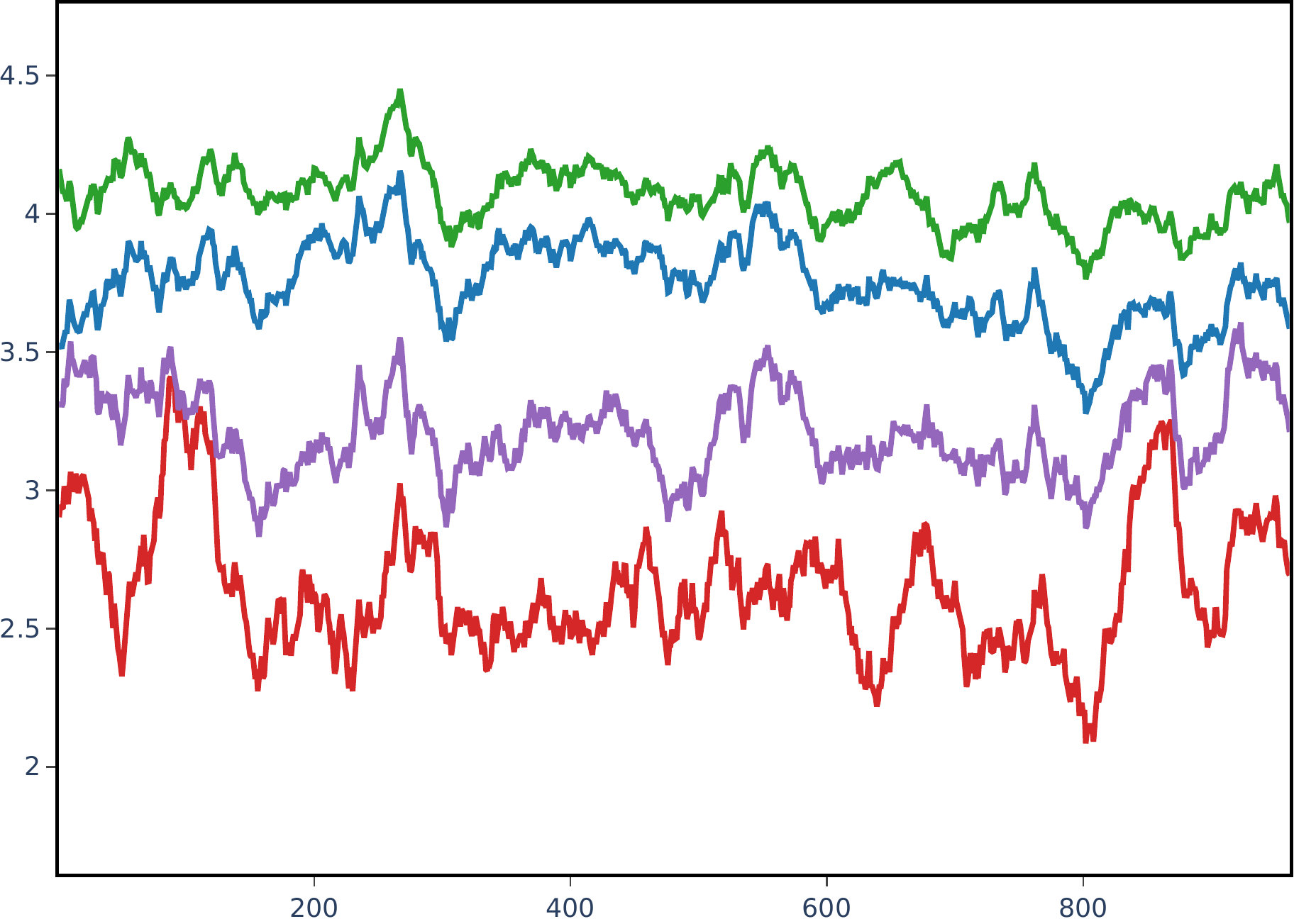}\\
            \midrule
            Legend & \includegraphics[scale=0.12]{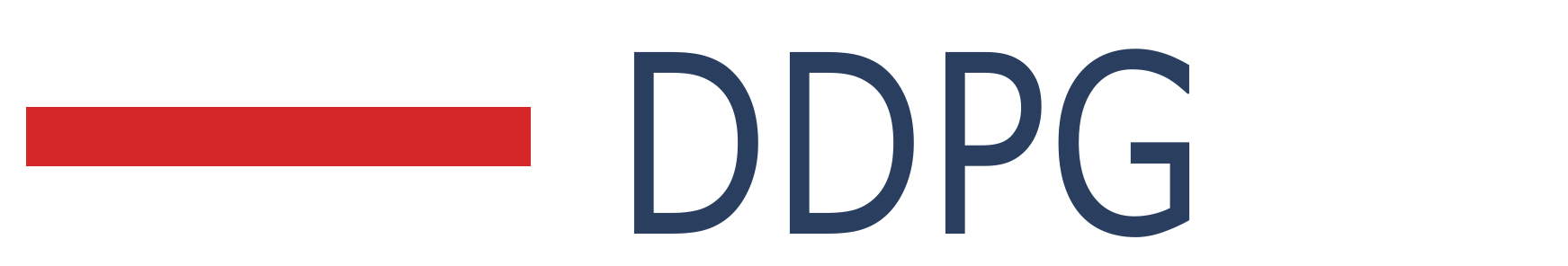} &
            \includegraphics[scale=0.12]{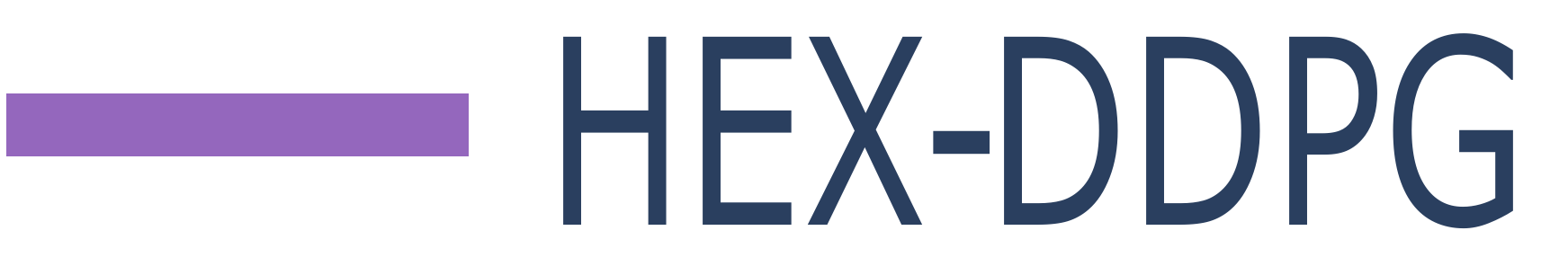} &
            \includegraphics[scale=0.12]{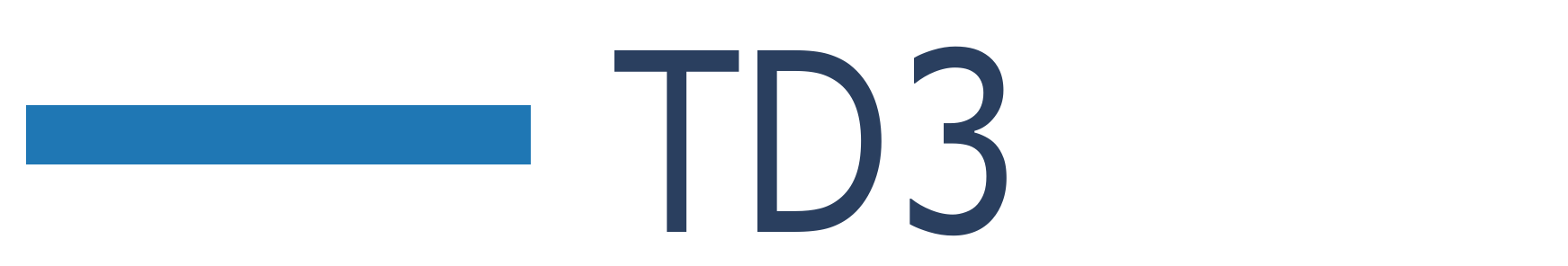} &
            \includegraphics[scale=0.12]{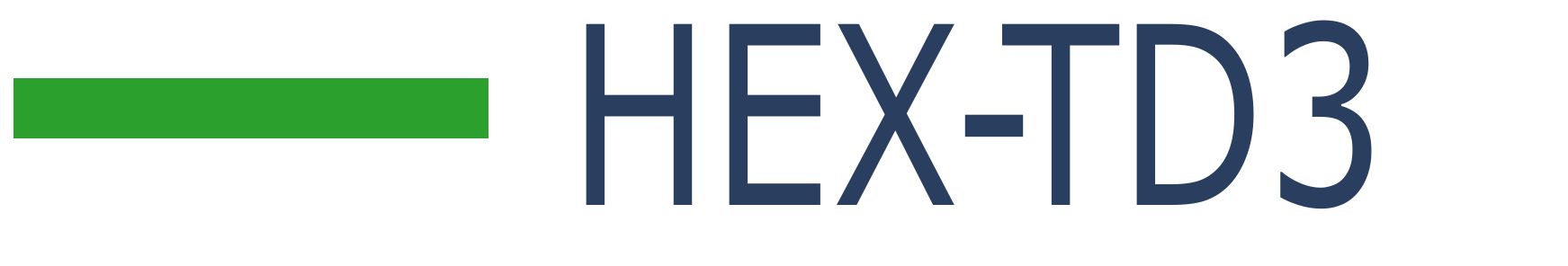} & \\
            \bottomrule
        \end{tabular}
        \caption{Learning curves of each RL method on each dataset and ML model for the decider-free scenario. Plotted values are episode-wise averages taken over ten trials with a rolling window average (window size $=40$) of these averages plotted.}
        \label{tab:lc-free}
\end{table}
First, our HEX-TD3 method is universally able to synthesize higher reward-producing policies across datasets and ML methods. This is, of course, training set performance (more on this shortly). Second, DDPG almost universally learns the lowest reward-producing policies across datasets and ML methods (also more on this shortly). TD3 and HEX-DDPG are often very comparable, falling between HEX-TD3 and DDPG in terms of average reward performance, although generally closer to HEX-TD3 -- i.e.,~DDPG tends to be well beneath the other three methods.

\subsubsection{Testing}

Table \ref{tab:free-testing} shows the average DBD by explainability method, dataset, and ML model. Table \ref{tab:free-test-rewards} in Appendix E shows the average rewards of these results. There are 25 total evaluations (\# datasets $\times$ \# of ML models). Of these a DRL method performs the best on 22/25. LIME performs the best on 2/25 and Grow on 1/25. Curiously, the 3/25 models that Grow and LIME perform the best on are all DT models.
\begin{table}[!htp]
\centering
\begin{tabular}{|cl|ccccc|}
\hline
\multicolumn{2}{|c|}{} & \multicolumn{5}{c|}{Model} \\ \hline
\multicolumn{1}{|l|}{Dataset} & \multicolumn{1}{c|}{Method} & NN & SVM & RF & LR & DT \\ \hline
\multicolumn{1}{|c|}{\multirow{6}{*}{Bank}} & DDPG & 0.28 & \textbf{0.173} & 0.091 & 0.339 & \textbf{0.31} \\
\multicolumn{1}{|c|}{} & HEX-DDPG & \textbf{0.277} & \textbf{0.173} & 0.111 & \textbf{0.33} & 0.331 \\
\multicolumn{1}{|c|}{} & TD3 & 0.298 & 0.212 & \textbf{0.09} & 0.376 & 0.337 \\
\multicolumn{1}{|c|}{} & HEX-TD3 & 0.3 & 0.227 & 0.108 & 0.366 & 0.343 \\
\multicolumn{1}{|c|}{} & LIME & 0.473 & 0.461 & 0.303 & 0.499 & 0.446 \\
\multicolumn{1}{|c|}{} & Grow & 0.476 & 0.462 & 0.335 & 0.495 & 0.427 \\ \hline
\multicolumn{1}{|c|}{\multirow{6}{*}{MIMIC}} & DDPG & \textbf{0.271} & \textbf{0.104} & \textbf{0.046} & 0.215 & 0.386 \\
\multicolumn{1}{|c|}{} & HEX-DDPG & 0.277 & 0.119 & 0.059 & \textbf{0.223} & 0.373 \\
\multicolumn{1}{|c|}{} & TD3 & 0.308 & 0.119 & 0.055 & 0.251 & 0.392 \\
\multicolumn{1}{|c|}{} & HEX-TD3 & 0.313 & 0.146 & 0.063 & 0.246 & 0.388 \\
\multicolumn{1}{|c|}{} & LIME & 0.427 & 0.338 & 0.154 & 0.426 & \textbf{0.321} \\
\multicolumn{1}{|c|}{} & Grow & 0.483 & 0.404 & 0.236 & 0.489 & 0.342 \\ \hline
\multicolumn{1}{|c|}{\multirow{6}{*}{Movie}} & DDPG & 0.24 & \textbf{0.316} & 0.076 & 0.454 & 0.438 \\
\multicolumn{1}{|c|}{} & HEX-DDPG & \textbf{0.207} & 0.335 & 0.097 & \textbf{0.424} & 0.448 \\
\multicolumn{1}{|c|}{} & TD3 & 0.247 & 0.356 & \textbf{0.072} & 0.457 & 0.433 \\
\multicolumn{1}{|c|}{} & HEX-TD3 & 0.235 & 0.381 & 0.092 & 0.452 & 0.45 \\
\multicolumn{1}{|c|}{} & LIME & 0.462 & 0.5 & 0.314 & 0.5 & 0.49 \\
\multicolumn{1}{|c|}{} & Grow & 0.45 & 0.416 & 0.217 & 0.49 & \textbf{0.331} \\ \hline
\multicolumn{1}{|c|}{\multirow{6}{*}{Student}} & DDPG & 0.246 & \textbf{0.176} & 0.092 & 0.374 & 0.381 \\
\multicolumn{1}{|c|}{} & HEX-DDPG & 0.198 & 0.212 & \textbf{0.082} & 0.247 & 0.372 \\
\multicolumn{1}{|c|}{} & TD3 & 0.239 & 0.178 & 0.092 & 0.371 & 0.385 \\
\multicolumn{1}{|c|}{} & HEX-TD3 & \textbf{0.192} & 0.206 & 0.083 & \textbf{0.246} & 0.363 \\
\multicolumn{1}{|c|}{} & LIME & 0.31 & 0.195 & 0.101 & 0.37 & \textbf{0.225} \\
\multicolumn{1}{|c|}{} & Grow & 0.42 & 0.304 & 0.126 & 0.449 & 0.234 \\ \hline
\multicolumn{1}{|c|}{\multirow{6}{*}{News}} & DDPG & \textbf{0.367} & 0.425 & \textbf{0.05} & \textbf{0.26} & 0.493 \\
\multicolumn{1}{|c|}{} & HEX-DDPG & 0.382 & 0.385 & 0.067 & 0.274 & 0.493 \\
\multicolumn{1}{|c|}{} & TD3 & 0.387 & 0.429 & 0.056 & 0.3 & 0.496 \\
\multicolumn{1}{|c|}{} & HEX-TD3 & 0.377 & \textbf{0.383} & 0.068 & 0.278 & \textbf{0.492} \\
\multicolumn{1}{|c|}{} & LIME & 0.496 & 0.448 & 0.294 & 0.494 & 0.498 \\
\multicolumn{1}{|c|}{} & Grow & 0.497 & 0.482 & 0.271 & 0.483 & 0.436 \\ \hline
\end{tabular}
\caption{Decider-free scenario. Average DBD. \textbf{Bold} indicates the lowest average DBD obtain on each dataset and ML model.\label{tab:free-testing}}
\end{table}

The success of the DDPG and TD3 methods, particularly DDPG, is interesting provided the learning curve results. The out-of-the-box methods perform the best on 11/25 experiments and our HEX methods also perform the best on 11/25. More concretely, DDPG performed the best on 9/25 evaluations, TD3 performed the best on 2/25, HEX-DDPG performed the best on 7/25, and HEX-TD3 performed the best on 4/25. Recall that DDPG produced the lowest learning curves across all datasets and methods and HEX-TD3 produced the highest. We suspect that the added greediness of our methods may, at times, lead to overfitting. Nevertheless, these results show there are benefits to our proposed additions even in non-HITL scenarios. We again stress that a DRL approach outperforms other state-of-the-art methods on 22/25 evaluations. 

These results also reveal several additional artifacts worth mentioning. First, the linear model, logistic regression (LR), tends to have larger DBD values across the various datasets and explainability methods. This is interesting because the method itself is naturally interpretable (due to it's linear nature). Second, random forest (RF), a highly non-linear model, generally has lower DBD values across the various explainability methods, particularly the DRL methods. This is ideal since explainability methods are adopted to overcome the unexplainable nature of nonlinear, black box models in the first place.

\subsection{HITL Decider Scenario}
This subsection presents the results of the HITL decider scenario. We consider three simulated decider scenarios using 10, 50, and 90\% of features as the UAPs for each, respectively (i.e., 90, 50, and 10\% of features are preferred by the decider). We randomly select a different set of features for each UAP scenario, for each of the 10 trials, holding the selection consistent across each explainability method and ML model. We begin by examining the learning curves of the various DRL methods, subsequently examining the test set performance of all models first in terms of average DBD and then in terms of average UEP. Note that DDPG and TD3, along with LIME and Grow, do not explicitly consider a HITL. We therefore only report one set of results for each of these explainability methods.

\subsubsection{Learning Curves}

The DRL learning curves for the HITL decider scenario are shown in Table \ref{tab:lc-hitl}. Note that there are three learning curves for each HEX method, but only one for the out-of-the-box methods, since the out-of-the-box methods do not consider the decider's preferred features.
\begin{table}[htbp]
        \centering
        \begin{tabular}{cM{24mm}M{24mm}M{24mm}M{24mm}M{24mm}}
           \toprule
             & Bank & MIMIC & Movie & News & Student\\
            \midrule
            DT & \includegraphics[scale=0.12]{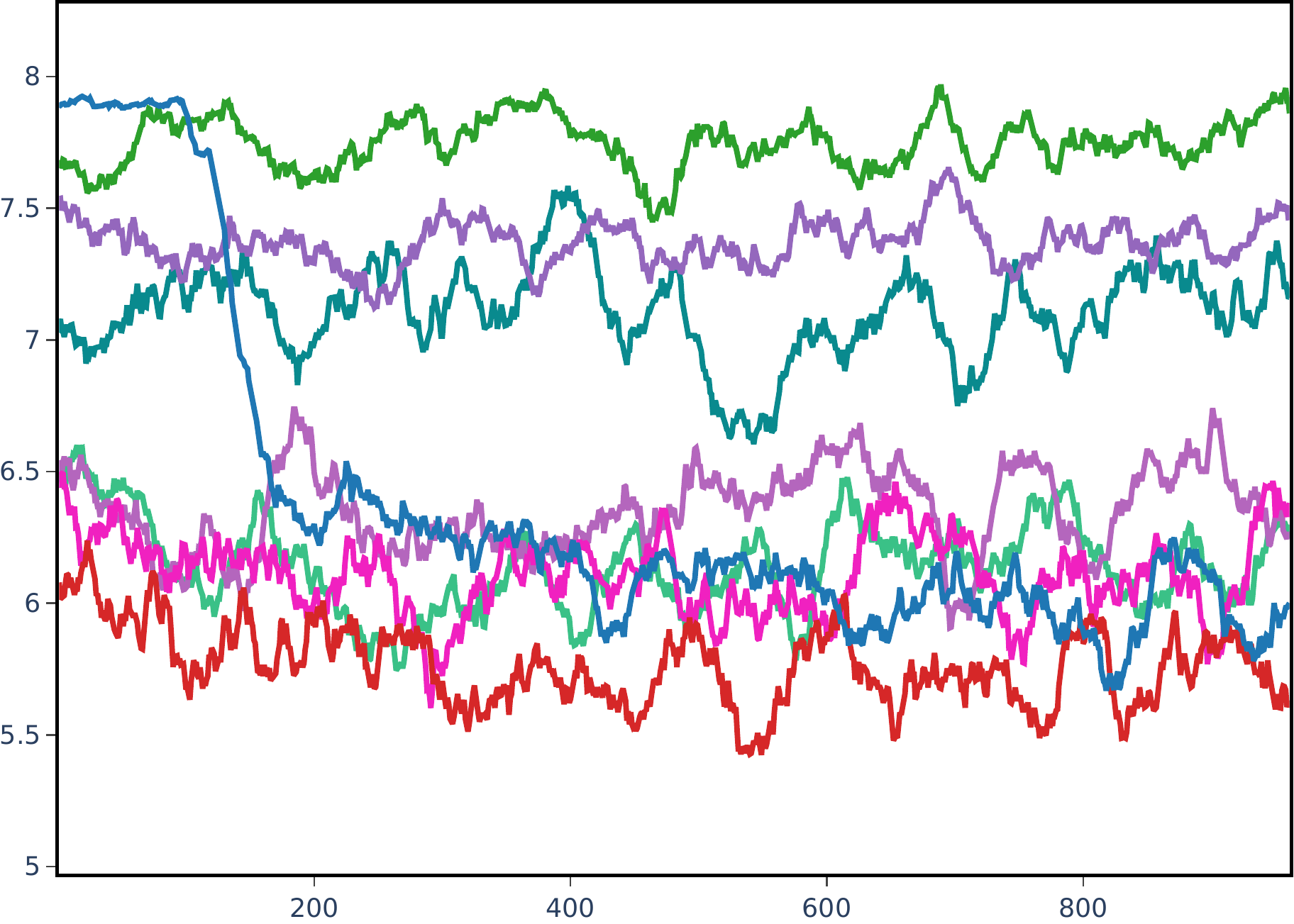} &  \includegraphics[scale=0.12]{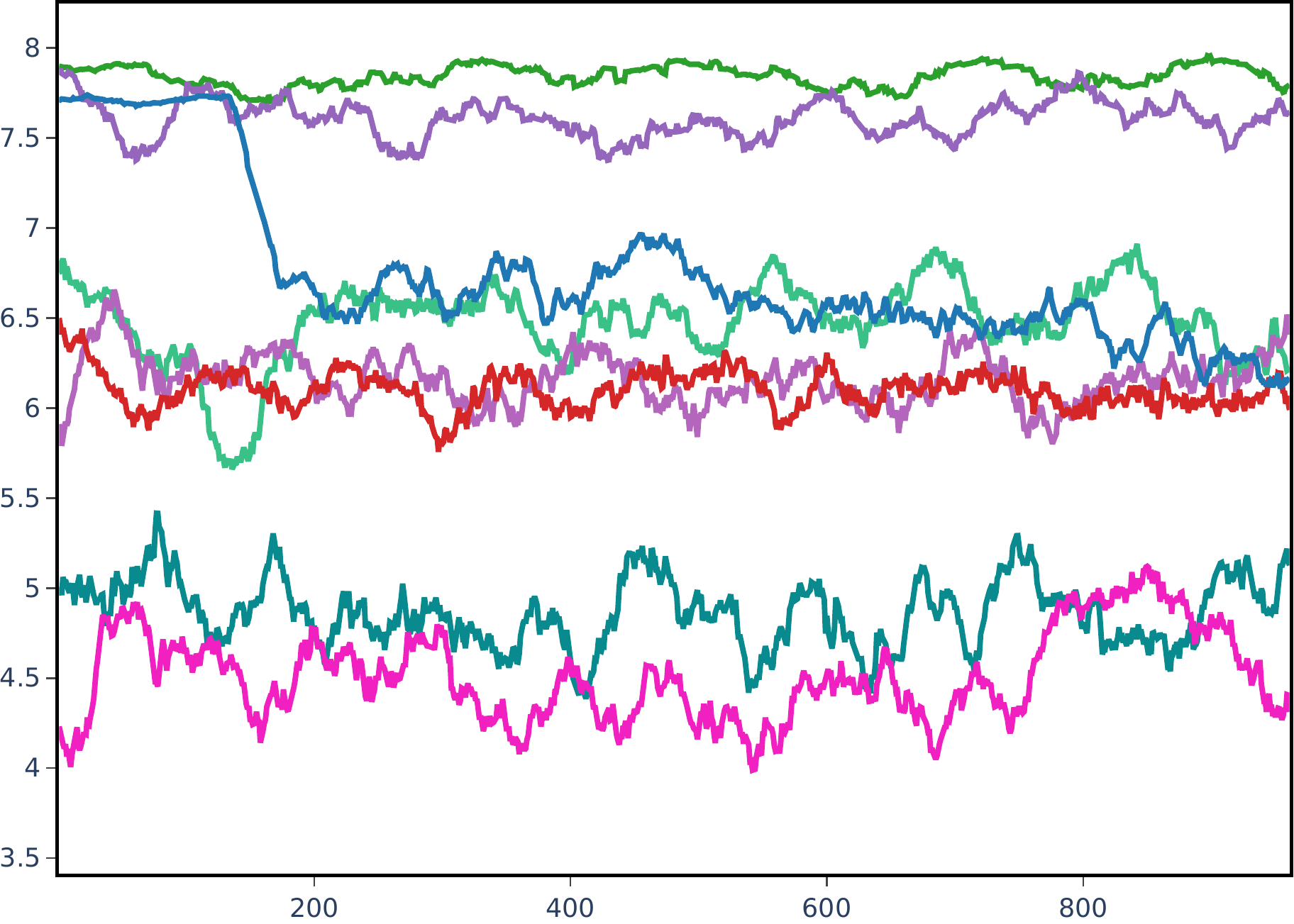} & \includegraphics[scale=0.12]{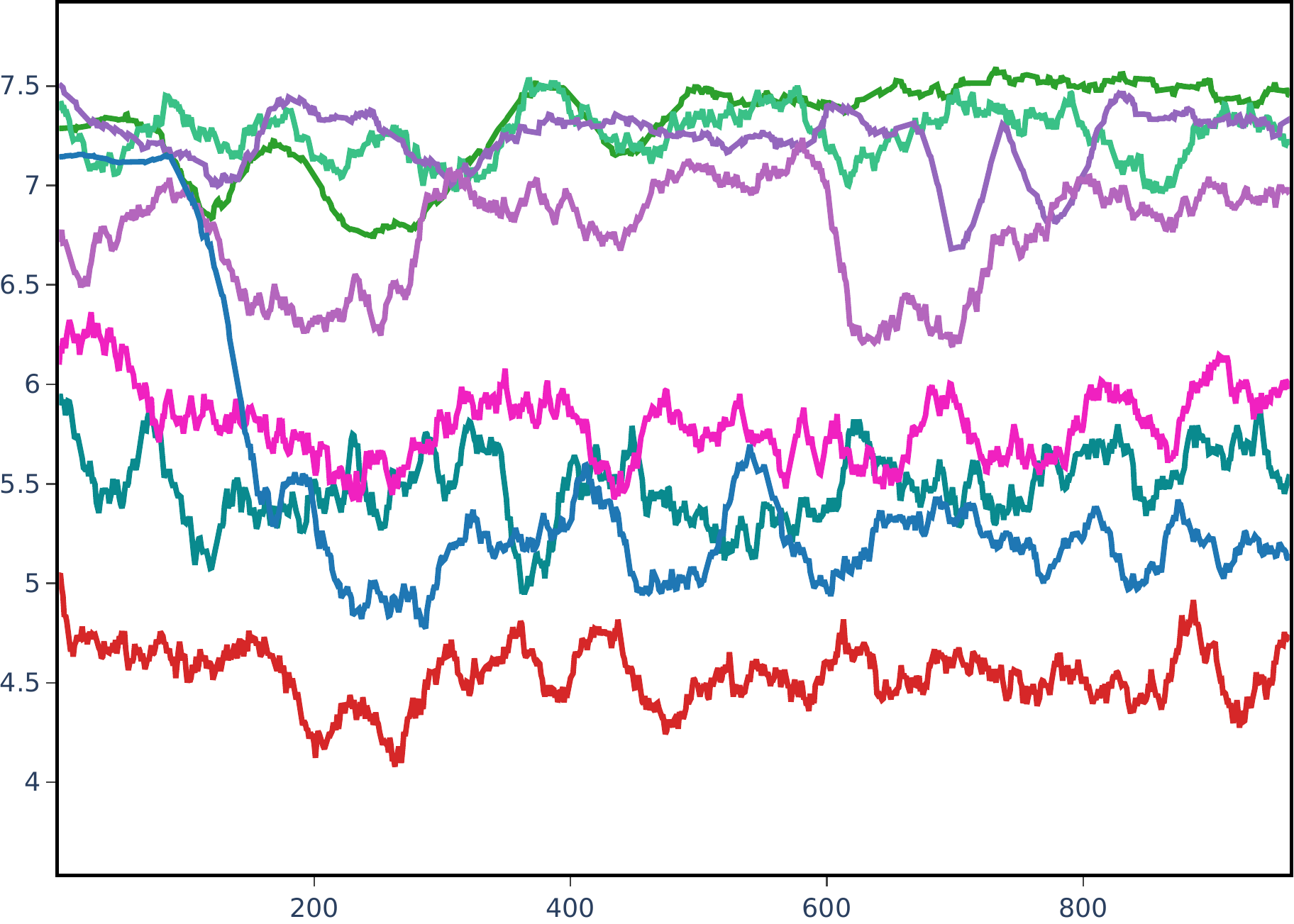} & \includegraphics[scale=0.12]{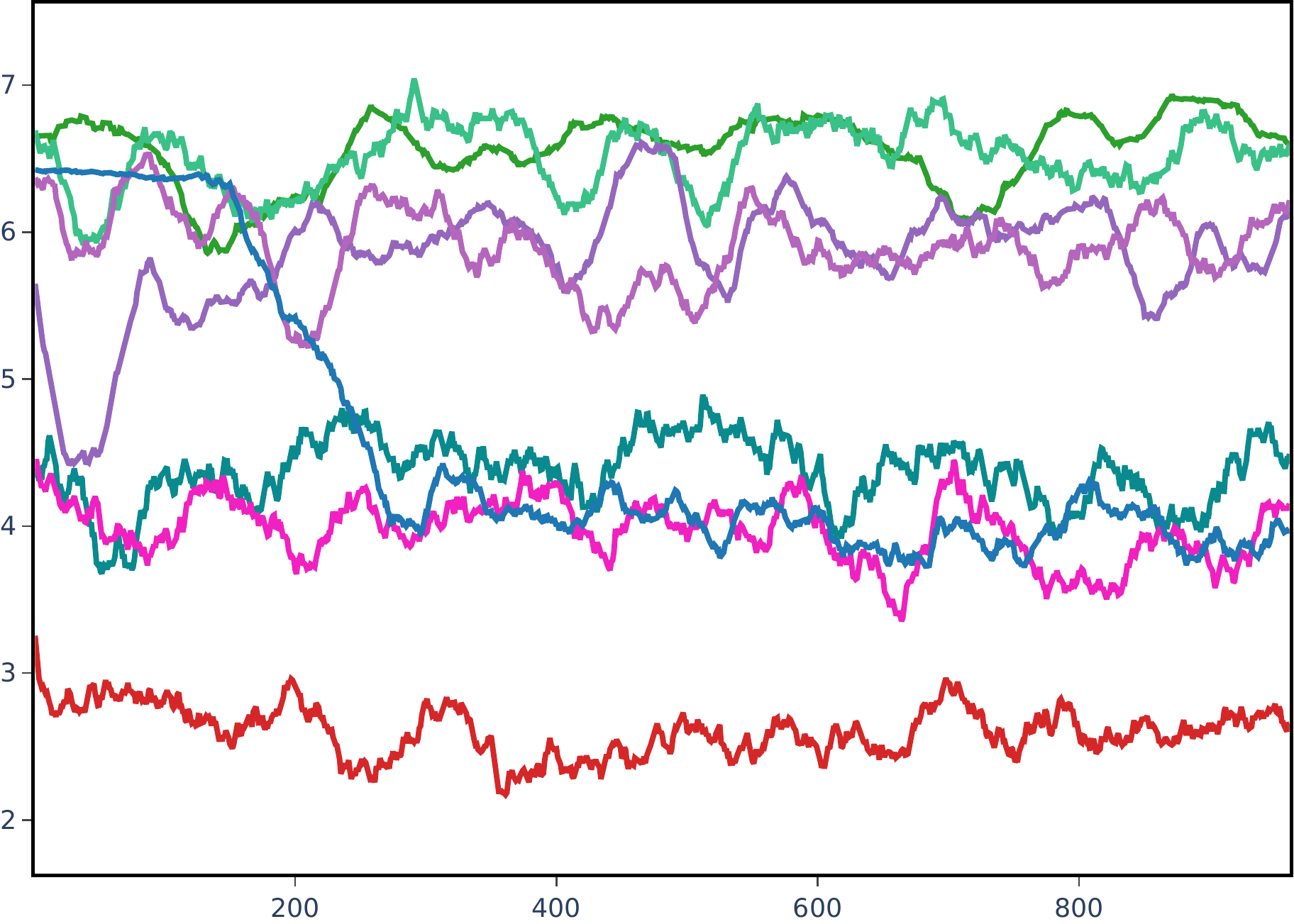} & \includegraphics[scale=0.12]{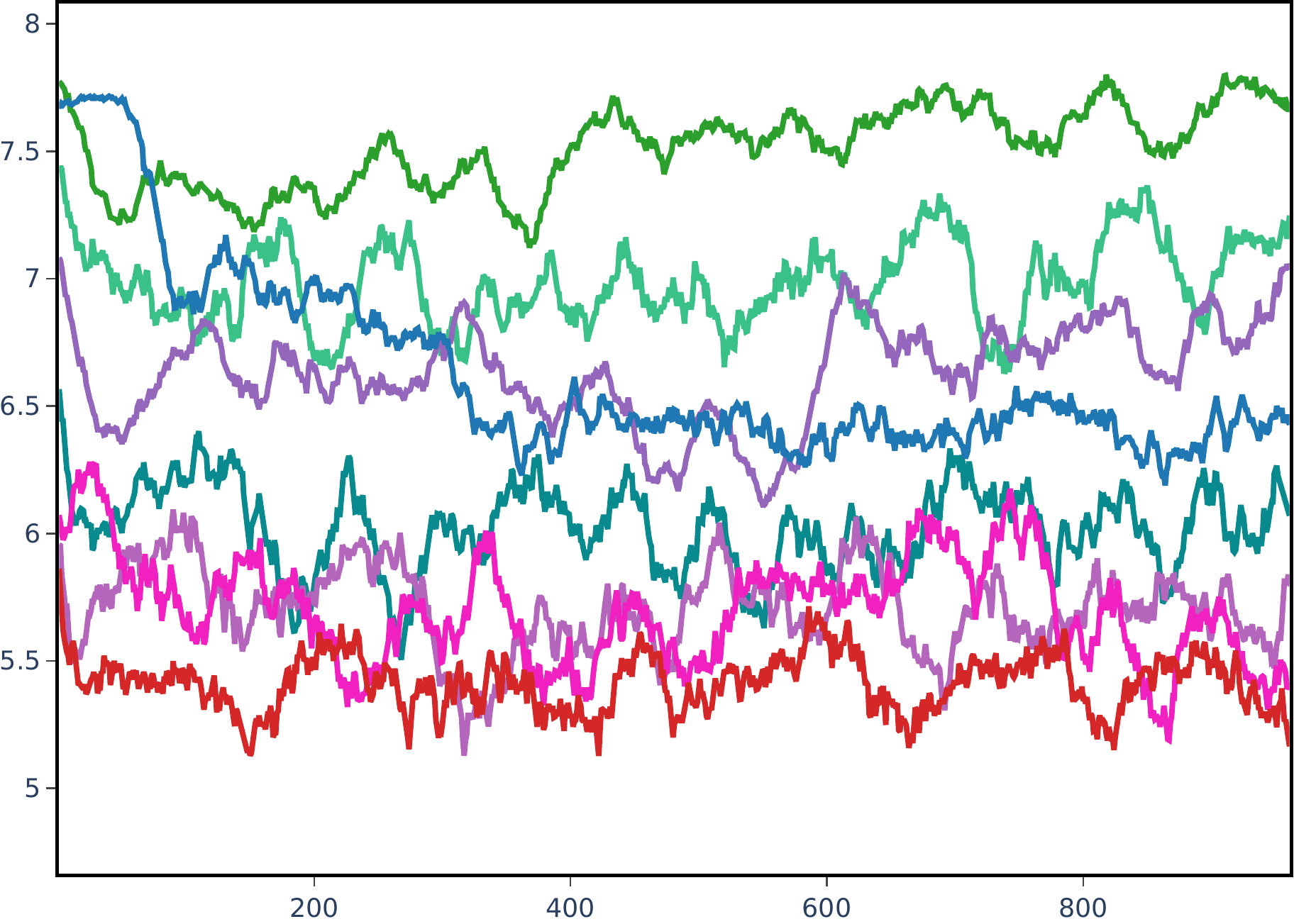}\\
            LR & \includegraphics[scale=0.12]{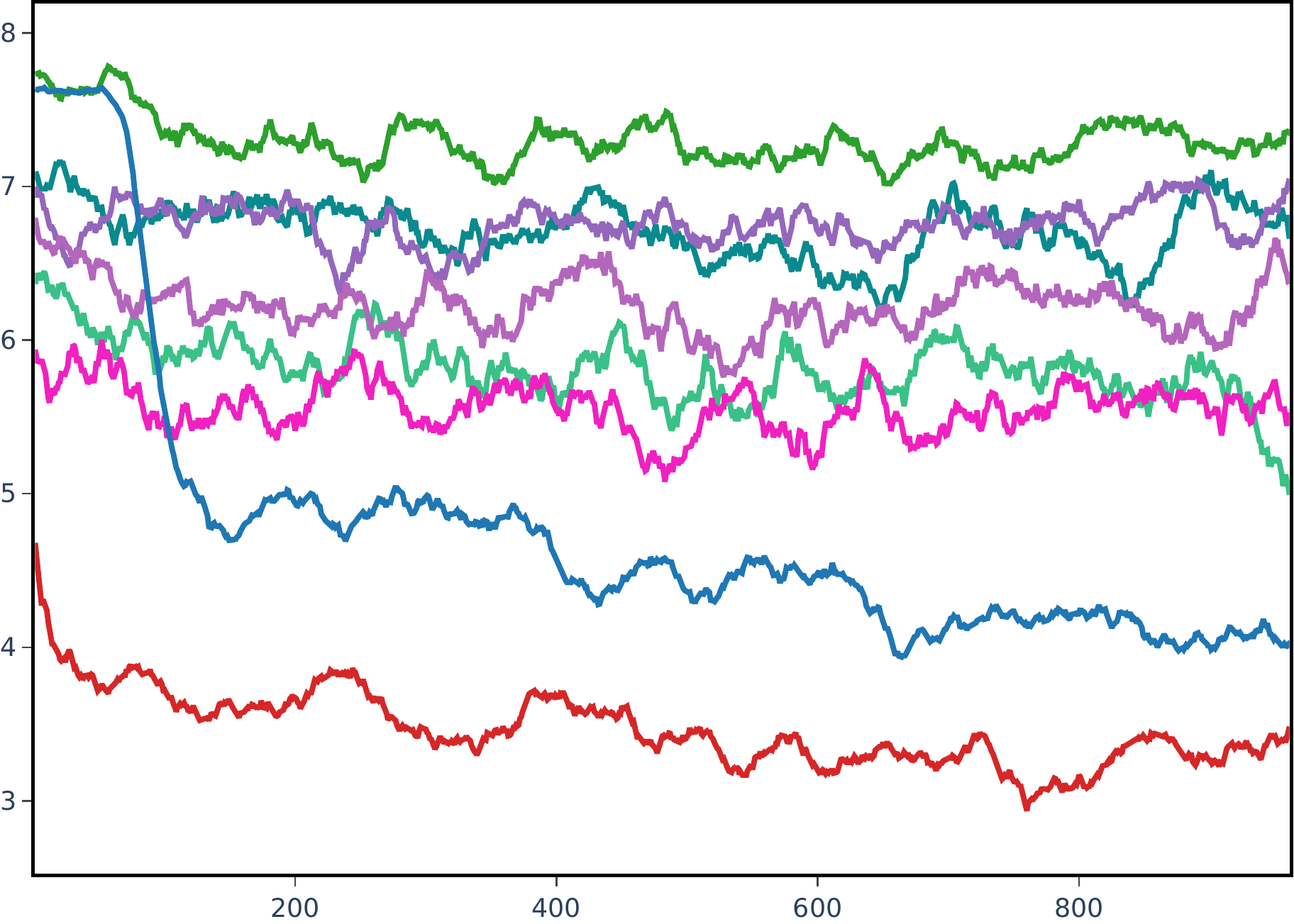}  &  \includegraphics[scale=0.12]{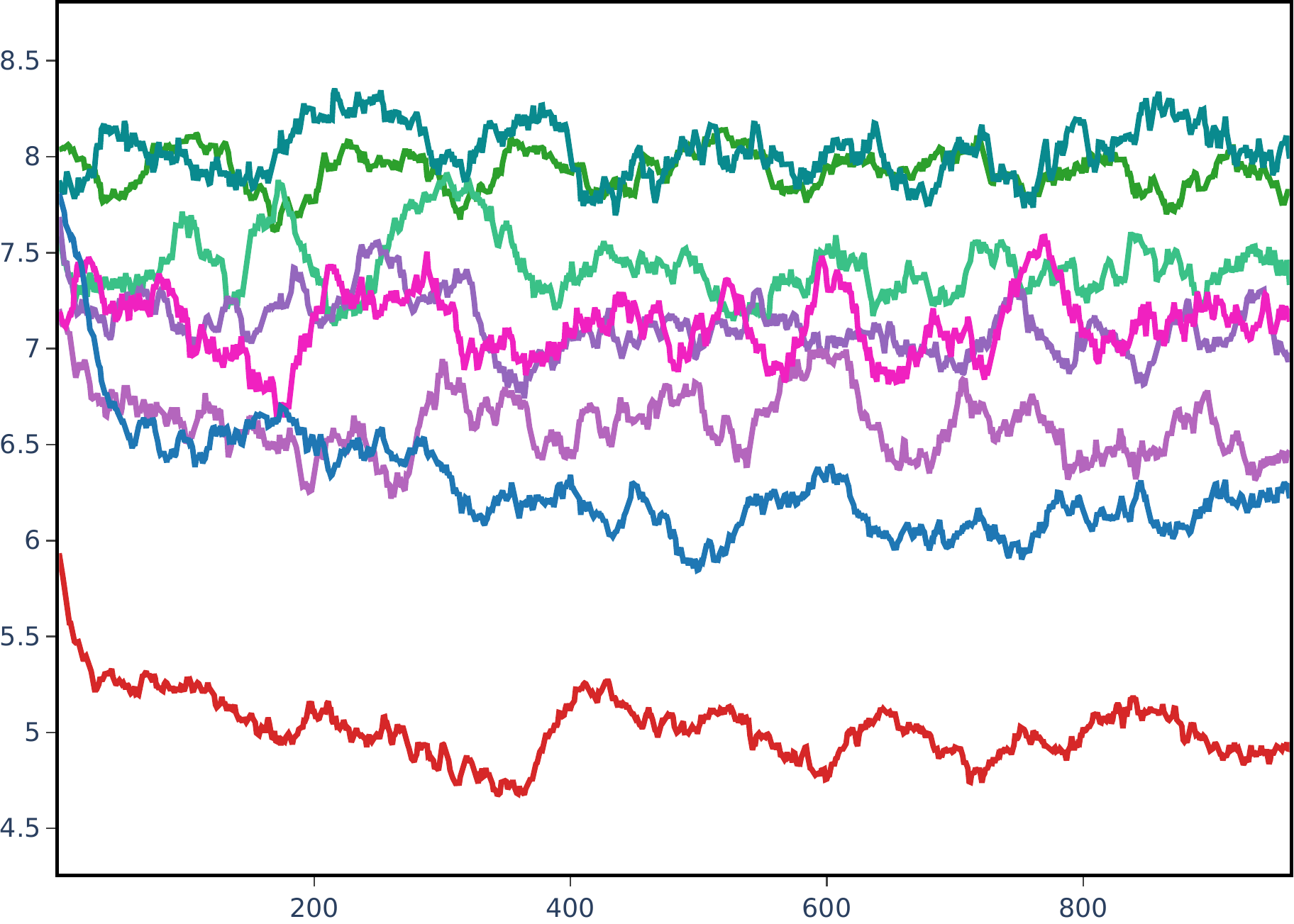} & \includegraphics[scale=0.12]{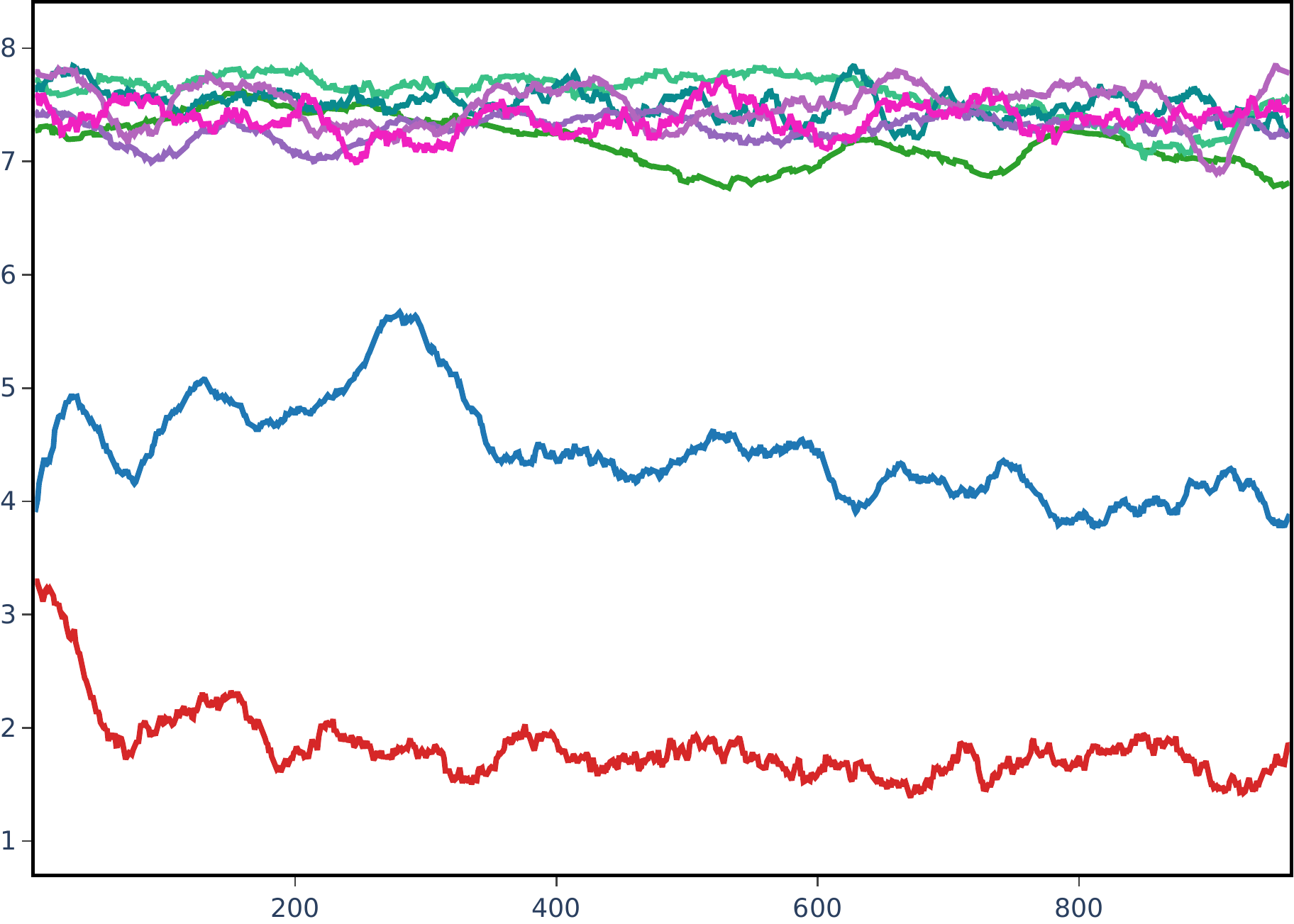} & \includegraphics[scale=0.12]{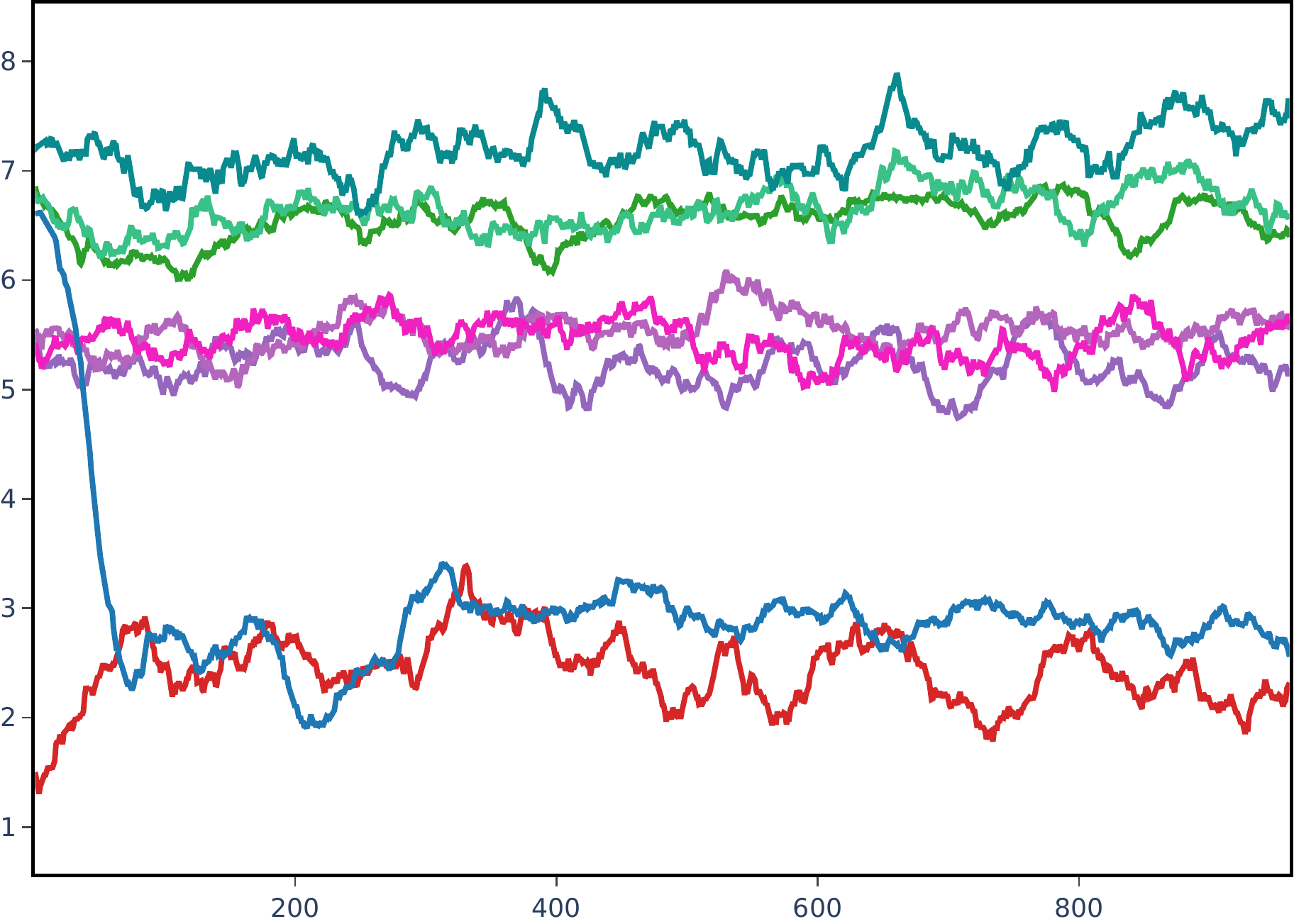} & \includegraphics[scale=0.12]{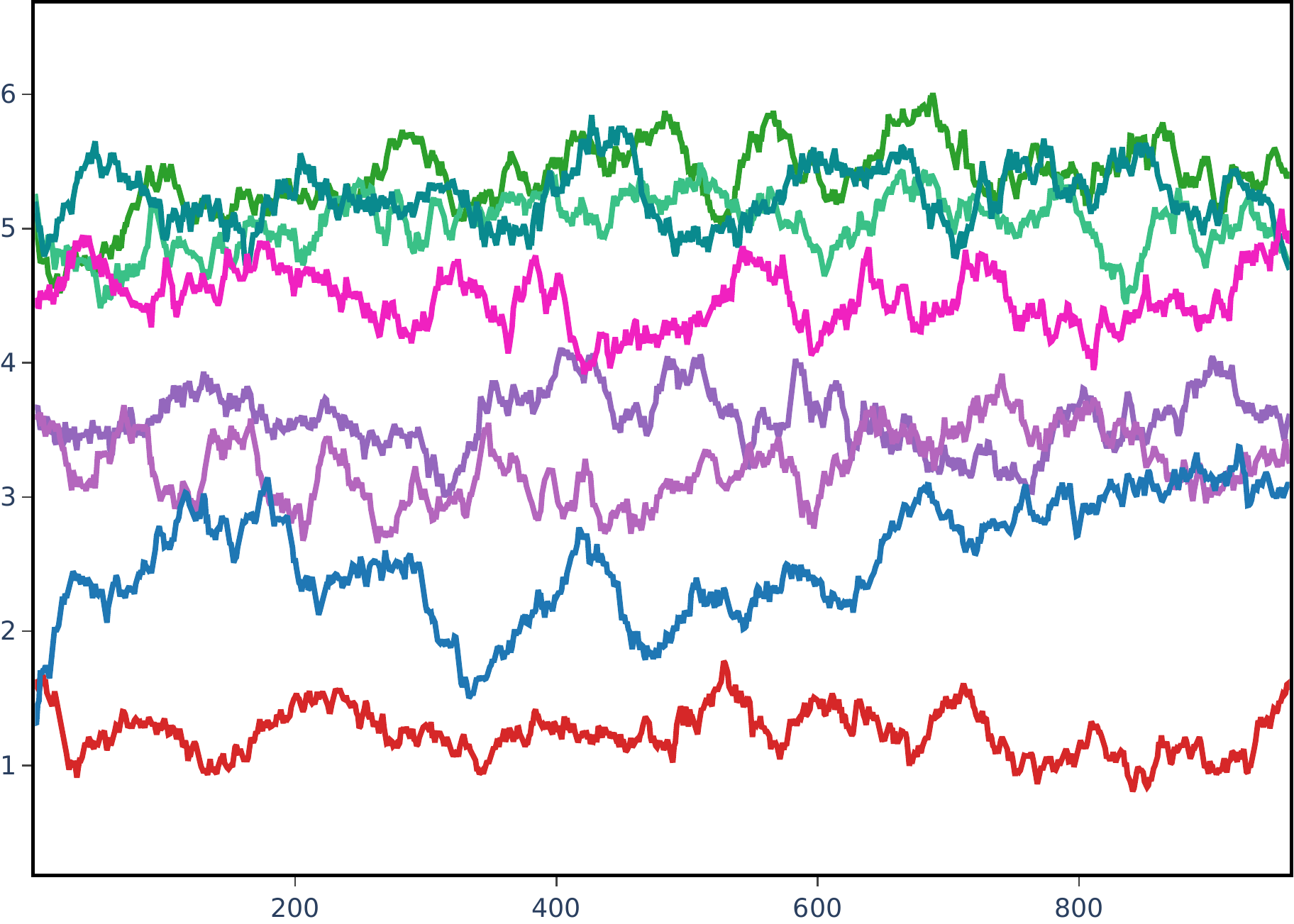} \\
            NN & \includegraphics[scale=0.12]{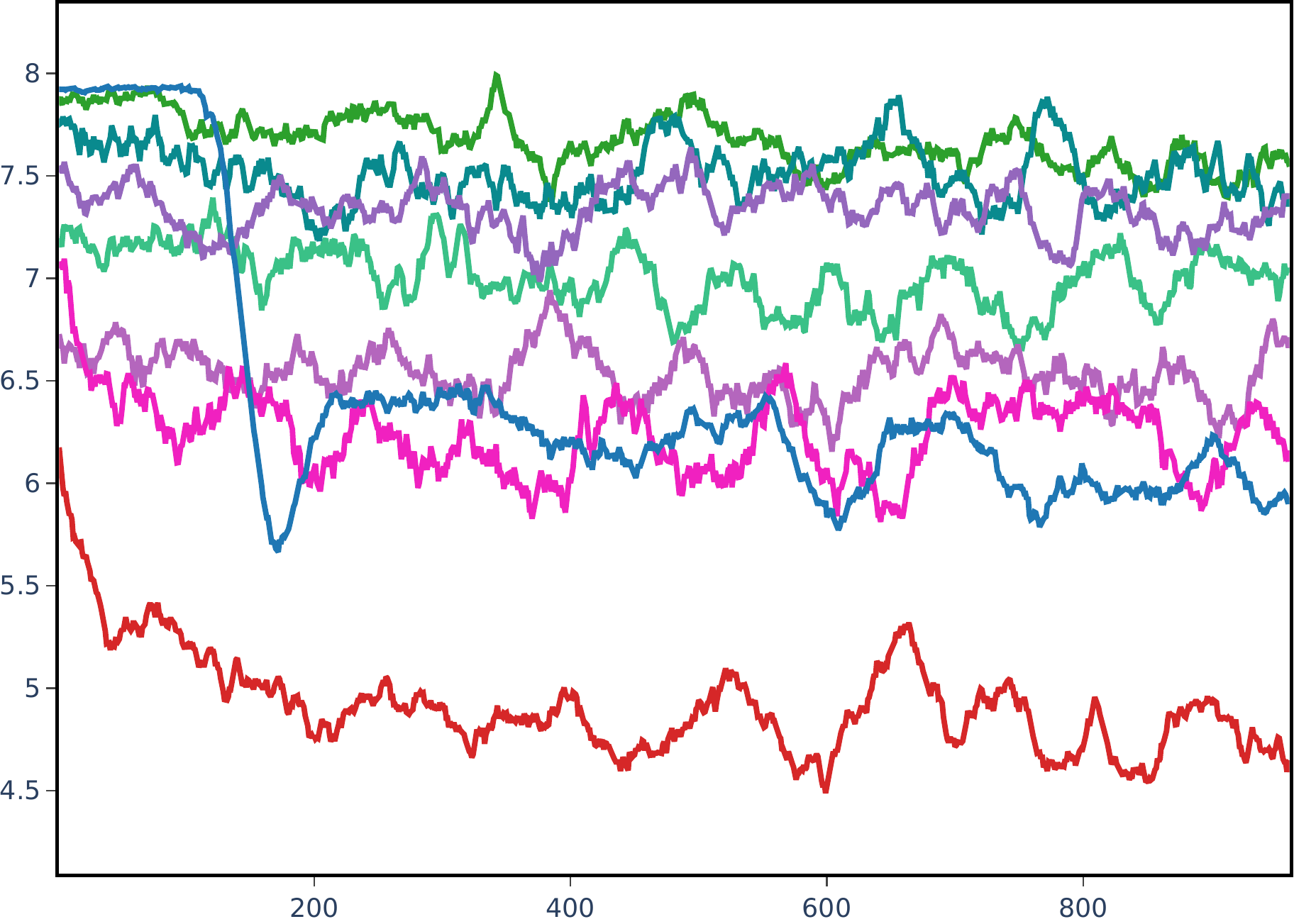}  &  \includegraphics[scale=0.12]{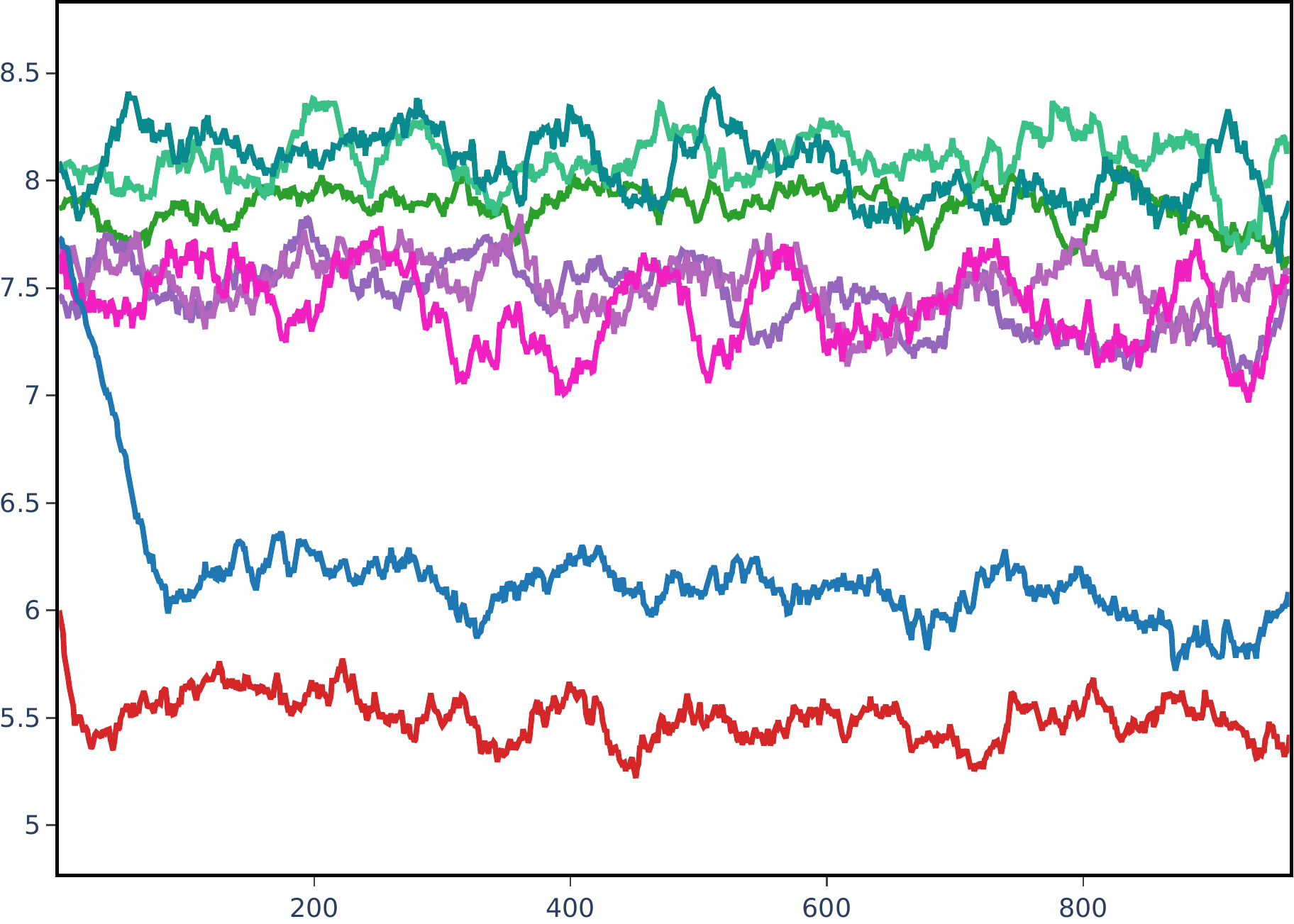} & \includegraphics[scale=0.12]{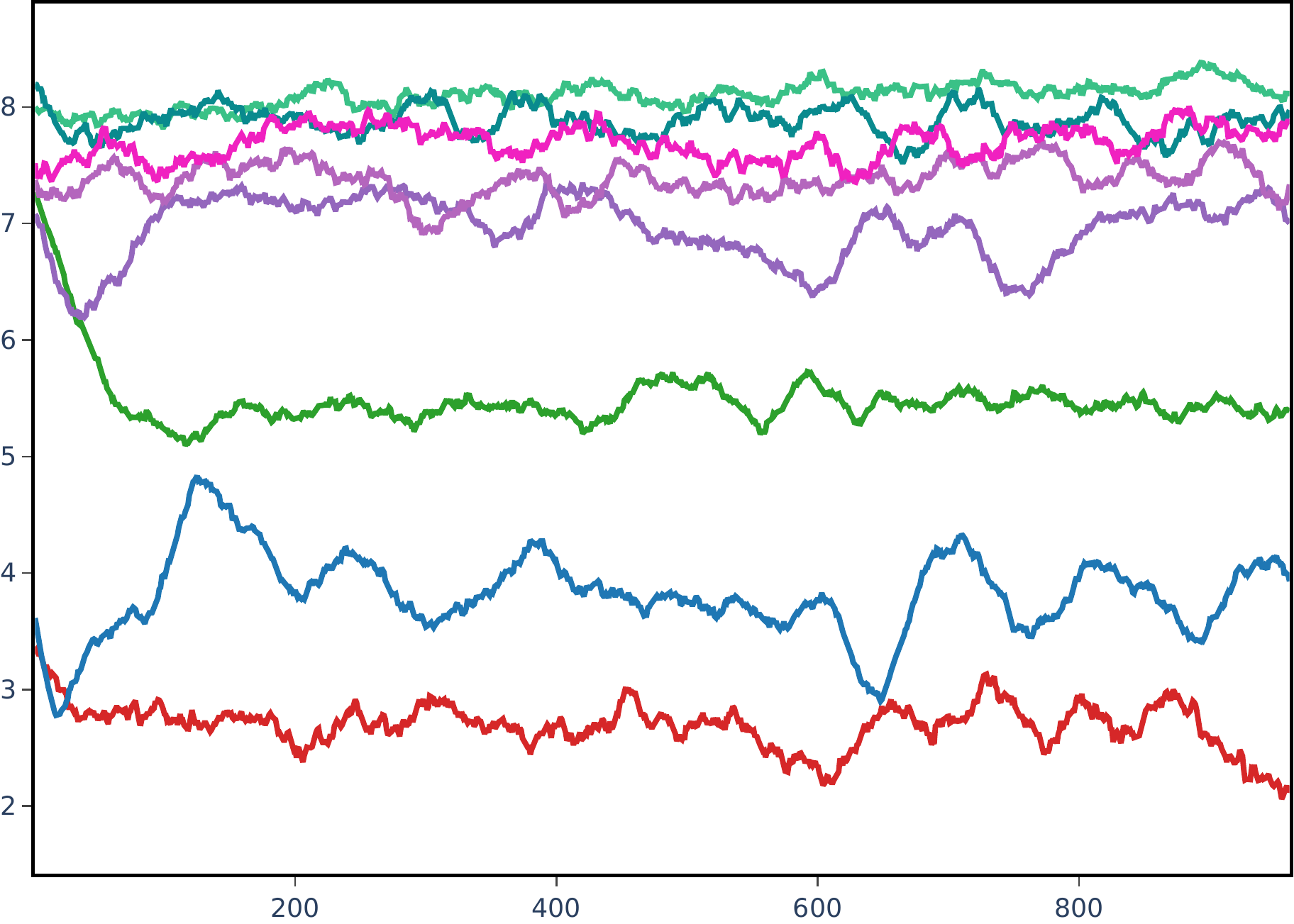} & \includegraphics[scale=0.12]{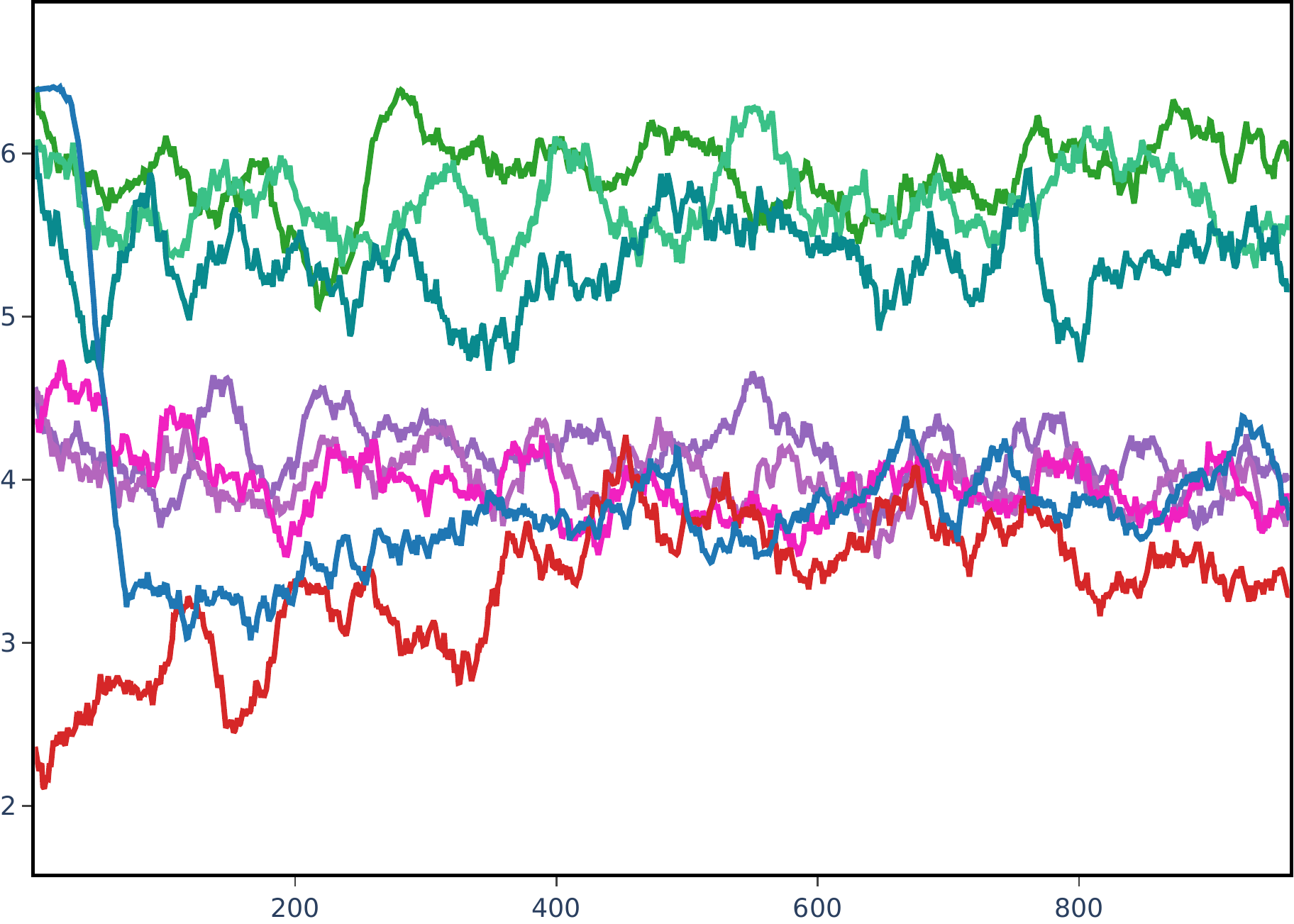} & \includegraphics[scale=0.12]{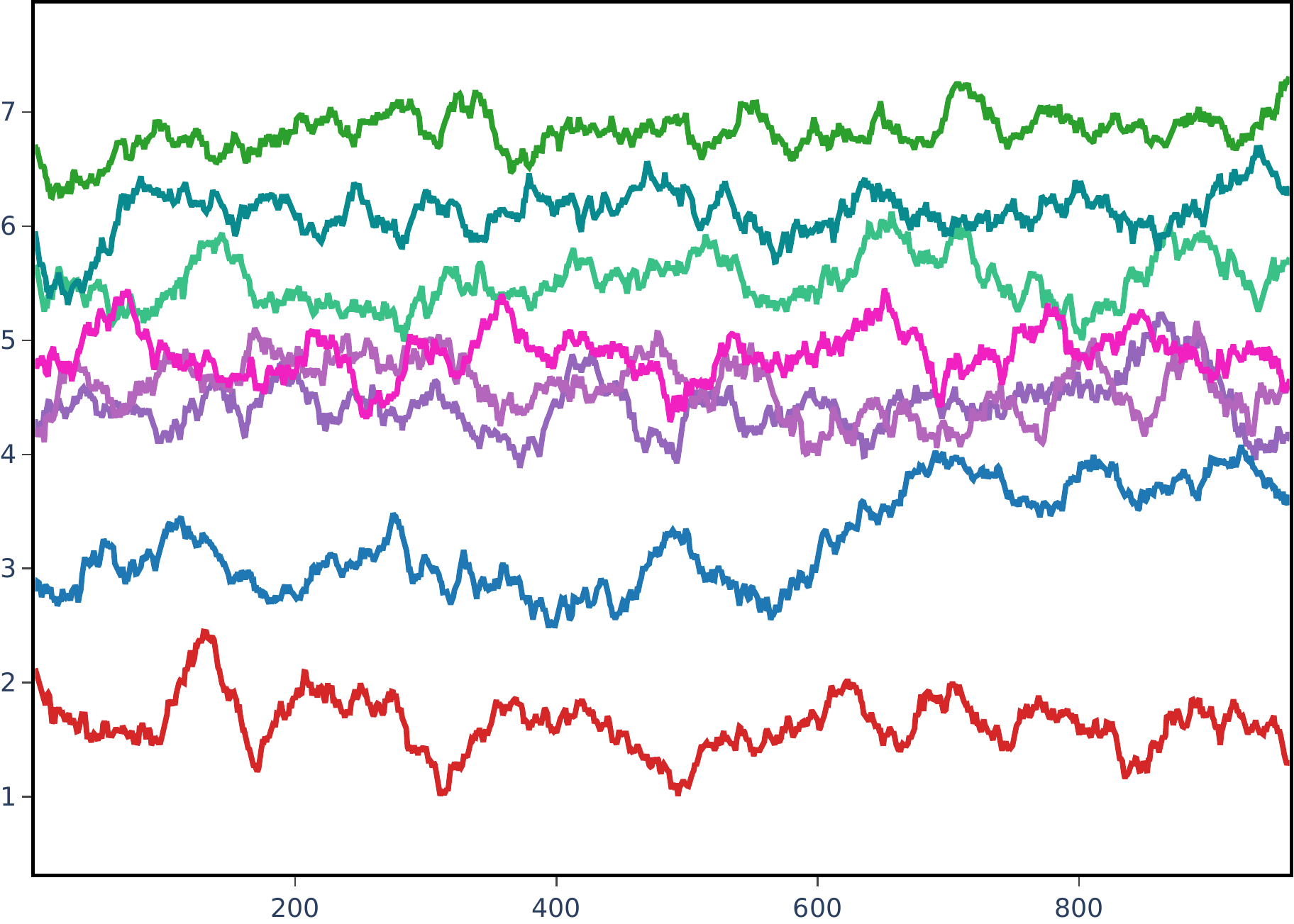} \\
            RF & \includegraphics[scale=0.12]{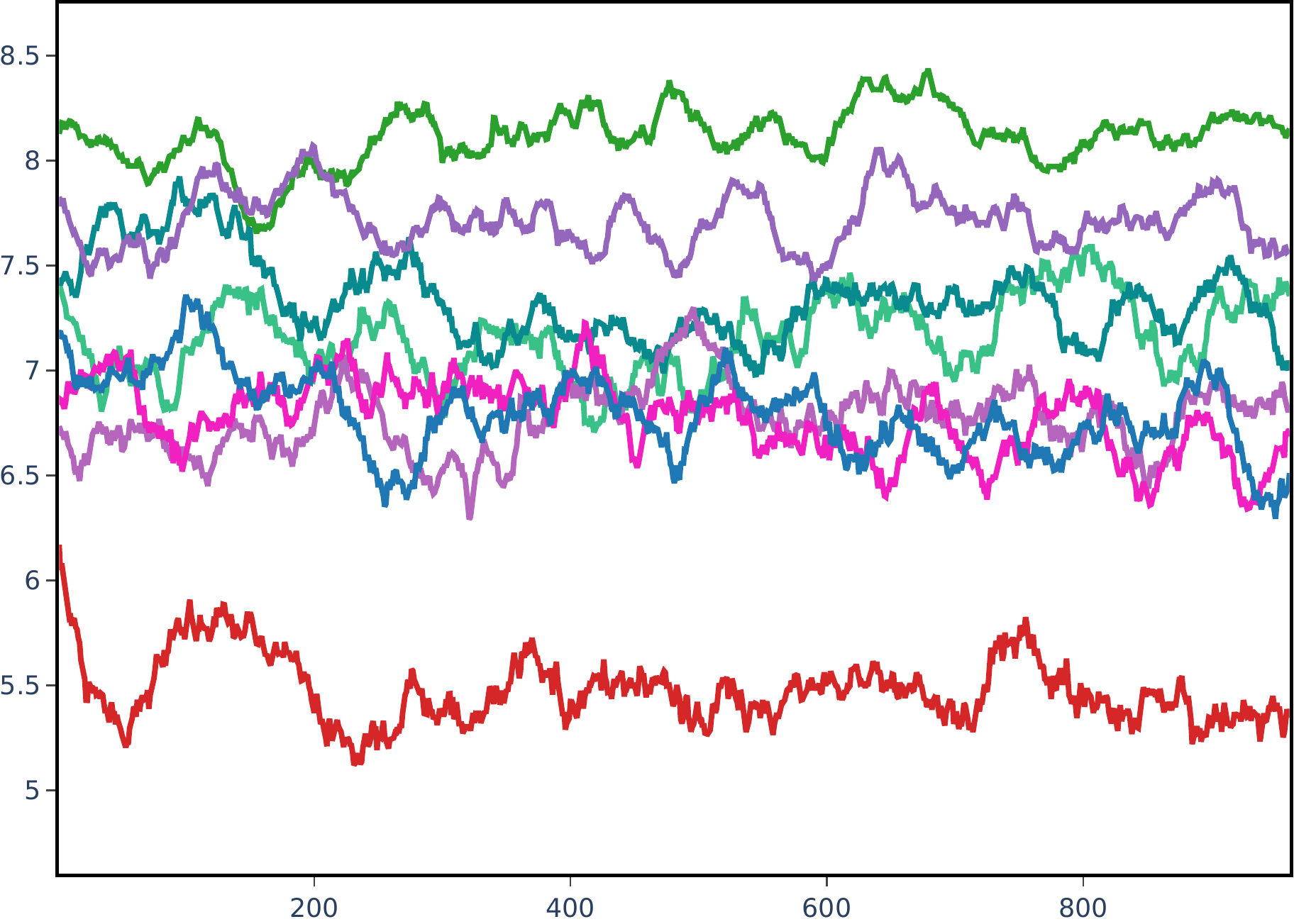}  &  \includegraphics[scale=0.12]{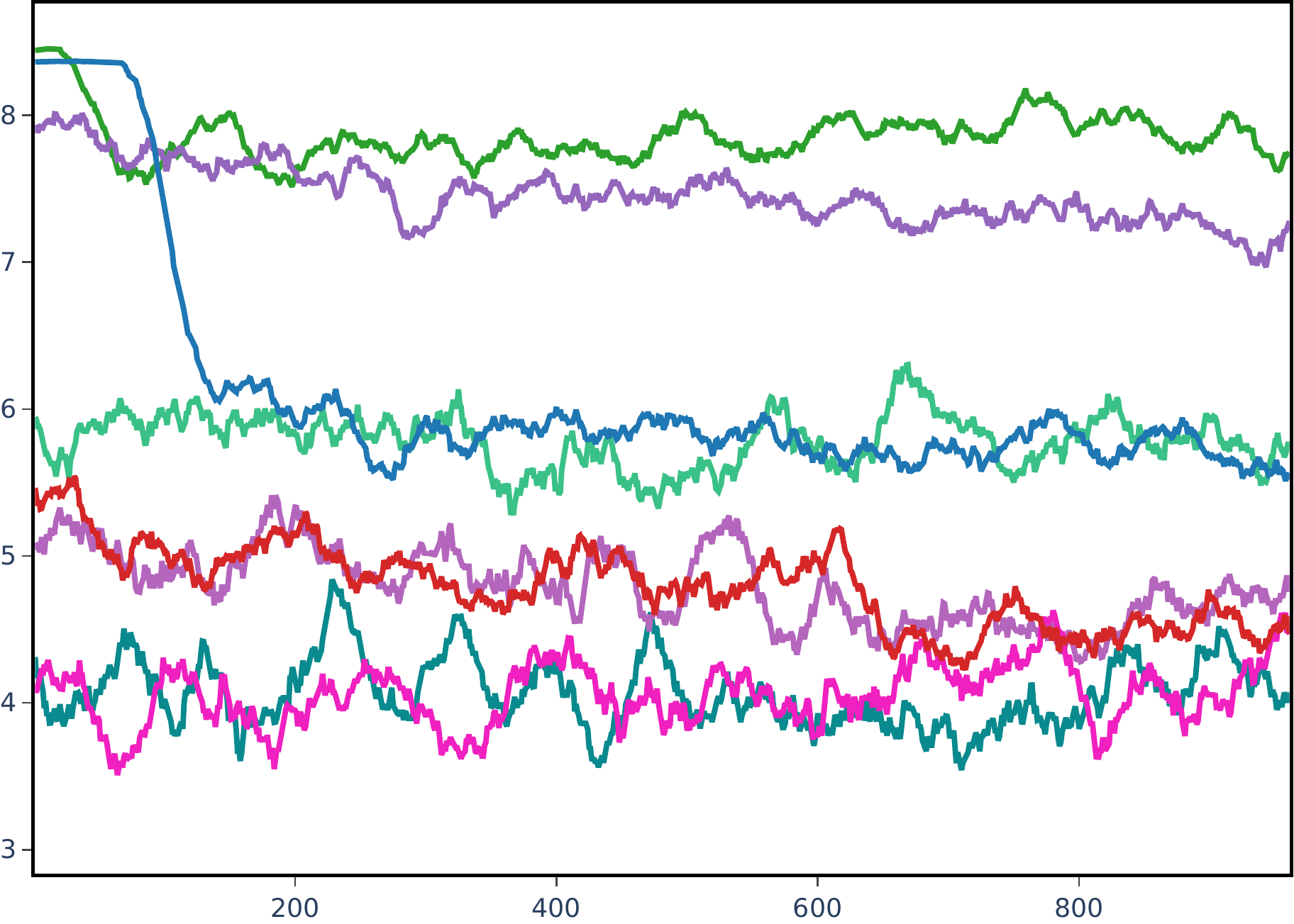} & \includegraphics[scale=0.12]{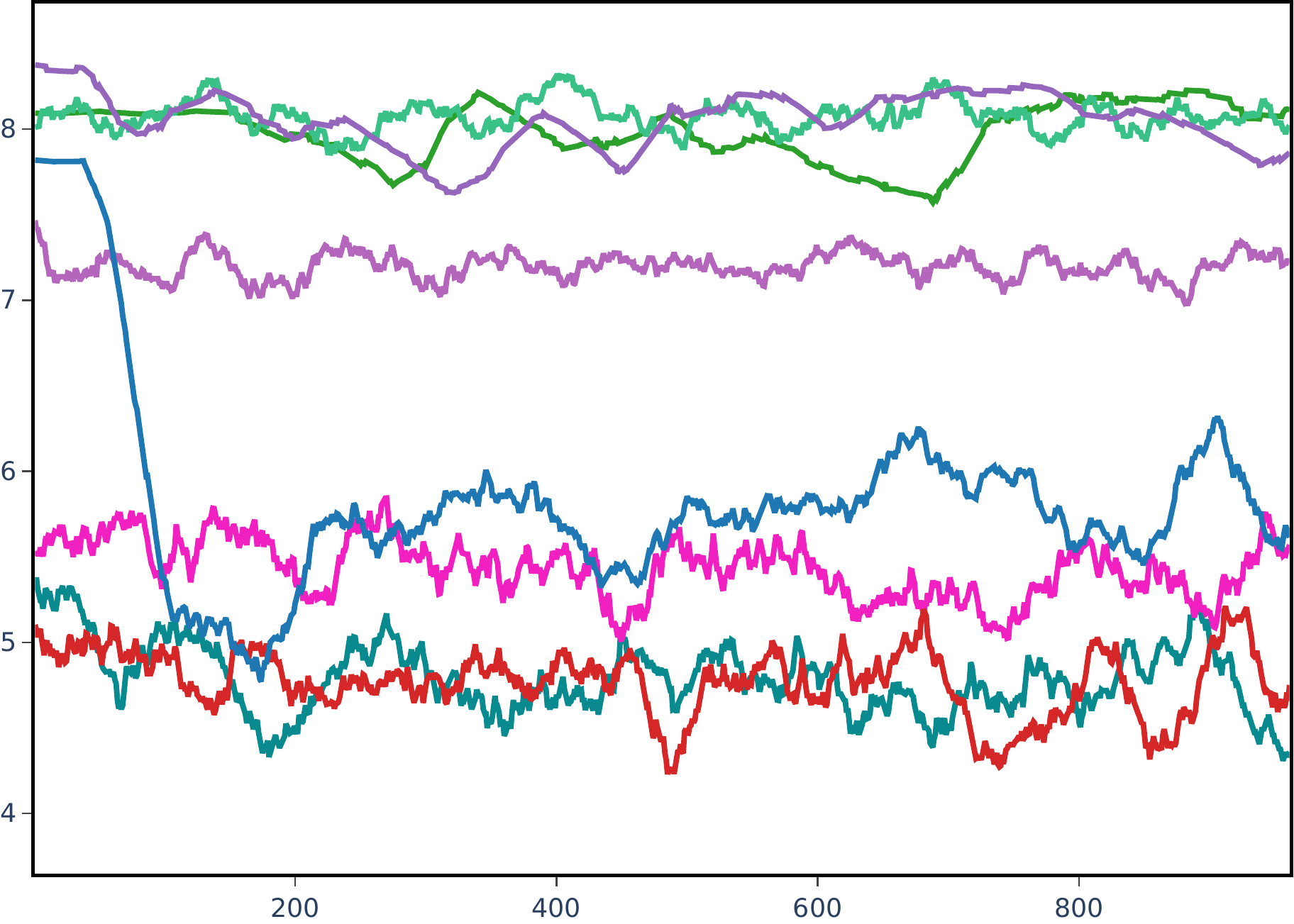} & \includegraphics[scale=0.12]{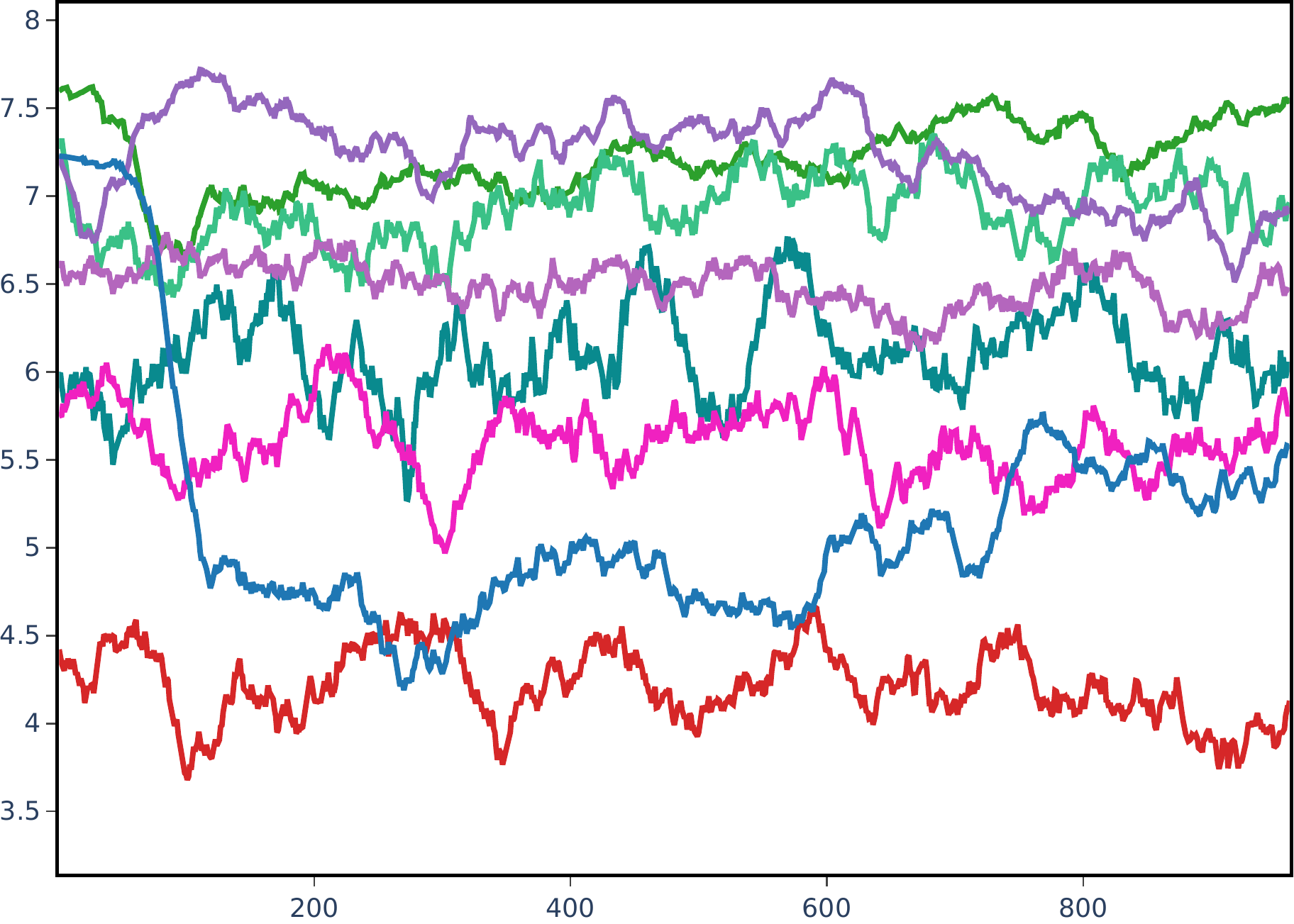} & \includegraphics[scale=0.12]{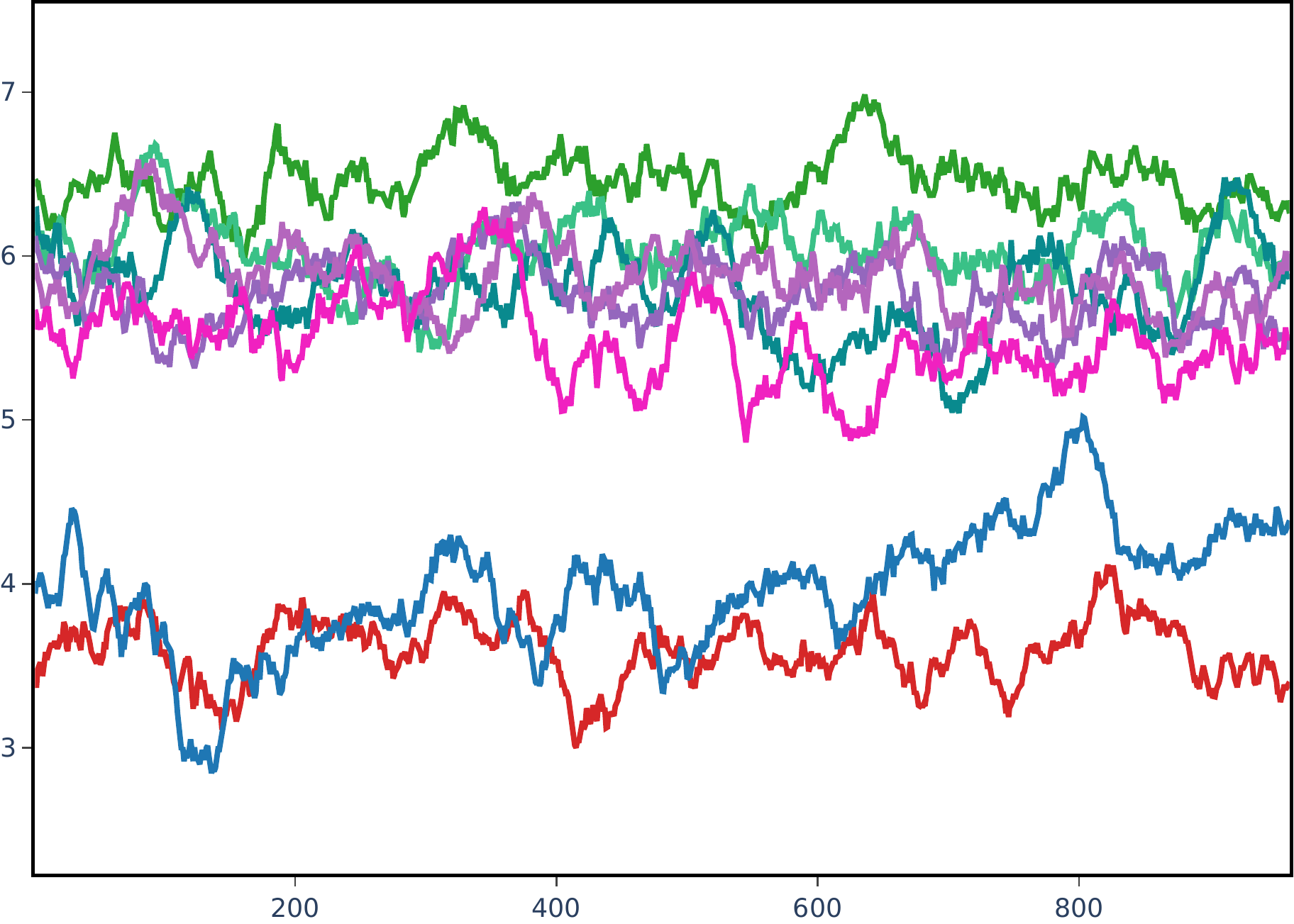} \\
            SVM & \includegraphics[scale=0.12]{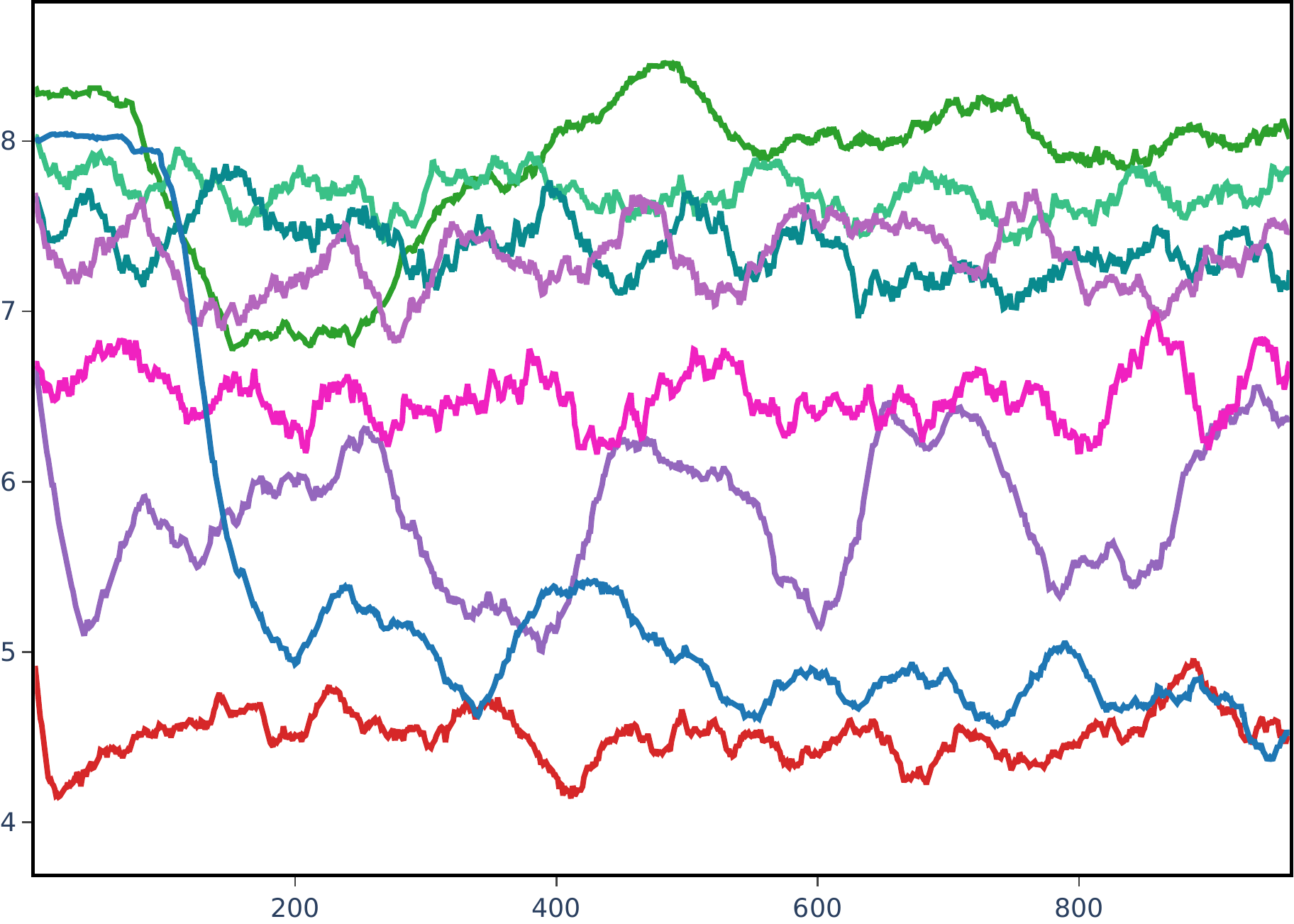}  &  \includegraphics[scale=0.12]{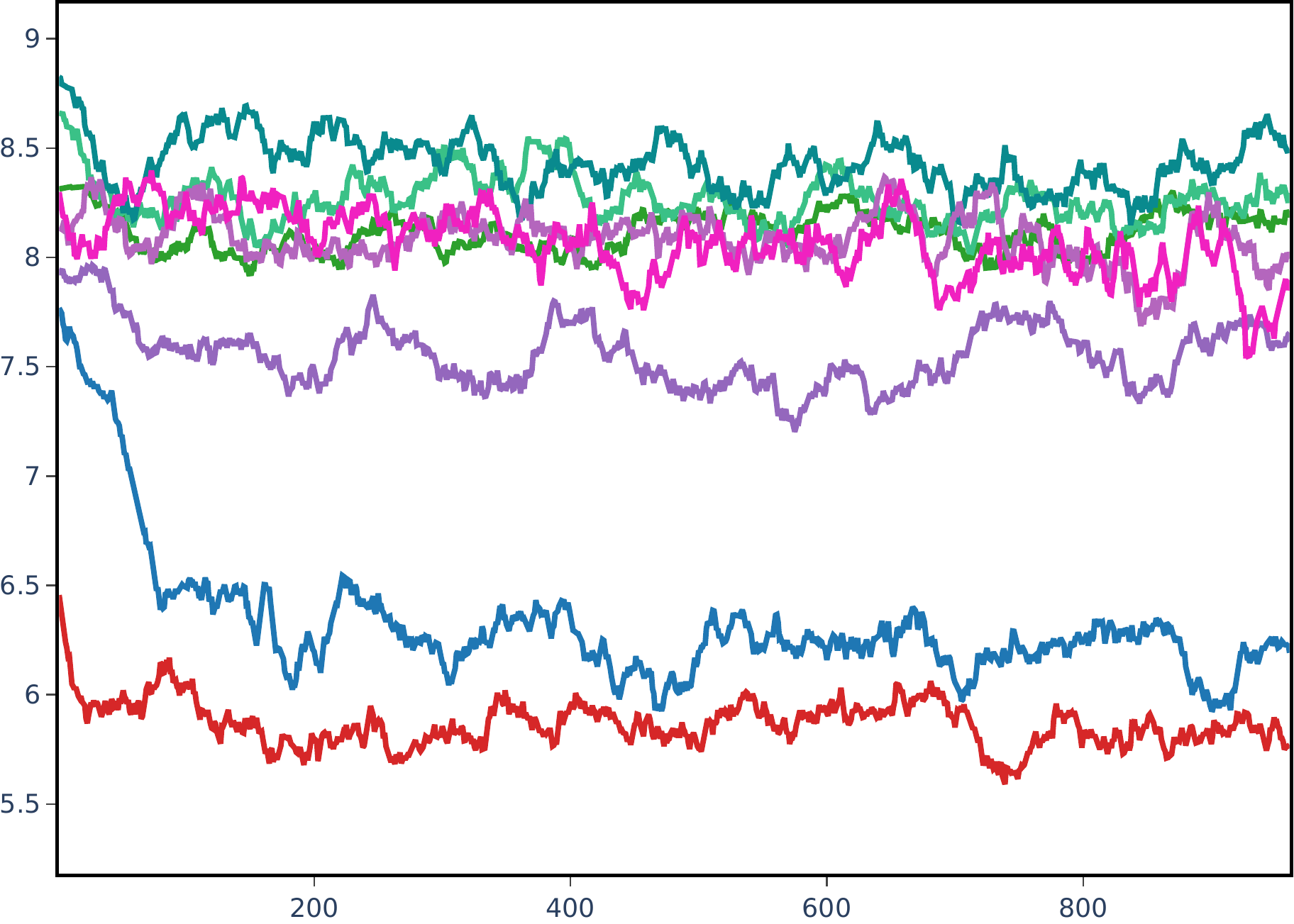} & \includegraphics[scale=0.12]{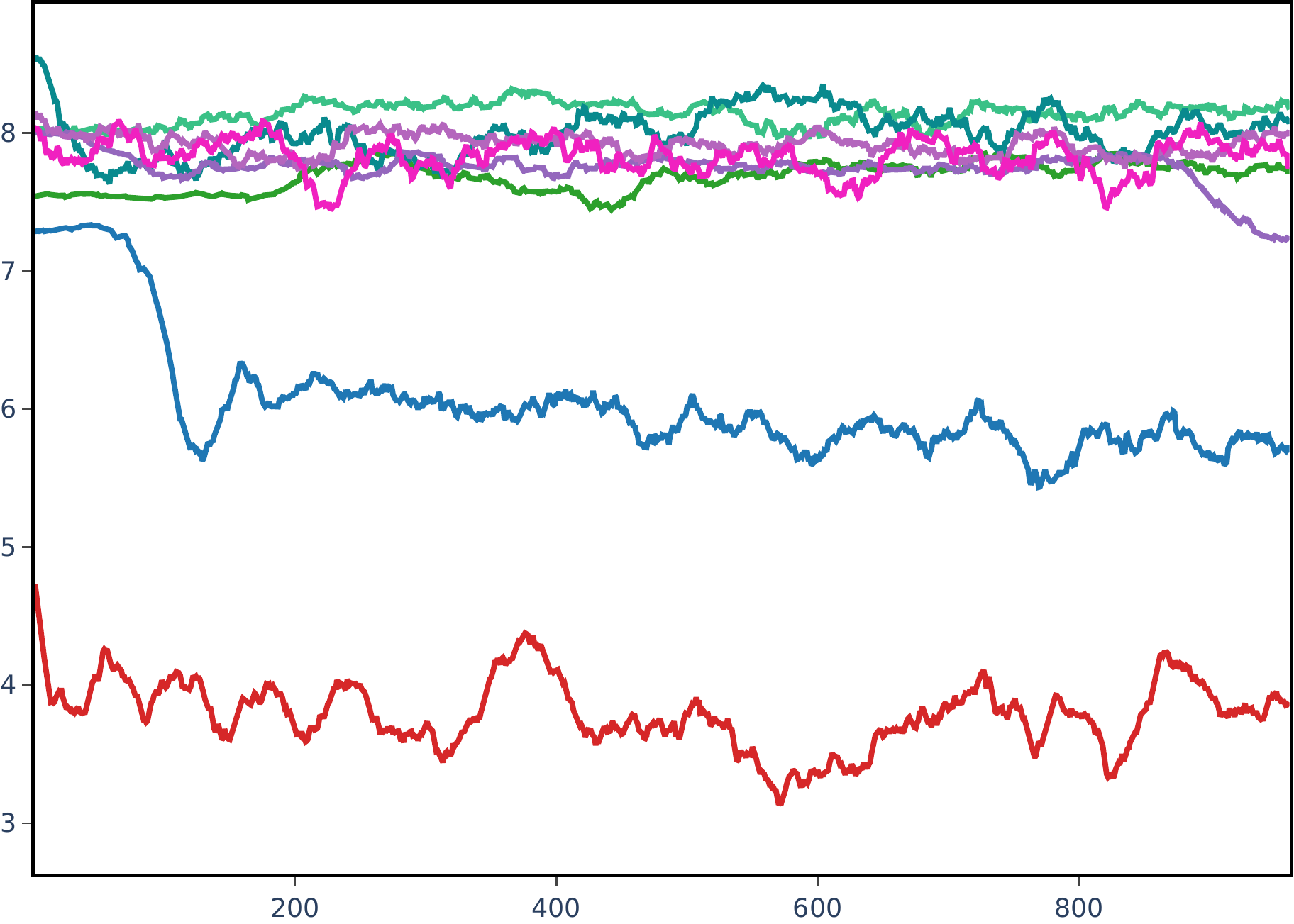} & \includegraphics[scale=0.12]{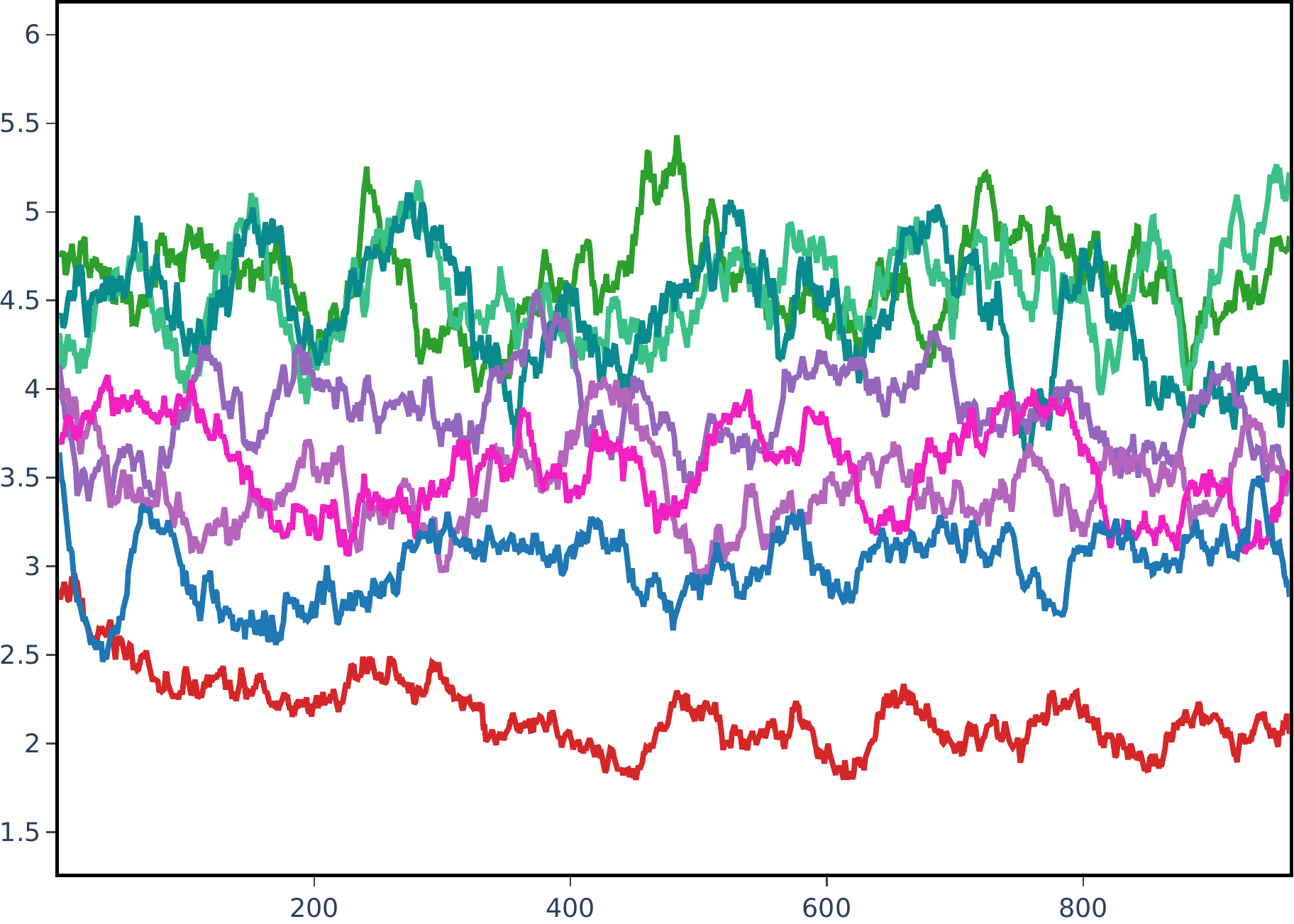} & \includegraphics[scale=0.12]{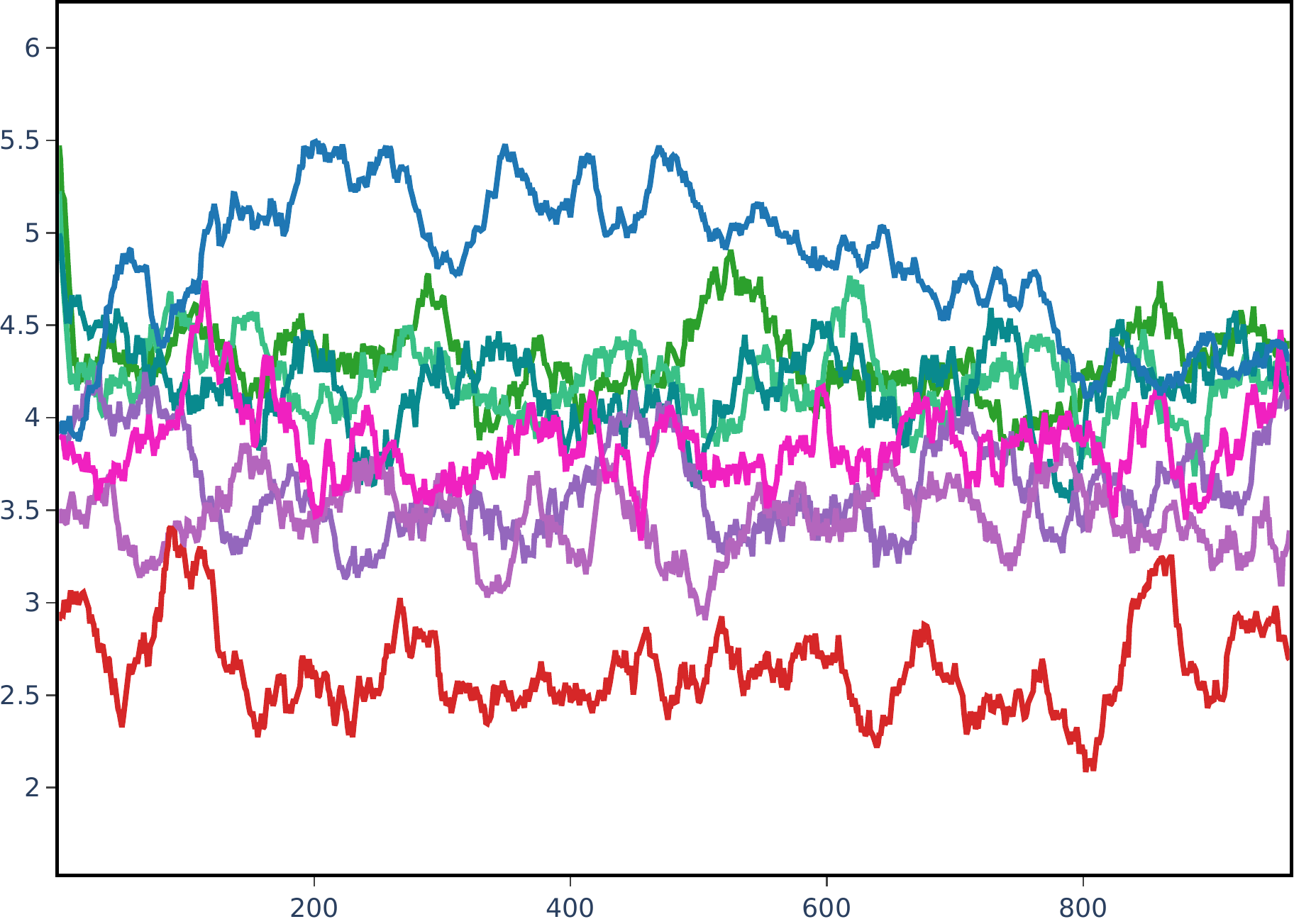}\\
            \midrule
            Legend &  \includegraphics[scale=0.30]{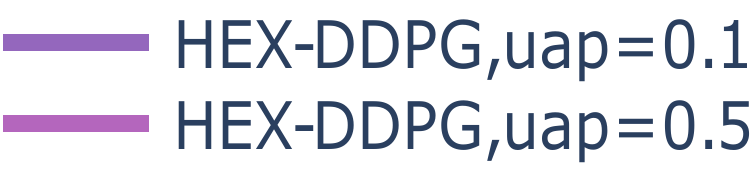} &  \includegraphics[scale=0.30]{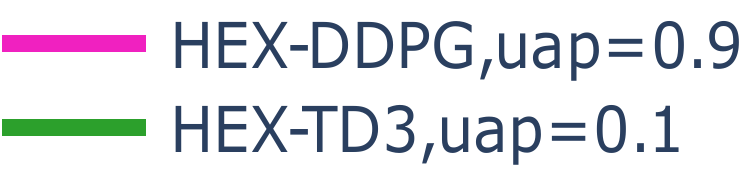} & \includegraphics[scale=0.30]{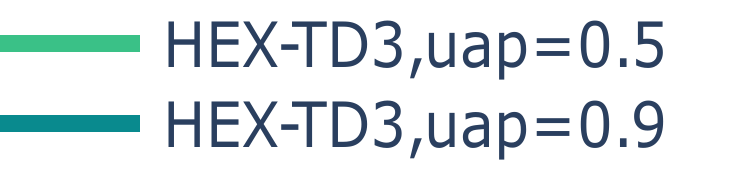} &  \includegraphics[scale=0.30]{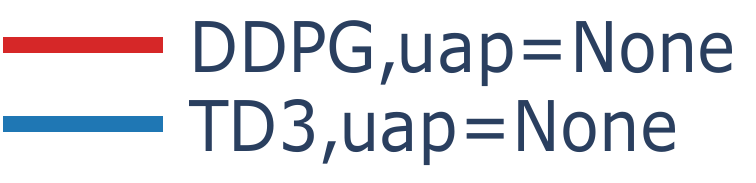} & \\
            \bottomrule
        \end{tabular}
        \caption{Learning curves of each RL method on each dataset and ML model for the HITL decider scenario. Plotted values are per-episode averages over ten trials with a rolling window average (window size $=40$) of these averages plotted.}
        \label{tab:lc-hitl}
    \end{table}
    
First, we notice that DDPG and TD3 have persistently lower learning curves than either of the HEX methods, even across different UAP values. In particular, DDPG has the lowest learning curve with the exception of the MIMIC DT, MIMIC RF, and Movie RF. On the other hand, the highest learning curves tend to be the HEX models with low unacceptable feature percentage (i.e., $UAP=0.10$), although this varies somewhat by dataset and method. Surprisingly, however, even when UAP is high, the HEX models are still relatively comparable and at times even outperform models with lower UAP values. This is particularly encouraging since ideally high quality explanations will still be obtained even when considering a decider's preferred features. This may, however, indicate some level of overfitting, which is better assessed via test set performance, provided and discussed in the next subsection.

\subsubsection{Testing}

We assess test performance of the various explainability methods in two different ways in this simulated, HITL decider scenario. Similar to the decider-free scenario, we first look at average DBD. Second, we examine the average UEP of the various methods to better orient the lens through which we view the average DBD results.

Table \ref{tab:hitl-testing} shows the average DBD across various datasets, ML methods, UAPs, and explainability methods; Table \ref{tab:hitl-test-rewards} in Appendix F shows the average rewards. We \textbf{bolden} the lowest (best) DBD value among our HEX methods by dataset and ML model, also boldening the lowest DBD value among non-HEX methods. If one of the HEX methods has a lower DBD than the non-HEX methods the result is colored \textcolor{red}{red}. If, on the other hand, a non-HEX method has a lower DBD, the result is displayed in \textcolor{blue}{blue}. Note that DDPG, TD3, LIME, and Grow, do not consider a HITL decider. Evaluations are therefore only shown once per dataset and ML method. Further note that because these methods do not consider a HITL decider they return explanations using all features. This is why we also show the average UEP in Table \ref{tab:hitl-perc}, which we discuss next.
\begin{longtable}[!htbp]
{|ccl|ccccc|}
\hline
\multicolumn{3}{|c|}{} & \multicolumn{5}{c|}{Model} \\ \hline
\multicolumn{1}{|c|}{Dataset} & \multicolumn{1}{c|}{UAP} & \multicolumn{1}{c|}{Method} & NN & SVM & RF & LR & DT \\ \hline
\multicolumn{1}{|c|}{} & \multicolumn{1}{c|}{} & HEX-DDPG & 0.305 & \textbf{0.197} & 0.138 & 0.36 & \textbf{0.321} \\
\multicolumn{1}{|c|}{} & \multicolumn{1}{c|}{\multirow{-2}{*}{0.3}} & HEX-TD3 & 0.311 & 0.236 & \textbf{0.137} & 0.379 & 0.331 \\
\multicolumn{1}{|c|}{} & \multicolumn{1}{c|}{} & HEX-DDPG & \textbf{0.303} & 0.202 & 0.15 & \textbf{0.35} & 0.327 \\
\multicolumn{1}{|c|}{} & \multicolumn{1}{c|}{\multirow{-2}{*}{0.5}} & HEX-TD3 & 0.316 & 0.231 & 0.152 & 0.382 & 0.334 \\
\multicolumn{1}{|c|}{} & \multicolumn{1}{c|}{} & HEX-DDPG & 0.335 & 0.237 & 0.181 & 0.372 & 0.334 \\
\multicolumn{1}{|c|}{} & \multicolumn{1}{c|}{\multirow{-2}{*}{0.7}} & HEX-TD3 & 0.343 & 0.258 & 0.179 & 0.397 & 0.338 \\ \cline{2-8} 
\multicolumn{1}{|c|}{} & \multicolumn{1}{c|}{} & DDPG & {\color[HTML]{0070C0} \textbf{0.28}} & {\color[HTML]{0070C0} \textbf{0.173}} & 0.091 & {\color[HTML]{0070C0} \textbf{0.339}} & {\color[HTML]{0070C0} \textbf{0.31}} \\
\multicolumn{1}{|c|}{} & \multicolumn{1}{c|}{} & TD3 & 0.298 & 0.212 & {\color[HTML]{0070C0} \textbf{0.09}} & 0.376 & 0.337 \\
\multicolumn{1}{|c|}{} & \multicolumn{1}{c|}{} & LIME & 0.473 & 0.461 & 0.303 & 0.499 & 0.446 \\
\multicolumn{1}{|c|}{\multirow{-10}{*}{Bank}} & \multicolumn{1}{c|}{\multirow{-4}{*}{None}} & Grow & 0.476 & 0.462 & 0.335 & 0.495 & 0.427 \\ \hline
\multicolumn{1}{|c|}{} & \multicolumn{1}{c|}{} & HEX-DDPG & 0.263 & 0.103 & \textbf{0.053} & {\color[HTML]{FF0000} \textbf{0.209}} & 0.376 \\
\multicolumn{1}{|c|}{} & \multicolumn{1}{c|}{\multirow{-2}{*}{0.3}} & HEX-TD3 & 0.293 & 0.131 & 0.055 & 0.226 & 0.379 \\
\multicolumn{1}{|c|}{} & \multicolumn{1}{c|}{} & HEX-DDPG & {\color[HTML]{FF0000} \textbf{0.247}} & {\color[HTML]{FF0000} \textbf{0.081}} & 0.066 & 0.22 & 0.376 \\
\multicolumn{1}{|c|}{} & \multicolumn{1}{c|}{\multirow{-2}{*}{0.5}} & HEX-TD3 & 0.268 & 0.104 & 0.065 & 0.231 & 0.372 \\
\multicolumn{1}{|c|}{} & \multicolumn{1}{c|}{} & HEX-DDPG & 0.254 & 0.09 & 0.056 & 0.221 & {\color[HTML]{FF0000} \textbf{0.366}} \\
\multicolumn{1}{|c|}{} & \multicolumn{1}{c|}{\multirow{-2}{*}{0.7}} & HEX-TD3 & 0.265 & 0.118 & 0.059 & 0.221 & {\color[HTML]{FF0000} \textbf{0.366}} \\ \cline{2-8} 
\multicolumn{1}{|c|}{} & \multicolumn{1}{c|}{} & DDPG & \textbf{0.271} & \textbf{0.104} & {\color[HTML]{0070C0} \textbf{0.046}} & \textbf{0.215} & \textbf{0.386} \\
\multicolumn{1}{|c|}{} & \multicolumn{1}{c|}{} & TD3 & 0.308 & 0.119 & 0.055 & 0.251 & 0.392 \\
\multicolumn{1}{|c|}{} & \multicolumn{1}{c|}{} & LIME & 0.427 & 0.338 & 0.154 & 0.426 & 0.321 \\
\multicolumn{1}{|c|}{\multirow{-10}{*}{MIMIC}} & \multicolumn{1}{c|}{\multirow{-4}{*}{None}} & Grow & 0.483 & 0.404 & 0.236 & 0.489 & 0.342 \\ \hline
\multicolumn{1}{|c|}{} & \multicolumn{1}{c|}{} & HEX-DDPG & 0.183 & 0.297 & 0.082 & 0.419 & 0.439 \\
\multicolumn{1}{|c|}{} & \multicolumn{1}{c|}{\multirow{-2}{*}{0.3}} & HEX-TD3 & 0.203 & 0.337 & 0.076 & 0.435 & 0.439 \\
\multicolumn{1}{|c|}{} & \multicolumn{1}{c|}{} & HEX-DDPG & 0.173 & 0.288 & 0.077 & 0.416 & 0.437 \\
\multicolumn{1}{|c|}{} & \multicolumn{1}{c|}{\multirow{-2}{*}{0.5}} & HEX-TD3 & 0.199 & 0.321 & 0.081 & 0.43 & 0.439 \\
\multicolumn{1}{|c|}{} & \multicolumn{1}{c|}{} & HEX-DDPG & {\color[HTML]{FF0000} \textbf{0.172}} & {\color[HTML]{FF0000} \textbf{0.245}} & {\color[HTML]{FF0000} \textbf{0.065}} & {\color[HTML]{FF0000} \textbf{0.388}} & {\color[HTML]{FF0000} \textbf{0.423}} \\
\multicolumn{1}{|c|}{} & \multicolumn{1}{c|}{\multirow{-2}{*}{0.7}} & HEX-TD3 & 0.186 & 0.286 & 0.066 & 0.406 & 0.424 \\ \cline{2-8} 
\multicolumn{1}{|c|}{} & \multicolumn{1}{c|}{} & DDPG & \textbf{0.24} & \textbf{0.316} & 0.076 & \textbf{0.454} & 0.438 \\
\multicolumn{1}{|c|}{} & \multicolumn{1}{c|}{} & TD3 & 0.247 & 0.356 & \textbf{0.072} & 0.457 & \textbf{0.433} \\
\multicolumn{1}{|c|}{} & \multicolumn{1}{c|}{} & LIME & 0.462 & 0.5 & 0.314 & 0.5 & 0.49 \\
\multicolumn{1}{|c|}{\multirow{-10}{*}{Movie}} & \multicolumn{1}{c|}{\multirow{-4}{*}{None}} & Grow & 0.45 & 0.416 & 0.217 & 0.49 & 0.331 \\ \hline
\multicolumn{1}{|c|}{} & \multicolumn{1}{c|}{} & HEX-DDPG & 0.195 & 0.205 & {\color[HTML]{FF0000} \textbf{0.084}} & 0.228 & 0.366 \\
\multicolumn{1}{|c|}{} & \multicolumn{1}{c|}{\multirow{-2}{*}{0.3}} & HEX-TD3 & 0.195 & 0.205 & 0.087 & 0.229 & 0.366 \\
\multicolumn{1}{|c|}{} & \multicolumn{1}{c|}{} & HEX-DDPG & {\color[HTML]{FF0000} \textbf{0.193}} & 0.206 & 0.089 & 0.22 & \textbf{0.341} \\
\multicolumn{1}{|c|}{} & \multicolumn{1}{c|}{\multirow{-2}{*}{0.5}} & HEX-TD3 & 0.195 & \textbf{0.203} & 0.086 & {\color[HTML]{FF0000} \textbf{0.218}} & 0.342 \\
\multicolumn{1}{|c|}{} & \multicolumn{1}{c|}{} & HEX-DDPG & 0.207 & 0.21 & 0.097 & 0.249 & 0.336 \\
\multicolumn{1}{|c|}{} & \multicolumn{1}{c|}{\multirow{-2}{*}{0.7}} & HEX-TD3 & 0.209 & 0.212 & 0.096 & 0.249 & 0.338 \\ \cline{2-8} 
\multicolumn{1}{|c|}{} & \multicolumn{1}{c|}{} & DDPG & 0.246 & {\color[HTML]{0070C0} \textbf{0.176}} & \textbf{0.092} & 0.374 & 0.381 \\
\multicolumn{1}{|c|}{} & \multicolumn{1}{c|}{} & TD3 & \textbf{0.239} & 0.178 & \textbf{0.092} & 0.371 & 0.385 \\
\multicolumn{1}{|c|}{} & \multicolumn{1}{c|}{} & LIME & 0.31 & 0.195 & 0.101 & \textbf{0.37} & {\color[HTML]{0070C0} \textbf{0.225}} \\
\multicolumn{1}{|c|}{\multirow{-10}{*}{Student}} & \multicolumn{1}{c|}{\multirow{-4}{*}{None}} & Grow & 0.42 & 0.304 & 0.126 & 0.449 & 0.234 \\ \hline
\multicolumn{1}{|c|}{} & \multicolumn{1}{c|}{} & HEX-DDPG & 0.384 & 0.373 & \textbf{0.065} & 0.285 & 0.488 \\
\multicolumn{1}{|c|}{} & \multicolumn{1}{c|}{\multirow{-2}{*}{0.3}} & HEX-TD3 & 0.387 & 0.371 & 0.067 & \textbf{0.273} & 0.49 \\
\multicolumn{1}{|c|}{} & \multicolumn{1}{c|}{} & HEX-DDPG & \textbf{0.381} & 0.363 & 0.073 & 0.286 & 0.477 \\
\multicolumn{1}{|c|}{} & \multicolumn{1}{c|}{\multirow{-2}{*}{0.5}} & HEX-TD3 & 0.387 & 0.368 & 0.072 & 0.282 & 0.478 \\
\multicolumn{1}{|c|}{} & \multicolumn{1}{c|}{} & HEX-DDPG & \textbf{0.381} & {\color[HTML]{FF0000} \textbf{0.35}} & 0.073 & 0.288 & 0.476 \\
\multicolumn{1}{|c|}{} & \multicolumn{1}{c|}{\multirow{-2}{*}{0.7}} & HEX-TD3 & 0.383 & 0.356 & 0.072 & 0.283 & {\color[HTML]{FF0000} \textbf{0.475}} \\ \cline{2-8} 
\multicolumn{1}{|c|}{} & \multicolumn{1}{c|}{} & DDPG & {\color[HTML]{0070C0} \textbf{0.367}} & \textbf{0.425} & {\color[HTML]{0070C0} \textbf{0.05}} & {\color[HTML]{0070C0} \textbf{0.26}} & \textbf{0.493} \\
\multicolumn{1}{|c|}{} & \multicolumn{1}{c|}{} & TD3 & 0.387 & 0.429 & 0.056 & 0.3 & 0.496 \\
\multicolumn{1}{|c|}{} & \multicolumn{1}{c|}{} & LIME & 0.496 & 0.448 & 0.294 & 0.494 & 0.498 \\
\multicolumn{1}{|c|}{\multirow{-10}{*}{News}} & \multicolumn{1}{c|}{\multirow{-4}{*}{None}} & Grow & 0.497 & 0.482 & 0.271 & 0.483 & 0.436 \\ \hline
\caption{Average DBD for our HITL decider scenario. \textbf{Bold} indicates the lowest DBD among all HEX methods and all non-HEX method on each dataset and ML model and all (HEX/non-HEX considered separately). \textbf{\textcolor{blue}{Blue}} indicates that a non-HEX method received the lowest overall within-ML method and dataset DBD, while \textbf{\textcolor{red}{red}} indicates that a HEX method received the lowest overall within-method and dataset DBD.\label{tab:hitl-testing}}
\end{longtable}

First, we are surprised how often one of the HEX methods outperforms the non-HEX methods since the non-HEX methods are free to use the entire feature space to produce explanations, while the HEX methods use 0-distrust projection and are thus limited to certain features. In fact, of the 25 model-dataset combinations, one of the HEX models performs better on 14. Second, among the various HEX results, we are surprised that the lowest UAP (UAP$=0.10$) does not always produce the lowest DBD result. In fact, of the 25 model-dataset experiments, a UAP$=0.10$ model only produces the lowest within-HEX DBD value on 8 of these. This finding suggests that, at times, more human-interpretable solutions may naturally produce better explanatory points.

To further contextualize and orient the lens through which the DBD results of Table \ref{tab:hitl-testing} are viewed, we provide Table \ref{tab:hitl-perc}, below, which shows the average UEP used to generate explanations. We show the average UEP of each non-HEX method for each UAP scenario generated. The bottom of the table shows the HEX method results for all datasets and UAP values. Since the HEX methods use 0-distrust projection, they never use unacceptable features (i.e., these values are always 0).
\begin{longtable}[!htbp]
{|ccl|ccccc|}
\hline
\multicolumn{3}{|c|}{} & \multicolumn{5}{c|}{Model} \\ \hline
\multicolumn{1}{|l|}{Dataset} & \multicolumn{1}{c|}{UAP} & Method & NN & SVM & RF & LR & DT \\ \hline
\multicolumn{1}{|c|}{} & \multicolumn{1}{c|}{} & DDPG & 0.84 & 0.84 & 0.836 & 0.835 & 0.835 \\
\multicolumn{1}{|c|}{} & \multicolumn{1}{c|}{} & TD3 & 0.834 & 0.836 & 0.836 & 0.838 & 0.837 \\
\multicolumn{1}{|c|}{} & \multicolumn{1}{c|}{} & LIME & 0.572 & 0.572 & 0.572 & 0.572 & 0.572 \\
\multicolumn{1}{|c|}{} & \multicolumn{1}{c|}{\multirow{-4}{*}{0.3}} & Grow & 0.744 & 0.744 & 0.744 & 0.744 & 0.744 \\ \cline{2-8} 
\multicolumn{1}{|c|}{} & \multicolumn{1}{c|}{} & DDPG & 0.757 & 0.766 & 0.759 & 0.755 & 0.755 \\
\multicolumn{1}{|c|}{} & \multicolumn{1}{c|}{} & TD3 & 0.755 & 0.762 & 0.757 & 0.761 & 0.759 \\
\multicolumn{1}{|c|}{} & \multicolumn{1}{c|}{} & LIME & 0.651 & 0.65 & 0.65 & 0.65 & 0.65 \\
\multicolumn{1}{|c|}{} & \multicolumn{1}{c|}{\multirow{-4}{*}{0.5}} & Grow & 0.616 & 0.616 & 0.616 & 0.616 & 0.616 \\ \cline{2-8} 
\multicolumn{1}{|c|}{} & \multicolumn{1}{c|}{} & DDPG & 0.7 & 0.703 & 0.693 & 0.69 & 0.69 \\
\multicolumn{1}{|c|}{} & \multicolumn{1}{c|}{} & TD3 & 0.7 & 0.702 & 0.699 & 0.703 & 0.7 \\
\multicolumn{1}{|c|}{} & \multicolumn{1}{c|}{} & LIME & 0.763 & 0.763 & 0.763 & 0.763 & 0.763 \\
\multicolumn{1}{|c|}{\multirow{-12}{*}{Bank}} & \multicolumn{1}{c|}{\multirow{-4}{*}{0.7}} & Grow & 0.518 & 0.518 & 0.518 & 0.518 & 0.518 \\ \hline
\multicolumn{1}{|c|}{} & \multicolumn{1}{c|}{} & DDPG & 0.762 & 0.763 & 0.764 & 0.763 & 0.764 \\
\multicolumn{1}{|c|}{} & \multicolumn{1}{c|}{} & TD3 & 0.766 & 0.765 & 0.765 & 0.764 & 0.765 \\
\multicolumn{1}{|c|}{} & \multicolumn{1}{c|}{} & LIME & 0.517 & 0.517 & 0.516 & 0.517 & 0.516 \\
\multicolumn{1}{|c|}{} & \multicolumn{1}{c|}{\multirow{-4}{*}{0.3}} & Grow & 0.754 & 0.754 & 0.754 & 0.754 & 0.754 \\ \cline{2-8} 
\multicolumn{1}{|c|}{} & \multicolumn{1}{c|}{} & DDPG & 0.612 & 0.613 & 0.613 & 0.613 & 0.613 \\
\multicolumn{1}{|c|}{} & \multicolumn{1}{c|}{} & TD3 & 0.612 & 0.613 & 0.612 & 0.612 & 0.613 \\
\multicolumn{1}{|c|}{} & \multicolumn{1}{c|}{} & LIME & 0.6 & 0.601 & 0.6 & 0.6 & 0.6 \\
\multicolumn{1}{|c|}{} & \multicolumn{1}{c|}{\multirow{-4}{*}{0.5}} & Grow & 0.6 & 0.6 & 0.6 & 0.6 & 0.6 \\ \cline{2-8} 
\multicolumn{1}{|c|}{} & \multicolumn{1}{c|}{} & DDPG & 0.544 & 0.546 & 0.546 & 0.546 & 0.545 \\
\multicolumn{1}{|c|}{} & \multicolumn{1}{c|}{} & TD3 & 0.541 & 0.542 & 0.542 & 0.543 & 0.543 \\
\multicolumn{1}{|c|}{} & \multicolumn{1}{c|}{} & LIME & 0.754 & 0.754 & 0.754 & 0.754 & 0.754 \\
\multicolumn{1}{|c|}{\multirow{-12}{*}{MIMIC}} & \multicolumn{1}{c|}{\multirow{-4}{*}{0.7}} & Grow & 0.515 & 0.515 & 0.515 & 0.515 & 0.515 \\ \hline
\multicolumn{1}{|c|}{} & \multicolumn{1}{c|}{} & DDPG & 0.8 & 0.797 & 0.799 & 0.797 & 0.797 \\
\multicolumn{1}{|c|}{} & \multicolumn{1}{c|}{} & TD3 & 0.817 & 0.806 & 0.806 & 0.805 & 0.805 \\
\multicolumn{1}{|c|}{} & \multicolumn{1}{c|}{} & LIME & 0.539 & 0.539 & 0.539 & 0.539 & 0.538 \\
\multicolumn{1}{|c|}{} & \multicolumn{1}{c|}{\multirow{-4}{*}{0.3}} & Grow & 0.742 & 0.742 & 0.742 & 0.742 & 0.742 \\ \cline{2-8} 
\multicolumn{1}{|c|}{} & \multicolumn{1}{c|}{} & DDPG & 0.701 & 0.697 & 0.698 & 0.693 & 0.693 \\
\multicolumn{1}{|c|}{} & \multicolumn{1}{c|}{} & TD3 & 0.725 & 0.709 & 0.707 & 0.706 & 0.705 \\
\multicolumn{1}{|c|}{} & \multicolumn{1}{c|}{} & LIME & 0.633 & 0.633 & 0.634 & 0.634 & 0.633 \\
\multicolumn{1}{|c|}{} & \multicolumn{1}{c|}{\multirow{-4}{*}{0.5}} & Grow & 0.613 & 0.613 & 0.613 & 0.613 & 0.613 \\ \cline{2-8} 
\multicolumn{1}{|c|}{} & \multicolumn{1}{c|}{} & DDPG & 0.618 & 0.615 & 0.617 & 0.611 & 0.612 \\
\multicolumn{1}{|c|}{} & \multicolumn{1}{c|}{} & TD3 & 0.651 & 0.633 & 0.631 & 0.63 & 0.627 \\
\multicolumn{1}{|c|}{} & \multicolumn{1}{c|}{} & LIME & 0.755 & 0.754 & 0.755 & 0.754 & 0.755 \\
\multicolumn{1}{|c|}{\multirow{-12}{*}{Movie}} & \multicolumn{1}{c|}{\multirow{-4}{*}{0.7}} & Grow & 0.512 & 0.512 & 0.512 & 0.512 & 0.512 \\ \hline
\multicolumn{1}{|c|}{} & \multicolumn{1}{c|}{} & DDPG & 0.842 & 0.842 & 0.841 & 0.842 & 0.843 \\
\multicolumn{1}{|c|}{} & \multicolumn{1}{c|}{} & TD3 & 0.823 & 0.83 & 0.828 & 0.827 & 0.83 \\
\multicolumn{1}{|c|}{} & \multicolumn{1}{c|}{} & LIME & 0.523 & 0.523 & 0.523 & 0.523 & 0.523 \\
\multicolumn{1}{|c|}{} & \multicolumn{1}{c|}{\multirow{-4}{*}{0.3}} & Grow & 0.756 & 0.756 & 0.756 & 0.756 & 0.756 \\ \cline{2-8} 
\multicolumn{1}{|c|}{} & \multicolumn{1}{c|}{} & DDPG & 0.745 & 0.745 & 0.747 & 0.748 & 0.75 \\
\multicolumn{1}{|c|}{} & \multicolumn{1}{c|}{} & TD3 & 0.722 & 0.737 & 0.734 & 0.735 & 0.736 \\
\multicolumn{1}{|c|}{} & \multicolumn{1}{c|}{} & LIME & 0.651 & 0.651 & 0.651 & 0.651 & 0.651 \\
\multicolumn{1}{|c|}{} & \multicolumn{1}{c|}{\multirow{-4}{*}{0.5}} & Grow & 0.616 & 0.616 & 0.616 & 0.616 & 0.616 \\ \cline{2-8} 
\multicolumn{1}{|c|}{} & \multicolumn{1}{c|}{} & DDPG & 0.67 & 0.67 & 0.67 & 0.671 & 0.673 \\
\multicolumn{1}{|c|}{} & \multicolumn{1}{c|}{} & TD3 & 0.642 & 0.654 & 0.648 & 0.649 & 0.651 \\
\multicolumn{1}{|c|}{} & \multicolumn{1}{c|}{} & LIME & 0.77 & 0.77 & 0.77 & 0.77 & 0.77 \\
\multicolumn{1}{|c|}{\multirow{-12}{*}{Student}} & \multicolumn{1}{c|}{\multirow{-4}{*}{0.7}} & Grow & 0.493 & 0.493 & 0.493 & 0.493 & 0.493 \\ \hline
\multicolumn{1}{|c|}{} & \multicolumn{1}{c|}{} & DDPG & 0.869 & 0.844 & 0.845 & 0.846 & 0.84 \\
\multicolumn{1}{|c|}{} & \multicolumn{1}{c|}{} & TD3 & 0.844 & 0.839 & 0.839 & 0.839 & 0.838 \\
\multicolumn{1}{|c|}{} & \multicolumn{1}{c|}{} & LIME & 0.491 & 0.491 & 0.491 & 0.491 & 0.491 \\
\multicolumn{1}{|c|}{} & \multicolumn{1}{c|}{\multirow{-4}{*}{0.3}} & Grow & 0.744 & 0.744 & 0.744 & 0.744 & 0.744 \\ \cline{2-8} 
\multicolumn{1}{|c|}{} & \multicolumn{1}{c|}{} & DDPG & 0.793 & 0.756 & 0.759 & 0.761 & 0.752 \\
\multicolumn{1}{|c|}{} & \multicolumn{1}{c|}{} & TD3 & 0.766 & 0.754 & 0.753 & 0.754 & 0.752 \\
\multicolumn{1}{|c|}{} & \multicolumn{1}{c|}{} & LIME & 0.61 & 0.61 & 0.61 & 0.61 & 0.61 \\
\multicolumn{1}{|c|}{} & \multicolumn{1}{c|}{\multirow{-4}{*}{0.5}} & Grow & 0.61 & 0.61 & 0.61 & 0.61 & 0.61 \\ \cline{2-8} 
\multicolumn{1}{|c|}{} & \multicolumn{1}{c|}{} & DDPG & 0.73 & 0.684 & 0.688 & 0.69 & 0.678 \\
\multicolumn{1}{|c|}{} & \multicolumn{1}{c|}{} & TD3 & 0.69 & 0.676 & 0.676 & 0.678 & 0.676 \\
\multicolumn{1}{|c|}{} & \multicolumn{1}{c|}{} & LIME & 0.744 & 0.744 & 0.744 & 0.744 & 0.744 \\
\multicolumn{1}{|c|}{\multirow{-12}{*}{News}} & \multicolumn{1}{c|}{\multirow{-4}{*}{0.7}} & Grow & 0.491 & 0.491 & 0.491 & 0.491 & 0.491 \\ \hline
\multicolumn{1}{|l|}{All} & \multicolumn{1}{l|}{All} & HEX-DDPG & {\color[HTML]{FF0000} \textbf{0}} & {\color[HTML]{FF0000} \textbf{0}} & {\color[HTML]{FF0000} \textbf{0}} & {\color[HTML]{FF0000} \textbf{0}} & {\color[HTML]{FF0000} \textbf{0}} \\
\multicolumn{1}{|l|}{All} & \multicolumn{1}{l|}{All} & HEX-TD3 & {\color[HTML]{FF0000} \textbf{0}} & {\color[HTML]{FF0000} \textbf{0}} & {\color[HTML]{FF0000} \textbf{0}} & {\color[HTML]{FF0000} \textbf{0}} & {\color[HTML]{FF0000} \textbf{0}} \\ \hline
\caption{Average UEP by dataset, ML model, and UAP scenario.\label{tab:hitl-perc}}
\end{longtable}
First, Table \ref{tab:hitl-perc} shows that the non-HEX methods rely heavily on the UAP-selected features, highlighting the benefits of our proposed HEX methods. Second, a curious artifact discovered in presenting and examining Table \ref{tab:hitl-perc} is the uniformity of feature reliance in generating explanations across different machine learning models among both LIME and Growing Spheres -- i.e., the UEP is consistent across ML models within dataset and UAP scenario. This result is unsurprising, however, provided how these methods operate. While both LIME and Growing Spheres are model agnostic, they do not explicitly consider the ML model when producing explanations, instead relying on the underlying training data. This finding highlights two things. First, the lack of explicit model consideration is a short-coming of these methods since explanatory models should explain the predictions made by an ML model, rather than the training data. Second, in a federated learning or other limited or reduced training data availability scenario, exploration of the ML model-learned probability space is key to reliably explaining the model. We suspect that UEP uniformity of LIME and Growing Spheres across ML models is partially attributable to the limited, underlying training data and lack of probability space exploration.

\section{Conclusions}
In this work we propose a human-in-the-loop (HITL) model-agnostic classification explainability method, adopting a markov decision process (MDP) approach, thus permitting the use of deep reinforcement learning (DRL) methods to synthesize explanatory policies. We augment two state-of-the-art DRL methods, DDPG and TD3, with both HITL and federated learning considerations, further updating the methods based on the relationship we prove to exist between buffer and policy degeneracy. We compare our proposed methods, HEX-DDPG and HEX-TD3, against out-of-the-box DRL methods and two competing state-of-the-art explainability methods, LIME and Growing Spheres. We conduct these evaluations across two different scenarios, one that does not consider a HITL and another that explicitly considers a HITL decider. Both scenarios consider five different well-known machine learning models each of which are trained on five different decision-focused datasets spanning business, medicine, and education. Our experiments show that HEX-DDPG and HEX-TD3 are competitive with and often outperform both baseline DRL methods and competing state-of-the-art explainability methods. Furthermore, our methods frequently outperform both baseline DRL methods, LIME, and Growing Spheres in our HITL decider scenario, even using our proposed $0$-distrust projection method, which restricts explanations to using only decider-trusted features.

\section*{Appendix A: Proof of Theorem \ref{thm:bd-pd} \label{sec:thm}}

\thispagestyle{empty}
\setcounter{equation}{0} %
\renewcommand{\theequation}{A.\arabic{equation}}
\setcounter{table}{0}
\renewcommand{\thetable}{A.\arabic{table}}
\textbf{Proof sketch}: The theorem is of the form
\begin{align}
    A > B \implies A > B^{\prime}.
\end{align}
By showing that $B^{\prime}$ is directly derived from $B$ -- i.e.,~$B \implies B^{\prime}$ -- we show $ \left ( A > B \cap B \implies B^{\prime} \right) \implies A > B^{\prime}$ to be true vacuously. The proof is below.
\begin{proof}
Assume Buffer Degeneracy holds.
Let $\pi^{\theta^*}_l = \argmin_{\pi}\{ \mathbb{E}_{\mathbf{x} \sim \rho^{\pi}_l, \mathbf{z} \sim \pi^{\theta}, r \sim R} \left[\left(Q \left(\mathbf{x},\mathbf{z} \vert \theta^{Q} \right) - y \right)^2 \right] \}$ and $\pi^{\theta^*}_{l^{\prime}} = \argmin_{\pi}\{ \mathbb{E}_{\mathbf{x} \sim \rho^{\pi}_{l^{\prime}}, \mathbf{z} \sim \pi^{\theta}, r \sim R} \left[\left(Q \left(\mathbf{x},\mathbf{z} \vert \theta^{Q} \right) - y \right)^2 \right] \}$. Let $\rho^{y}_l = \{y:\mathbf{x} \sim \rho^{\pi}_l, \mathbf{z} \sim \pi^{\theta^*}_l, r \sim R\}$ and $\tilde{\rho}^{y}_{l^{\prime}} = \{y: \mathbf{x} \sim \rho^{\pi}_{l}, \mathbf{z} \sim \pi^{\theta^*}_{l^{\prime}}, r \sim R\}$. Since buffer degeneracy holds and $\pi^{\theta^*}_{l^{\prime}}$ is the minimizer of the loss on the degenerate buffer, then
\begin{align}
    \mathbb{E}_{y_l \sim \rho^{y}_l}\left[y_l\right] > & \mathbb{E}_{\tilde{y}_{l^{\prime}} \sim \tilde{\rho}^{y}_{l^{\prime}}}\left[\tilde{y}_{l^{\prime}}\right]: l^{\prime} > l,
\end{align}
as desired.
\end{proof}

\section*{Appendix B: Dataset Details \label{sec:dset}}

\thispagestyle{empty}
\setcounter{equation}{0} %
\renewcommand{\theequation}{B.\arabic{equation}}
\setcounter{table}{0}
\renewcommand{\thetable}{B.\arabic{table}}

To assess our proposed method we select several datasets spanning a variety of decision-focused domains, as well as a dataset that is commonly used to evaluate MLX methods. We provide a brief description of each as follows:
\begin{itemize}
    \item \textbf{Bank}: A bank solicited customers via a phone call marketing campaign to entice them into subscribing to a term deposit. Predictive models derived from this data can be used to prioritize solicitation of potential customers in subsequent campaigns. The raw data is available at: \url{https://archive.ics.uci.edu/ml/datasets/bank+marketing} \citep{moro2014data}; $n=41188$, $p=61$,  $c\in\{\text{subscribe},\text{didn't subscribe}\}$, $\text{count}(c=\text{subscribe})=4640$. 
    \item \textbf{MIMIC}: A healthcare dataset consisting of patients experiencing sepsis. The models derived from this dataset can be used to help categorize patients as low/high risk and thus aid in treatment decisions. The raw data is available at: \url{https://physionet.org/content/mimiciii-demo/1.4/} \citep{johnson2016mimic}; $n=1122$, $p=25$, $c \in \{\text{alive},\text{dead}\}$, $\text{count}(c=\text{dead})=243$.
    \item \textbf{Movie}: A movie dataset that measures various film characteristics available only during the pre-production phase of development; box office profitability as the outcome. ML models can be used to help potential investors decided whether to invest a film or not. Obtained from \cite{lash2016early}; $n=2506$, $p=119$, $c \in \{\text{profitable},\text{not profitable}\}$, $\text{count}(c=\text{profitable})=757$.
    \item \textbf{News}: A text-based dataset consisting of different newsgroup posts. We adopt the posts from the Christianity and atheism newsgroups as in \cite{ribeiro2016should}. Features are the 300 most common corpus terms. The raw data is available at: \url{https://archive.ics.uci.edu/ml/datasets/Twenty+Newsgroups} \citep{joachims1996probabilistic}; $n=1796$, $p=300$, $c \in \{\text{Christianity},\text{atheism}\}$, $\text{count}(c=\text{Christianity})=997$.
    \item \textbf{Student}: A dataset consisting of Portuguese high school students with final grades as outcomes. Predictive models can be used to prioritize/decide which students may need extra assistance. The raw data is available at: \url{https://archive.ics.uci.edu/ml/datasets/student+performance} \citep{cortez2008using}; $n=650$, $p=43$, $c \in \{\text{good grade},\allowbreak \text{not good grade}\}$, $\text{count}(c=\text{not good grade})=455$.
\end{itemize}

\section*{Appendix C: Machine Learning Model Training and Evaluation \label{sec:train}}

\thispagestyle{empty}
\setcounter{equation}{0} %
\renewcommand{\theequation}{C.\arabic{equation}}
\setcounter{table}{0}
\renewcommand{\thetable}{C.\arabic{table}}

We conduct our experiments using five different machine learning models, each of which is trained on each of our selected datasets. We partition each dataset randomly into training, validation, and testing, where 70\%, 15\%, and 15\% of the full dataset are allocated to each partition, respectively. The training set is used to train models, the validation set is used for parameter selection and training the reinforcement learning models, and the test set is used to assess explainability.

We perform a grid search to optimize predictive model performance, depending on the model (i.e., if hyperparameters are available). We provide the following details for each model and summarize the search space for each, where relevant:
\begin{itemize}
    \item \textbf{LR}: Optimized using ADAM with an initial learning rate of $0.01$, a batch size of 64, a maximum of 4000 epochs, with early stopping and a patience value of 200.
    \item \textbf{NN}: Fully connected layers, with relu activation, a single hidden layer, a batch size of 64, a maximum of 4000 epochs, with early stopping and a patience value of 200, optimized used ADAM with an initial learning rate of $0.01$. We search across the following hidden node numbers: 3, 5, 10, 25, 50, 100.
    \item \textbf{SVM}: We perform a search over the kernel, where $kernel \in \{\text{linear},\text{RBF},\text{polynomial}\}$. A degree of 2 was used with the polynomial kernel and the gamma parameter of the RBF kernel was set to $\frac{1}{p}$. We further impose miss-classification penalties on the loss according to proportional class imbalance -- i.e., the penalty imposed on positive miss-classifications is $p_{c^{+}}=\frac{\vert \mathbf{c} \vert }{\sum_{c \in \mathbf{c}^-} 1}$ and $p_{c^{-}}= p_{c^{+}}^{-1}$. 
    \item \textbf{DT}: We employ CART, using the gini index as the splitting criteria with a minimum number of leaf nodes of 2. We search across the following maximum depth values: 5, 10 , 30, 50. We also impose miss-classification penalties in a similar, but slightly different way to that of SVMs, where $p_{c^{+}}=\frac{\sum_{c \in \mathbf{c}^-} 1}{\vert \mathbf{c} \vert }$ and $p_{y^{-}}= 1 - p_{y^{+}}$.
    \item \textbf{RF}: We use a bag of CART classifiers described by DT, above, performing a full grid search across two parameters: maximum depth and the number of weak learners. We search across the same max depth values as that of DT and the following number of weak learners: 50, 100, 200, 400, 600. We also impose the same miss-classification penalties as that of DT.
\end{itemize}

We report the training, validation, and test set performance on each of the optimized classification models, below.
\begin{table}[!htp]
\centering
\begin{tabular}{lllllll}
\hline
\multicolumn{7}{c}{Bank}                                                                                                                                 \\ \hline
\multicolumn{1}{l|}{\multirow{2}{*}{}} & \multicolumn{2}{c|}{Train}                & \multicolumn{2}{c|}{Validation}          & \multicolumn{2}{c}{Test} \\ \cline{2-7} 
\multicolumn{1}{l|}{}                  & Accuracy  & \multicolumn{1}{l|}{AUC}      & Accuracy & \multicolumn{1}{l|}{AUC}      & Accuracy    & AUC        \\ \hline
\multicolumn{1}{l|}{LR}                & 0.9726986 & \multicolumn{1}{l|}{0.911175} & 0.973934 & \multicolumn{1}{l|}{0.912431} & 0.972485    & 0.912269   \\
\multicolumn{1}{l|}{NN}                & 0.9787762 & \multicolumn{1}{l|}{0.917349} & 0.978133 & \multicolumn{1}{l|}{0.912269} & 0.97627     & 0.912269   \\
\multicolumn{1}{l|}{DT}                & 0.8480855 & \multicolumn{1}{l|}{0.881607} & 0.84202  & \multicolumn{1}{l|}{0.87335}  & 0.849142    & 0.880499   \\
\multicolumn{1}{l|}{RF}                & 0.8788152 & \multicolumn{1}{l|}{0.912631} & 0.865976 & \multicolumn{1}{l|}{0.884968} & 0.870346    & 0.876768   \\
\multicolumn{1}{l|}{SVM}               & 0.9423904 & \multicolumn{1}{l|}{0.785564} & 0.905147 & \multicolumn{1}{l|}{0.677491} & 0.909032    & 0.687833   \\ \hline
\end{tabular}
\end{table}

\begin{table}[!htp]
\centering
\begin{tabular}{lllllll}
\hline
\multicolumn{7}{c}{Mimic}                                                                                                                               \\ \hline
\multicolumn{1}{l|}{\multirow{2}{*}{}} & \multicolumn{2}{c|}{Train}               & \multicolumn{2}{c|}{Validation}          & \multicolumn{2}{c}{Test} \\ \cline{2-7} 
\multicolumn{1}{l|}{}                  & Accuracy & \multicolumn{1}{l|}{AUC}      & Accuracy & \multicolumn{1}{l|}{AUC}      & Accuracy    & AUC        \\ \hline
\multicolumn{1}{l|}{LR}                & 0.941238 & \multicolumn{1}{l|}{0.877551} & 0.91674  & \multicolumn{1}{l|}{0.840237} & 0.955639    & 0.905325   \\
\multicolumn{1}{l|}{NN}                & 0.949067 & \multicolumn{1}{l|}{0.882653} & 0.920311 & \multicolumn{1}{l|}{0.846154} & 0.944155    & 0.905325   \\
\multicolumn{1}{l|}{DT}                & 0.859694 & \multicolumn{1}{l|}{0.863633} & 0.757396 & \multicolumn{1}{l|}{0.65991}  & 0.757396    & 0.679361   \\
\multicolumn{1}{l|}{RF}                & 0.936224 & \multicolumn{1}{l|}{0.91462}  & 0.828402 & \multicolumn{1}{l|}{0.734541} & 0.881657    & 0.836712   \\
\multicolumn{1}{l|}{SVM}               & 0.922194 & \multicolumn{1}{l|}{0.829096} & 0.840237 & \multicolumn{1}{l|}{0.693489} & 0.87574     & 0.77457    \\ \hline
\end{tabular}
\end{table}

\begin{table}[!htp]
\centering
\begin{tabular}{lllllll}
\hline
\multicolumn{7}{c}{Movie}                                                                                                                               \\ \hline
\multicolumn{1}{l|}{\multirow{2}{*}{}} & \multicolumn{2}{c|}{Train}               & \multicolumn{2}{c|}{Validation}          & \multicolumn{2}{c}{Test} \\ \cline{2-7} 
\multicolumn{1}{l|}{}                  & Accuracy & \multicolumn{1}{l|}{AUC}      & Accuracy & \multicolumn{1}{l|}{AUC}      & Accuracy    & AUC        \\ \hline
\multicolumn{1}{l|}{LR}                & 0.900309 & \multicolumn{1}{l|}{0.812999} & 0.903491 & \multicolumn{1}{l|}{0.819149} & 0.873911    & 0.829787   \\
\multicolumn{1}{l|}{NN}                & 0.912744 & \multicolumn{1}{l|}{0.831243} & 0.849536 & \multicolumn{1}{l|}{0.789894} & 0.805978    & 0.726064   \\
\multicolumn{1}{l|}{DT}                & 0.855188 & \multicolumn{1}{l|}{0.829737} & 0.829787 & \multicolumn{1}{l|}{0.793625} & 0.816489    & 0.776651   \\
\multicolumn{1}{l|}{RF}                & 0.8626   & \multicolumn{1}{l|}{0.816247} & 0.848404 & \multicolumn{1}{l|}{0.787164} & 0.837766    & 0.784485   \\
\multicolumn{1}{l|}{SVM}               & 0.794185 & \multicolumn{1}{l|}{0.676512} & 0.81117  & \multicolumn{1}{l|}{0.698507} & 0.803191    & 0.692782   \\ \hline
\end{tabular}
\end{table}

\begin{table}[!htp]
\centering
\begin{tabular}{lllllll}
\hline
\multicolumn{7}{c}{News}                                                                                                                                \\ \hline
\multicolumn{1}{l|}{\multirow{2}{*}{}} & \multicolumn{2}{c|}{Train}               & \multicolumn{2}{c|}{Validation}          & \multicolumn{2}{c}{Test} \\ \cline{2-7} 
\multicolumn{1}{l|}{}                  & Accuracy & \multicolumn{1}{l|}{AUC}      & Accuracy & \multicolumn{1}{l|}{AUC}      & Accuracy    & AUC        \\ \hline
\multicolumn{1}{l|}{LR}                & 0.99691  & \multicolumn{1}{l|}{0.974522} & 0.989095 & \multicolumn{1}{l|}{0.948148} & 0.986927    & 0.937037   \\
\multicolumn{1}{l|}{NN}                & 0.998119 & \multicolumn{1}{l|}{0.984873} & 0.989376 & \multicolumn{1}{l|}{0.951852} & 0.98011     & 0.937037   \\
\multicolumn{1}{l|}{DT}                & 0.944268 & \multicolumn{1}{l|}{0.938628} & 0.925926 & \multicolumn{1}{l|}{0.92}     & 0.892593    & 0.883333   \\
\multicolumn{1}{l|}{RF}                & 0.984076 & \multicolumn{1}{l|}{0.982111} & 0.959259 & \multicolumn{1}{l|}{0.955833} & 0.92963     & 0.921667   \\
\multicolumn{1}{l|}{SVM}               & 0.999204 & \multicolumn{1}{l|}{0.999106} & 0.977778 & \multicolumn{1}{l|}{0.976667} & 0.948148    & 0.944167   \\ \hline
\end{tabular}
\end{table}

\begin{table}[!htp]
\centering
\begin{tabular}{lllllll}
\hline
\multicolumn{7}{c}{Student}                                                                                                                             \\ \hline
\multicolumn{1}{l|}{\multirow{2}{*}{}} & \multicolumn{2}{c|}{Train}               & \multicolumn{2}{c|}{Validation}          & \multicolumn{2}{c}{Test} \\ \cline{2-7} 
\multicolumn{1}{l|}{}                  & Accuracy & \multicolumn{1}{l|}{AUC}      & Accuracy & \multicolumn{1}{l|}{AUC}      & Accuracy    & AUC        \\ \hline
\multicolumn{1}{l|}{LR}                & 0.865161 & \multicolumn{1}{l|}{0.769231} & 0.841747 & \multicolumn{1}{l|}{0.742268} & 0.79966     & 0.742268   \\
\multicolumn{1}{l|}{NN}                & 0.887303 & \multicolumn{1}{l|}{0.782418} & 0.831013 & \multicolumn{1}{l|}{0.752577} & 0.786003    & 0.71134    \\
\multicolumn{1}{l|}{DT}                & 0.683516 & \multicolumn{1}{l|}{0.751095} & 0.701031 & \multicolumn{1}{l|}{0.757099} & 0.628866    & 0.666075   \\
\multicolumn{1}{l|}{RF}                & 0.828571 & \multicolumn{1}{l|}{0.856652} & 0.804124 & \multicolumn{1}{l|}{0.82074}  & 0.690722    & 0.680527   \\
\multicolumn{1}{l|}{SVM}               & 0.87033  & \multicolumn{1}{l|}{0.821052} & 0.71134  & \multicolumn{1}{l|}{0.665568} & 0.670103    & 0.576826   \\ \hline
\end{tabular}
\end{table}

\section*{Appendix D: Additional Experiment Model Parameterization Details}

\thispagestyle{empty}
\setcounter{equation}{0} %
\renewcommand{\theequation}{D.\arabic{equation}}
\setcounter{table}{0}
\renewcommand{\thetable}{D.\arabic{table}}

Several additional experiment details are provided below, including parameterization of the different models attempted. All code was written and executed in Python 3.6.
\begin{itemize}
    \item \textbf{DRL Models}: We generally found two hidden layers with 50 hidden units each to work well across our various datasets and DRL methods and therefore adopted this parameterization for all DRL models. Each model was optimized using the Adam optimizer with default parameterization. Models were constructed, trained, etc. using Tensorflow 2.4.
    \item \textbf{LIME}: We obtained the code from the \cite{ribeiro2016should} Github repository at \url{https://github.com/marcotcr/lime} (installed using PyPI). We found the following parameterized to work well and adopted it for use with all models and datasets: $num\_samples=300, num\_features=p$.
    \item \textbf{Grow}: We obtained the code from \cite{laugel2018comparison} Github repository at \url{https://github.com/thibaultlaugel/growingspheres}. We found the following parameterization to work well and adopted it for use with all ML models and datasets: $n\_in\_layer=200$, $first\_radius=1.1$, $dicrease\_radius=2.0$.
\end{itemize}

\section*{Appendix E: Decider-free Testing Rewards}

\thispagestyle{empty}
\setcounter{equation}{0} %
\renewcommand{\theequation}{E.\arabic{equation}}
\setcounter{table}{0}
\renewcommand{\thetable}{E.\arabic{table}}

Table \ref{tab:free-test-rewards}, below, shows the average reward over 10 trials, with 100 test instances in each trial, by explainability method, ML method, and dataset for our ``decider free`` explainability scenario. \textbf{Bold} shows the highest reward obtain by an explainability method for each ML model-dataset combination.
\begin{longtable}[htp]
{|cl|lllll|}
\hline
\multicolumn{2}{|c|}{} & \multicolumn{5}{c|}{Model} \\ \hline
\multicolumn{1}{|l|}{Dataset} & \multicolumn{1}{c|}{Method} & \multicolumn{1}{c}{NN} & \multicolumn{1}{c}{SVM} & \multicolumn{1}{c}{RF} & \multicolumn{1}{c}{LR} & \multicolumn{1}{c|}{DT} \\ \hline
\multicolumn{1}{|c|}{\multirow{6}{*}{Bank}} & DDPG & 8.185 & 8.291 & 8.188 & 7.934 & 8.137 \\
\multicolumn{1}{|c|}{} & Our DDPG & \textbf{8.453} & \textbf{8.54} & \textbf{8.756} & \textbf{8.207} & \textbf{8.367} \\
\multicolumn{1}{|c|}{} & TD3 & 7.941 & 8.006 & 8.11 & 7.642 & 7.874 \\
\multicolumn{1}{|c|}{} & Our TD3 & 8.216 & 8.271 & 8.564 & 7.959 & 8.104 \\
\multicolumn{1}{|c|}{} & LIME & -1.392 & -0.982 & -0.693 & -1.197 & -1.113 \\
\multicolumn{1}{|c|}{} & Grow & -3.535 & -2.451 & -1.751 & -3.718 & -2 \\ \hline
\multicolumn{1}{|c|}{\multirow{6}{*}{MIMIC-3}} & DDPG & \textbf{8.072} & 7.982 & \textbf{8.545} & \textbf{8.004} & 7.92 \\
\multicolumn{1}{|c|}{} & Our DDPG & 7.936 & \textbf{8.474} & \textbf{8.545} & 7.902 & \textbf{7.976} \\
\multicolumn{1}{|c|}{} & TD3 & 7.943 & 7.899 & 8.367 & 7.955 & 7.735 \\
\multicolumn{1}{|c|}{} & Our TD3 & 7.892 & 8.255 & 8.359 & 7.95 & 7.765 \\
\multicolumn{1}{|c|}{} & LIME & -1.452 & -0.734 & 0.1 & -1.158 & 1.463 \\
\multicolumn{1}{|c|}{} & Grow & -3.445 & -2.151 & -1.538 & -3.725 & -1.724 \\ \hline
\multicolumn{1}{|c|}{\multirow{6}{*}{Movie}} & DDPG & 6.114 & 7.752 & 8.174 & 2.782 & 7.465 \\
\multicolumn{1}{|c|}{} & Our DDPG & \textbf{7.837} & \textbf{7.876} & \textbf{8.312} & \textbf{7.5} & \textbf{7.577} \\
\multicolumn{1}{|c|}{} & TD3 & 6.139 & 7.291 & 7.831 & 2.937 & 7.134 \\
\multicolumn{1}{|c|}{} & Our TD3 & 7.623 & 7.411 & 7.982 & 7.121 & 7.256 \\
\multicolumn{1}{|c|}{} & LIME & -0.737 & -1.82 & -1.167 & -2.287 & -1.21 \\
\multicolumn{1}{|c|}{} & Grow & -4.156 & -1.995 & -1.32 & -3.019 & -1.63 \\ \hline
\multicolumn{1}{|c|}{\multirow{6}{*}{Student}} & DDPG & 1.774 & 2.786 & 4.342 & 0.635 & \textbf{7.889} \\
\multicolumn{1}{|c|}{} & Our DDPG & 5.166 & 2.306 & 5.869 & 5.005 & 6.575 \\
\multicolumn{1}{|c|}{} & TD3 & 1.775 & \textbf{2.789} & 4.568 & 0.712 & 7.708 \\
\multicolumn{1}{|c|}{} & Our TD3 & \textbf{5.365} & 2.505 & \textbf{5.898} & \textbf{5.139} & 6.769 \\
\multicolumn{1}{|c|}{} & LIME & 1.416 & -0.713 & -1.117 & 2.086 & -0.249 \\
\multicolumn{1}{|c|}{} & Grow & -4.186 & -2.402 & -1.853 & -3.589 & -1.813 \\ \hline
\multicolumn{1}{|c|}{\multirow{6}{*}{News}} & DDPG & \textbf{7.003} & 4.149 & 7.717 & \textbf{7.24} & 6.93 \\
\multicolumn{1}{|c|}{} & Our DDPG & 4.786 & 4.6 & \textbf{7.831} & 5.928 & \textbf{6.985} \\
\multicolumn{1}{|c|}{} & TD3 & 6.473 & 4.122 & 7.298 & 6.766 & 6.396 \\
\multicolumn{1}{|c|}{} & Our TD3 & 5.08 & \textbf{4.657} & 7.509 & 6.179 & 6.659 \\
\multicolumn{1}{|c|}{} & LIME & 1.971 & 0.671 & 2.378 & 2.402 & 3.205 \\
\multicolumn{1}{|c|}{} & Grow & -5.433 & -3.544 & -2.133 & -5.382 & -2.19 \\ \hline
\caption{Average test rewards. \textbf{Bold} shows the largest average reward on each dataset-ML model combination. \label{tab:free-test-rewards}}
\end{longtable}

\section*{Appendix F: HITL Decider Testing Rewards}

\thispagestyle{empty}
\setcounter{equation}{0} %
\renewcommand{\theequation}{F.\arabic{equation}}
\setcounter{table}{0}
\renewcommand{\thetable}{F.\arabic{table}}

Table \ref{tab:hitl-test-rewards}, below, shows the results of our HITL decider scenario in terms of average reward taken over 10 trials, with 100 test instances in each trial, by explainability method, ML method, and dataset. Results for our HEX methods are also  shown by UAP scenario. Results for the out-of-the-box DRL, LIME, and Grow methods are shown without considering the UAP since these methods do not handle such considerations.
\begin{longtable}[!htp]
{|ccl|ccccc|}
\hline
\multicolumn{3}{|c|}{} & \multicolumn{5}{c|}{Model} \\ \hline
\multicolumn{1}{|l|}{Dataset} & \multicolumn{1}{l|}{\% Corrupt} & Method & NN & SVM & RF & LR & DT \\ \hline
\multicolumn{1}{|c|}{} & \multicolumn{1}{c|}{} & GS-DDPG & \textbf{6.718} & \textbf{7.117} & \textbf{6.75} & \textbf{5.991} & \textbf{6.896} \\
\multicolumn{1}{|c|}{} & \multicolumn{1}{c|}{\multirow{-2}{*}{0.3}} & GS-TD3 & 6.597 & 6.834 & 6.627 & 5.84 & 6.812 \\ \cline{2-8} 
\multicolumn{1}{|c|}{} & \multicolumn{1}{c|}{} & GS-DDPG & 6.358 & 6.755 & 5.961 & 5.558 & 6.122 \\
\multicolumn{1}{|c|}{} & \multicolumn{1}{c|}{\multirow{-2}{*}{0.5}} & GS-TD3 & 6.207 & 6.707 & 5.91 & 5.301 & 6.016 \\ \cline{2-8} 
\multicolumn{1}{|c|}{} & \multicolumn{1}{c|}{} & GS-DDPG & 4.618 & 5.266 & 4.022 & 3.863 & 4.757 \\
\multicolumn{1}{|c|}{} & \multicolumn{1}{c|}{\multirow{-2}{*}{0.7}} & GS-TD3 & 4.548 & 5.221 & 4.065 & 3.669 & 4.727 \\ \cline{2-8} 
\multicolumn{1}{|c|}{} & \multicolumn{1}{c|}{} & DDPG & {\color[HTML]{0070C0} \textbf{8.185}} & {\color[HTML]{0070C0} \textbf{8.291}} & {\color[HTML]{0070C0} \textbf{8.188}} & {\color[HTML]{0070C0} \textbf{7.934}} & {\color[HTML]{0070C0} \textbf{8.137}} \\
\multicolumn{1}{|c|}{} & \multicolumn{1}{c|}{} & TD3 & 7.941 & 8.006 & 8.11 & 7.642 & 7.874 \\
\multicolumn{1}{|c|}{} & \multicolumn{1}{c|}{} & LIME & -1.392 & -0.982 & -0.693 & -1.197 & -1.113 \\
\multicolumn{1}{|c|}{\multirow{-10}{*}{Bank}} & \multicolumn{1}{c|}{\multirow{-4}{*}{None}} & Grow & -3.535 & -2.451 & -1.751 & -3.718 & -2 \\ \hline
\multicolumn{1}{|c|}{} & \multicolumn{1}{c|}{} & GS-DDPG & {\color[HTML]{FF0000} \textbf{8.074}} & 8.787 & \textbf{8.222} & 7.788 & \textbf{7.702} \\
\multicolumn{1}{|c|}{} & \multicolumn{1}{c|}{\multirow{-2}{*}{0.3}} & GS-TD3 & 8.062 & 8.593 & 8.186 & \textbf{7.86} & 7.654 \\ \cline{2-8} 
\multicolumn{1}{|c|}{} & \multicolumn{1}{c|}{} & GS-DDPG & 7.623 & 8.978 & 5.788 & 6.888 & 6.597 \\
\multicolumn{1}{|c|}{} & \multicolumn{1}{c|}{\multirow{-2}{*}{0.5}} & GS-TD3 & 7.623 & 8.841 & 5.845 & 6.93 & 6.556 \\ \cline{2-8} 
\multicolumn{1}{|c|}{} & \multicolumn{1}{c|}{} & GS-DDPG & 7.838 & {\color[HTML]{FF0000} \textbf{9.074}} & 7.504 & 7.458 & 6.818 \\
\multicolumn{1}{|c|}{} & \multicolumn{1}{c|}{\multirow{-2}{*}{0.7}} & GS-TD3 & 7.933 & 8.952 & 7.527 & 7.714 & 6.832 \\ \cline{2-8} 
\multicolumn{1}{|c|}{} & \multicolumn{1}{c|}{} & DDPG & \textbf{8.072} & \textbf{7.982} & {\color[HTML]{0070C0} \textbf{8.545}} & {\color[HTML]{0070C0} \textbf{8.004}} & {\color[HTML]{0070C0} \textbf{7.92}} \\
\multicolumn{1}{|c|}{} & \multicolumn{1}{c|}{} & TD3 & 7.943 & 7.899 & 8.367 & 7.955 & 7.735 \\
\multicolumn{1}{|c|}{} & \multicolumn{1}{c|}{} & LIME & -1.452 & -0.734 & 0.1 & -1.158 & 1.463 \\
\multicolumn{1}{|c|}{\multirow{-10}{*}{MIMIC}} & \multicolumn{1}{c|}{\multirow{-4}{*}{None}} & Grow & -3.445 & -2.151 & -1.538 & -3.725 & -1.724 \\ \hline
\multicolumn{1}{|c|}{} & \multicolumn{1}{c|}{} & GS-DDPG & 8.077 & 8.218 & {\color[HTML]{FF0000} \textbf{8.26}} & 7.725 & \textbf{7.443} \\
\multicolumn{1}{|c|}{} & \multicolumn{1}{c|}{\multirow{-2}{*}{0.3}} & GS-TD3 & 7.999 & 7.841 & 8.021 & 7.508 & 7.269 \\ \cline{2-8} 
\multicolumn{1}{|c|}{} & \multicolumn{1}{c|}{} & GS-DDPG & 8.237 & 8.387 & 8.146 & 8.012 & 7.197 \\
\multicolumn{1}{|c|}{} & \multicolumn{1}{c|}{\multirow{-2}{*}{0.5}} & GS-TD3 & 8.039 & 8.072 & 7.925 & 7.742 & 7.023 \\ \cline{2-8} 
\multicolumn{1}{|c|}{} & \multicolumn{1}{c|}{} & GS-DDPG & {\color[HTML]{FF0000} \textbf{8.352}} & {\color[HTML]{FF0000} \textbf{8.622}} & 7.76 & {\color[HTML]{FF0000} \textbf{8.081}} & 6.609 \\
\multicolumn{1}{|c|}{} & \multicolumn{1}{c|}{\multirow{-2}{*}{0.7}} & GS-TD3 & 8.323 & 8.305 & 7.756 & 7.842 & 6.492 \\ \cline{2-8} 
\multicolumn{1}{|c|}{} & \multicolumn{1}{c|}{} & DDPG & \textbf{6.114} & \textbf{7.752} & \textbf{8.174} & 2.782 & {\color[HTML]{0070C0} \textbf{7.465}} \\
\multicolumn{1}{|c|}{} & \multicolumn{1}{c|}{} & TD3 & 6.139 & 7.291 & 7.831 & \textbf{2.937} & 7.134 \\
\multicolumn{1}{|c|}{} & \multicolumn{1}{c|}{} & LIME & -0.737 & -1.82 & -1.167 & -2.287 & -1.21 \\
\multicolumn{1}{|c|}{\multirow{-10}{*}{Movie}} & \multicolumn{1}{c|}{\multirow{-4}{*}{None}} & Grow & -4.156 & -1.995 & -1.32 & -3.019 & -1.63 \\ \hline
\multicolumn{1}{|c|}{} & \multicolumn{1}{c|}{} & GS-DDPG & 5.299 & \textbf{2.521} & {\color[HTML]{FF0000} \textbf{5.51}} & 5.386 & 6.12 \\
\multicolumn{1}{|c|}{} & \multicolumn{1}{c|}{\multirow{-2}{*}{0.3}} & GS-TD3 & {\color[HTML]{FF0000} \textbf{5.347}} & 2.461 & 5.427 & 5.469 & \textbf{6.28} \\ \cline{2-8} 
\multicolumn{1}{|c|}{} & \multicolumn{1}{c|}{} & GS-DDPG & 5.269 & 2.394 & 4.671 & 5.445 & 5.249 \\
\multicolumn{1}{|c|}{} & \multicolumn{1}{c|}{\multirow{-2}{*}{0.5}} & GS-TD3 & 5.233 & 2.444 & 4.883 & {\color[HTML]{FF0000} \textbf{5.553}} & 5.15 \\ \cline{2-8} 
\multicolumn{1}{|c|}{} & \multicolumn{1}{c|}{} & GS-DDPG & 4.503 & 1.915 & 4.268 & 4.283 & 4.612 \\
\multicolumn{1}{|c|}{} & \multicolumn{1}{c|}{\multirow{-2}{*}{0.7}} & GS-TD3 & 4.527 & 2.003 & 4.224 & 4.308 & 4.661 \\ \cline{2-8} 
\multicolumn{1}{|c|}{} & \multicolumn{1}{c|}{} & DDPG & 1.774 & 2.786 & 4.342 & 0.635 & {\color[HTML]{0070C0} \textbf{7.889}} \\
\multicolumn{1}{|c|}{} & \multicolumn{1}{c|}{} & TD3 & \textbf{1.775} & {\color[HTML]{0070C0} \textbf{2.789}} & \textbf{4.568} & 0.712 & 7.708 \\
\multicolumn{1}{|c|}{} & \multicolumn{1}{c|}{} & LIME & 1.416 & -0.713 & -1.117 & \textbf{2.086} & -0.249 \\
\multicolumn{1}{|c|}{\multirow{-10}{*}{Student}} & \multicolumn{1}{c|}{\multirow{-4}{*}{None}} & Grow & -4.186 & -2.402 & -1.853 & -3.589 & -1.813 \\ \hline
\multicolumn{1}{|c|}{} & \multicolumn{1}{c|}{} & GS-DDPG & 4.436 & 4.811 & {\color[HTML]{FF0000} \textbf{7.626}} & 5.675 & \textbf{6.812} \\
\multicolumn{1}{|c|}{} & \multicolumn{1}{c|}{\multirow{-2}{*}{0.3}} & GS-TD3 & \textbf{4.594} & {\color[HTML]{FF0000} \textbf{4.905}} & 7.515 & {\color[HTML]{000000} \textbf{6.005}} & 6.678 \\ \cline{2-8} 
\multicolumn{1}{|c|}{} & \multicolumn{1}{c|}{} & GS-DDPG & 3.98 & 4.743 & 6.902 & 5.439 & 6.324 \\
\multicolumn{1}{|c|}{} & \multicolumn{1}{c|}{\multirow{-2}{*}{0.5}} & GS-TD3 & 3.979 & 4.692 & 7.026 & 5.448 & 6.326 \\ \cline{2-8} 
\multicolumn{1}{|c|}{} & \multicolumn{1}{c|}{} & GS-DDPG & 3.733 & 4.674 & 6.57 & 5.156 & 5.123 \\
\multicolumn{1}{|c|}{} & \multicolumn{1}{c|}{\multirow{-2}{*}{0.7}} & GS-TD3 & 3.871 & 4.724 & 6.65 & 5.332 & 5.138 \\ \cline{2-8} 
\multicolumn{1}{|c|}{} & \multicolumn{1}{c|}{} & DDPG & {\color[HTML]{0070C0} \textbf{7.003}} & \textbf{4.149} & \textbf{7.717} & {\color[HTML]{0070C0} \textbf{7.24}} & {\color[HTML]{0070C0} \textbf{6.93}} \\
\multicolumn{1}{|c|}{} & \multicolumn{1}{c|}{} & TD3 & 6.473 & 4.122 & 7.298 & 6.766 & 6.396 \\
\multicolumn{1}{|c|}{} & \multicolumn{1}{c|}{} & LIME & 1.971 & 0.671 & 2.378 & 2.402 & 3.205 \\
\multicolumn{1}{|c|}{\multirow{-10}{*}{News}} & \multicolumn{1}{c|}{\multirow{-4}{*}{None}} & Grow & -5.433 & -3.544 & -2.133 & -5.382 & -2.19 \\ \hline
\caption{HITL scenario test rewards. \textbf{Bold} indicates the lowest reward among all HEX methods and all non-HEX methods by dataset and ML model (HEX/non-HEX considered separately). \textbf{\textcolor{blue}{Blue}} indicates that a non-HEX method received the lowest overall within-ML method reward, while \textbf{\textcolor{red}{red}} indicates that a HEX method received the lowest overall within-method reward.\label{tab:hitl-test-rewards}}
\end{longtable}

\bibliography{references}

\end{document}